\documentclass[sigconf]{aamas}

\usepackage{balance} %

\usepackage{mathtools}          %
\usepackage{mathrsfs}           %
\mathtoolsset{showonlyrefs}     %
\usepackage{graphicx}           %
\usepackage{subcaption}         %
\usepackage[space]{grffile}     %
\usepackage{url}                %

\usepackage{hyperref}
\usepackage{array}
\usepackage{booktabs}
\usepackage{multirow}
\usepackage{xspace}
\usepackage{amsmath}
\usepackage{cleveref}
\usepackage{caption,adjustbox}
\usepackage[
    separate-uncertainty=true,
    detect-weight,
    table-number-alignment=center,
	table-text-alignment=center,
	table-format=1.2(2),
    table-sign-mantissa,
    group-digits=integer,
    group-minimum-digits=4,
    group-separator={,},
]{siunitx}

\definecolor{mydarkblue}{rgb}{0,0.08,0.45}
\definecolor{mydarkgreen}{rgb}{0,0.45,0.15}
\hypersetup{colorlinks,linkcolor=red,urlcolor=mydarkblue,linktoc=page,citecolor=mydarkgreen}

\usepackage{amsmath,amsfonts,bm}

\def\eqref#1{equation~\ref{#1}}

\def\1{\bm{1}}

\DeclareMathAlphabet{\mathsfit}{\encodingdefault}{\sfdefault}{m}{sl}
\SetMathAlphabet{\mathsfit}{bold}{\encodingdefault}{\sfdefault}{bx}{n}

\setcopyright{none}
\acmConference[ALA '25]{Proc.\@ of the Adaptive and Learning Agents Workshop (ALA 2025)}{May 19 -- 20, 2025}{Detroit, Michigan, USA, ala-workshop.github.io}{Avalos, Aydeniz, M\"uller, Mohammedalamen (eds.)}
\copyrightyear{2025}
\acmYear{2025}
\acmDOI{}
\acmPrice{}
\acmISBN{}
\acmSubmissionID{14}

\title[Visual Encoders for Imitation Learning in Modern Video Games]{Visual Encoders for Data-Efficient Imitation Learning in Modern Video Games}

\author{Lukas Sch\"afer}
\affiliation{
  \institution{Microsoft Research}
  \city{Cambridge}
  \country{United Kingdom}}
\authornote{Contact at lukas.schaefer@microsoft.com.}

\author{Logan Jones}
\affiliation{
  \institution{Microsoft Gaming}
  \city{Redmond}
  \country{United States}}

\author{Anssi Kanervisto}
\affiliation{
  \institution{Microsoft Research}
  \city{Cambridge}
  \country{United Kingdom}}

\author{Yuhan Cao}
\affiliation{
  \institution{Microsoft Research}
  \city{Cambridge}
  \country{United Kingdom}}

\author{Tabish Rashid}
\affiliation{
  \institution{Microsoft Research}
  \city{Cambridge}
  \country{United Kingdom}}

\author{Raluca Georgescu}
\affiliation{
  \institution{Microsoft Research}
  \city{Cambridge}
  \country{United Kingdom}}

\author{David Bignell}
\affiliation{
  \institution{Microsoft Research}
  \city{Cambridge}
  \country{United Kingdom}}

\author{Siddhartha Sen}
\affiliation{
  \institution{Microsoft Research}
  \city{New York}
  \country{United States}}

\author{Andrea Treviño Gavito}
\affiliation{
  \institution{Microsoft Gaming}
  \city{Redmond}
  \country{United States}}

\author{Sam Devlin}
\affiliation{
  \institution{Microsoft Research}
  \city{Cambridge}
  \country{United Kingdom}}

\begin{abstract}
    Video games have served as useful benchmarks for the decision-making community, but going beyond Atari games towards modern games has been prohibitively expensive for the vast majority of the research community. Prior work in modern video games typically relied on game-specific integration to obtain game features and enable online training, or on existing large datasets. An alternative approach is to train agents using imitation learning to play video games purely from images. However, this setting poses a fundamental question: which visual encoders obtain representations that retain information critical for decision making? To answer this question, we conduct a systematic study of imitation learning with publicly available pre-trained visual encoders compared to the typical task-specific end-to-end training approach in Minecraft, Counter-Strike: Global Offensive, and Minecraft Dungeons. Our results show that end-to-end training can be effective with comparably low-resolution images and only minutes of demonstrations, but significant improvements can be gained by utilising pre-trained encoders such as DINOv2 depending on the game. In addition to enabling effective decision making, we show that pre-trained encoders can make decision-making research in video games more accessible by significantly reducing the cost of training.
\end{abstract}


\newcommand{\BibTeX}{\rm B\kern-.05em{\sc i\kern-.025em b}\kern-.08em\TeX}

\newcommand{\data}{\mathcal{D}}
\newcommand{\loss}{\mathcal{L}}
\newcommand{\pol}{\pi}

\newcommand{\citeneeded}[1]{\textsuperscript{\textcolor{red}{[citation needed]}}\xspace}

\begin{document}

\pagestyle{fancy}
\fancyhead{}

\maketitle

\section{Introduction}
Video games have served as useful benchmarks for the decision-making community, training agents in complex games using reinforcement learning (RL)~\citep{vinyals2019grandmaster,berner2019dota, wurman2022outracing}, imitation learning (IL)~\citep{kanervisto2020benchmarking,pearce2022counter,sestini2022towards}, or a combination of both paradigms~\citep{baker2022vpt,fan2022minedojo}. Beyond representing a valuable benchmark for complex decision-making tasks, video games represent a vast entertainment industry with many commercial applications of AI agents, including in game development, game testing or game design~\citep{jacob2020s,gillberg2023technical}.

Prior research in video games often necessitated close game-specific integration to obtain features and establish a scalable interface for training agents. However, game-specific integration introduces significant cost and requires domain expertise and engineering efforts. To enable decision-making agents for video games without depending on game-specific integration, we focus on training agents to play video games in a human-like manner, receiving only images from the game and producing actions corresponding to controller joystick and button inputs. This framework allows us to train agents entirely offline with behaviour cloning (BC), utilising previously collected human gameplay demonstrations, and evaluate agents without the need for game-specific integration. However, efficient training of agents from video game images necessitates a lower-dimensional representation of high-dimensional images. This motivates our main research question:

\begin{center}
    \emph{Which visual encoders learn representations that retain information for data-efficient decision making in modern video games?}
\end{center}

To answer this question, we conduct a comprehensive empirical study of 22 visual encoders in three modern video games: Minecraft~\citep{guss2019minerl}, Counter-Strike: Global Offensive (CS:GO)~\citep{pearce2022counter}, and Minecraft Dungeons. First, we examine which visual encoders enable effective decision-making when trained end-to-end from the BC loss. Considered encoders vary in network architecture (ResNets~\citep{he2016deep,liu2022convnext} or ViTs~\citep{dosovitskiy2021vit,steiner2022train}), image size, and the application of image augmentations. Second, we study 10 visual encoders that are pre-trained on large datasets of diverse real-world images. These pre-trained encoders promise generalisable representation without additional training and have shown promise in decision-making tasks~\citep{nair2022r3m,parisi2022unsurprising} but it is currently unclear whether these findings transfer to video games that represent a significant domain shift from the real-world images these encoders have been trained on. We identify four categories of pre-trained visual encoders and train agents using the representations from 10 different pre-trained encoders spanning all these categories: self-supervised trained encoders, language-contrastive trained encoders, supervised trained encoders, and reconstruction trained encoders. Third, motivated by the cost of collecting human gameplay data, we investigate the efficacy of these visual encoders for decision making when trained on smaller datasets. Our findings can be summaries as follows:

\begin{itemize}
    \item End-to-end trained encoders: small $128\times128$ images are sufficient to enable decision-making in games and image augmentation can significantly improve the performance.
    \item Pre-trained encoders: DINOv2~\citep{oquab2023dinov2} is among the best pre-trained encoder across all games and outperforms end-to-end trained encoders in Minecraft and Minecraft Dungeons.
    \item Data efficiency: capable decision-making agents can be trained with as little as 5 minutes of human gameplay data.%
    \item Training cost: pre-trained encoders can significantly reduce the cost of training in terms of time and memory requirements through pre-computing embeddings on the dataset.
\end{itemize}

Our approach and study provides a general framework for training decision-making agents from visual inputs in modern video games, and contributes valuable findings that facilitate further research. In particular, the strong performance of pre-trained encoders alongside their comparably little training cost is encouraging and significantly reduces the barrier of entry to conduct research on decision-making agents in video games.

\section{Related Work}
\label{sec:related_work}

\textbf{Learning Agents in Video Games.}
Video games have commonly served as benchmarks for decision-making agents, but training these agents typically assumed access to a programmatic interface for online training using reinforcement learning~\citep{vinyals2019grandmaster,berner2019dota} or large quantities of expert demonstrations used for offline imitation and supervised learning~\citep{baker2022vpt,fan2022minedojo,reed2022gato}. Recently, a plethora of work proposed to leverage pre-trained foundation models such as large language models as decision-making agents. The foundation models are either frozen or fine-tuned to directly act within their environment~\citep{wang2023voyager,lifshitz2023steve,cai2023groot,Team2024ScalingIA} or to collect training data~\citep{cai2023open}, but these works all rely on extensive datasets, game-specific integration, or both. Similar to our approach, \citet{pearce2022counter} and \citet{kanervisto2020benchmarking} evaluate imitation learning agents in videos games without programmatic interfaces. However, their studies only considered few and relatively simple visual encoders.

\textbf{Visual Encoders for Robotics.}
Prior research has studied the efficacy of visual encoders for imitation learning or reinforcement learning in robotic domains~\citep{nair2022r3m,parisi2022unsurprising,yuan2022pre,schneider2024surprising}. Most studies found that pre-trained encoders enable better generalisation and performance in decision-making agents than end-to-end trained visual encoders trained on smaller, task-specific data sets. However, not all studies align with these conclusions~\citep{schneider2024surprising}. These seemingly contradictory findings indicate that the question of which visual encoder works best depends for decision making is nuanced with the answer strongly depending on the underlying algorithm and setting. We further highlight that the images in the majority of robotics tasks and datasets strongly resemble the real world. This stands in contrast to video games that might feature highly stylised images. Therefore, it remains uncertain how these findings from robotics translate to video games that represents a strong domain shift from the training data of pre-trained visual encoders. Our study seeks to bridge this gap.

\textbf{Visual Encoders for Video Games.}
In the context of video games, pre-trained visual models have been employed to extract visual representations that differentiate between genres and styles \citep{trivedi2023towards}, indicating their ability to detect relevant features in games. However, domain-specific models trained using self-supervised representation learning techniques can yield higher-quality representations than certain pre-trained visual encoders \citep{trivedi2022learning}. Our study expands upon previous experiments\footnote{Code at \url{https://github.com/microsoft/imitation_learning_in_modern_video_games}.} by concentrating on modern video games and examining a broad spectrum of recent pre-trained and end-to-end trained visual encoder architectures.

\section{Imitation Learning for Video Games from Pixels}

\begin{figure}[t]
    \centering
    \includegraphics[width=.7\linewidth]{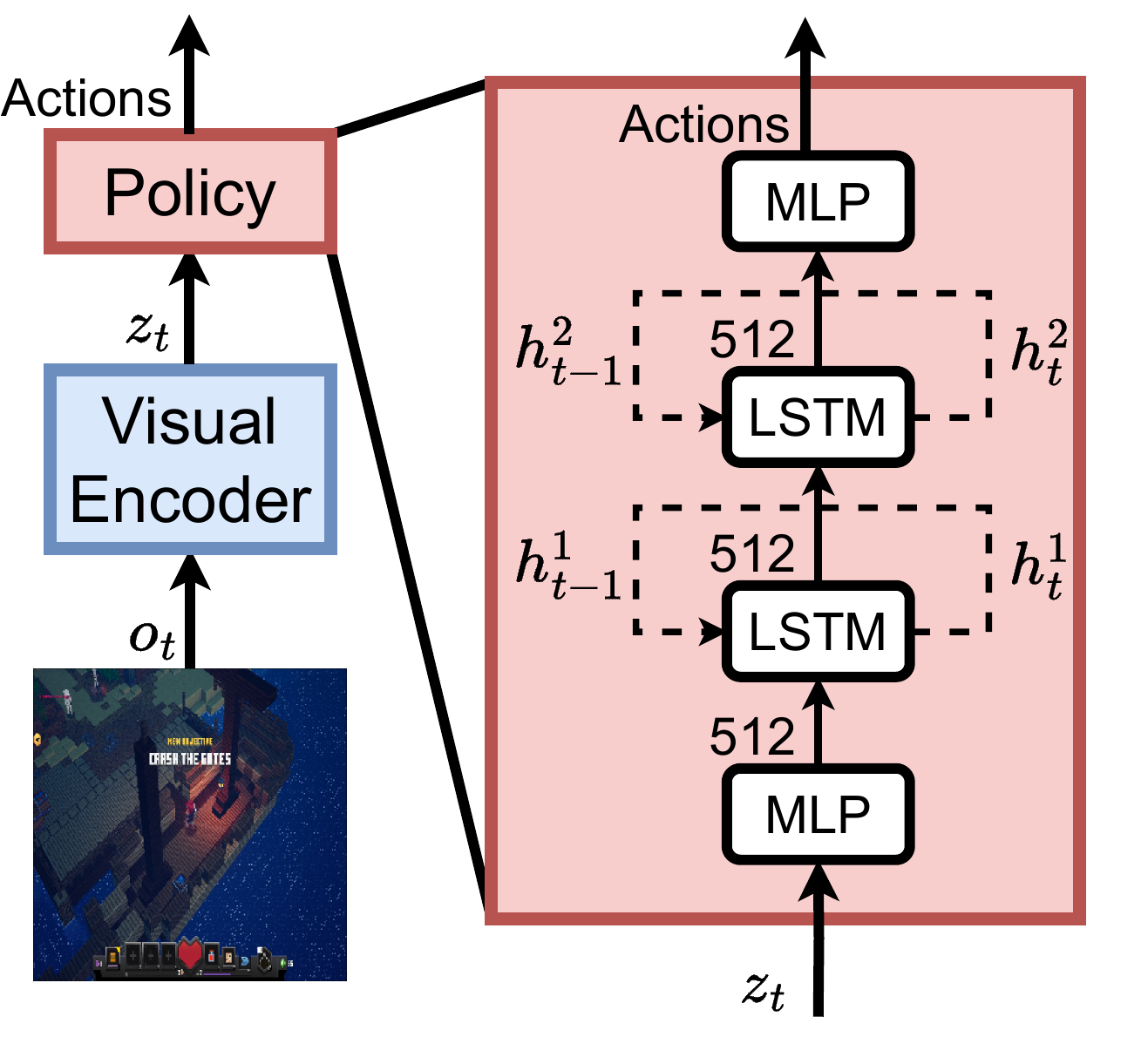}
    \caption{Illustration of the agent architecture. We consider end-to-end trained and pre-trained visual encoders.}
    \label{fig:architecture}

    \vspace{-1em}
\end{figure}

\textbf{Behaviour cloning} (BC) is an imitation learning approach that trains agents through supervised learning on a dataset of demonstrations, denoted as $\data = \{(o_1, a_1), \ldots, (o_N, a_N)\}$, where $N$ represents the total number of samples in the dataset. Each demonstration comprises a sequence of tuples $(o, a)$ of an image observation $o$ and a chosen action $a$. Using this data, a policy $\pol(a \mid o; \theta)$ is trained to mimic the distribution of actions within $\data$ by minimising the loss
\begin{equation}
    \loss(\theta) = \mathbb{E}_{(o, a) \sim \data, \hat{a} \sim \pol(\cdot \mid o; \theta)}\left[l(a, \hat{a})\right]
\end{equation}
where $l$ measures the discrepancy between the "true" action $a$ and the policy's sampled action $\hat{a}$. For continuous and discrete actions, we use the mean-squared error and cross-entropy loss, respectively.

\begin{table*}[t]
    \centering
    \caption{Overview of visual encoders considered in this study including the training category, image sizes, parameter counts and the size of computed embeddings. %
    For pre-trained encoders we only report the size of visual encoder used to embed images (and do not include the parameters of other model components such as language embedding networks of CLIP models).}
    \label{tab:encoder_overview}
    \begin{tabular}{l l c c c}
        \toprule
        Category & Model & Image size & Parameters & Embedding size\\
        \midrule
        \multirow{6}{*}{End-to-end} & Impala ResNet & $128\times128$ & 98K & 7200\\
         & Custom ResNet & $128\times128$ & 585K & 1024 \\
         & Custom ResNet & $256\times256$ & 586K & 1024 \\
         & ViT Tiny & $224\times224$ & 5.5M & 192\\
         & Custom ViT & $128\times128$ & 8.8M & 512 \\
         & Custom ViT & $256\times256$ & 8.9M & 512 \\
        \midrule
        \multirow{3}{*}{\shortstack[l]{Language contrastive\\pre-trained}} & CLIP ResNet50 & $224\times224$ & 38M & 1024\\
         & CLIP ViT-B/16 & $224\times224$ & 86M & 512\\
         & CLIP ViT-L/14 & $224\times224$ & 303M & 768\\
        \midrule
        \multirow{3}{*}{\shortstack[l]{Self-supervised\\pre-trained}} & DINOv2 ViT-S/14 & $224 \times 224$ & 21M & 384\\
         & DINOv2 ViT-B/14 & $224 \times 224$ & 86M & 768\\
         & DINOv2 ViT-L/14 & $224 \times 224$ & 303M & 1024\\
        \midrule
        \multirow{3}{*}{\shortstack[l]{Classification supervised\\ pre-trained}} & FocalNet Large FL4 & $224 \times 224$ & 205M & 1536\\
         & FocalNet XLarge FL4 & $224 \times 224$ & 364M & 2048\\
         & FocalNet Huge FL4 & $224 \times 224$ & 683M & 2816\\
         \midrule
         Reconstruction pre-trained & Stable Diffusion 2.1 VAE & $256 \times 256$ & 34M & 4096\\
        \bottomrule
    \end{tabular}
\end{table*}

\textbf{Image Processing.} Received images, sampled from the dataset during training or directly from the game during evaluation, are first resized to the required image size of the respective visual encoder (see \Cref{tab:encoder_overview})\footnote{For end-to-end encoders, we resize images using linear interpolation. For pre-trained encoders, we use bicubic interpolation to be consistent with the processing pipeline used during training.}. If image augmentation is used for an encoder, images are augmented after resizing using the same augmentations applied by \citet{baker2022vpt}. Finally, image colour values are normalised. \Cref{fig:architecture} illustrates the architecture of agents with an image $o_t$ being processed by the visual encoder to obtain an embedding $z_t$ that is given to the policy to obtain actions $a_t$.

\textbf{Policy network.} For all experiments, the policy architecture consists of a MLP projecting the embedding $z_t$ of the visual encoder to a dimension of 512 before being fed through a two-layered LSTM~\citep{hochreiter1997long} with hidden dimensions of 512. The LSTM processes the projected embedding and a hidden state $h_{t-1}$ which encodes the history of embeddings during a sequence (obtained as a sampled sequence during training or online evaluation). The output of the LSTM is then projected to as many dimensions as there are actions in the task using a MLP with one hidden layer of 512 dimensions. At each intermediate layer, the ReLU activation function is applied.

\textbf{End-to-end visual encoders.} For visual encoders trained end-to-end with the BC loss, we consider three ResNet~\citep{he2016deep} and three vision transformer (ViT)~\citep{dosovitskiy2021vit} architectures. For ResNets, we evaluate the Impala~\citep{espeholt2018impala} architecture as a commonly used visual encoder for decision-making agents. However, it outputs large embeddings for $128\times128$ images so we also evaluate two alternative ResNet architectures based on ConvNeXt~\citep{liu2022convnext} designed for $128\times128$ and $256\times256$ images, respectively. For ViTs, we evaluate the commonly used tiny model architecture proposed by \citet{steiner2022train} which outputs fairly small embeddings. For comparison, we also evaluate two ViT architectures with slightly larger models that output slightly larger embeddings. See \Cref{app:end_to_end_encoders} for full details on all end-to-end visual encoder architectures.\footnote{All appendices are available online at \url{https://arxiv.org/abs/2312.02312}.}

\textbf{Pre-trained visual encoders.} We consider four paradigms of pre-trained visual encoders with representative models being evaluated in our experiments: OpenAI's CLIP~\citep{radford2021clip} as language contrastive pre-trained encoders, DINOv2~\citep{oquab2023dinov2} as self-supervised pre-trained encoders with self-distillation objectives between a teacher and student network, FocalNet~\citep{yang2022focal} trained on ImageNet21K classification as supervised pre-trained encoders, and a variational autoencoder (VAE)~\citep{kingma2013auto} from stable diffusion~\citep{rombach2022stablediffusion} as reconstruction pre-trained encoder. These visual encoders have already been trained on large amounts of real-world images. During all our experiments, we freeze these encoders and only use them to obtain embeddings of images without any fine-tuning or further training. See \Cref{app:pretrained_encoders} for details on the evaluated models.

\textbf{Training details.}
For each network update, we sample 32 random sequences of 100 consecutive image-action pairs within the dataset. Before each training step, the hidden state and cell state of the LSTMs in the policy are reset and the mean BC loss is computed across all sequences with the hidden state accumulating across the 100 samples within a sequence. The Adam optimiser~\citep{kingma2014adam} is used with decoupled weight decay~\citep{loshchilov2019adamw} of $0.01$ and a learning rate of $3 \cdot 10^{-4}$. To stabilise training, gradients are normalised at $1.0$ and we use half precision for all training. In Minecraft Dungeons, we train each model for 1 million gradient updates. In Minecraft and CS:GO, models are trained for 500,000 gradient updates.

\begin{figure*}[t]
    \centering
    \begin{subfigure}{0.325\textwidth}
        \centering
        \includegraphics[width=\textwidth]{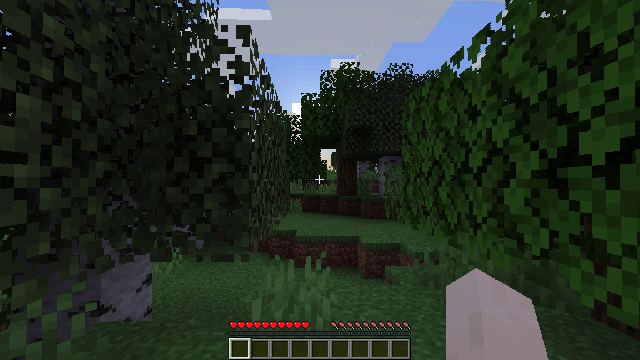}
        \caption{Minecraft}
        \label{fig:minecraft_screenshot}
    \end{subfigure}
        \begin{subfigure}{0.325\textwidth}
        \centering
        \includegraphics[width=\textwidth, trim=15 0 15 0, clip]{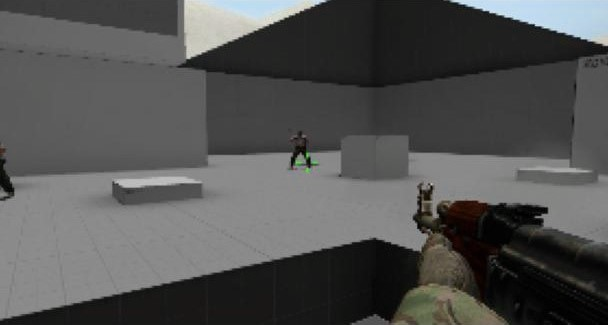}
        \caption{Counter-Strike}
        \label{fig:csgo_screenshot}
    \end{subfigure}
    \begin{subfigure}{0.325\textwidth}
        \centering
        \includegraphics[width=\textwidth]{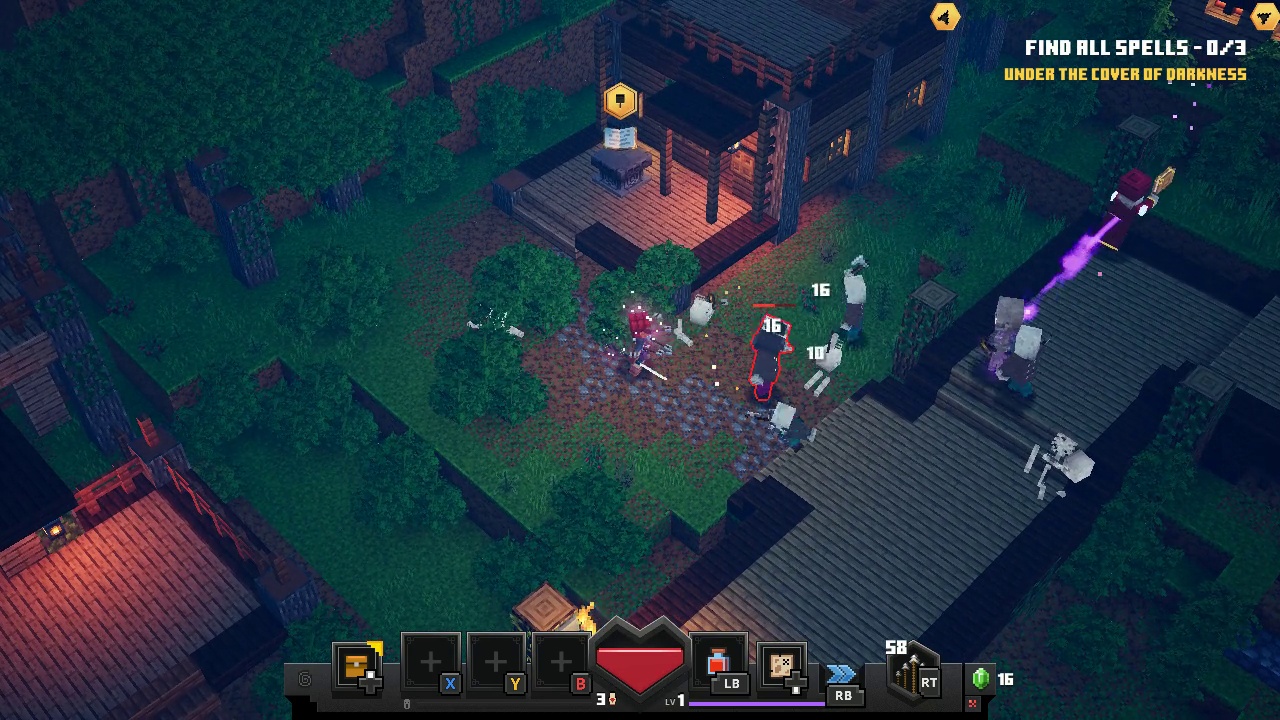}
        \caption{Minecraft Dungeons}
        \label{fig:dungeons_screenshot}
    \end{subfigure}
    \caption{Representative screenshots of all games studied in this paper.}
    \label{fig:game_screenshots}
\end{figure*}

\begin{table*}[t]
    \centering
    \caption{Summary of evaluation video games, %
    the size of the human demonstrations dataset, the perspective (and resolution) of input images, the action space given by the number of continuous (cont.) and discrete binary (bin.) actions, and the task horizon given by the maximum number of time steps per task.}
    \label{tab:env_summary}
        \begin{tabular}{l l l l l l l}
            \toprule
            \textbf{Game} & \textbf{Task} & \textbf{Dataset} & \textbf{Images} & \textbf{Actions} & 
            \textbf{Task horizon}\\
            \midrule
            Minecraft & Treechop & 40min & First-person ($640\times360$) & 2 cont., 8 bin. & \num{1200}\\
            Counter Strike & Clean Aim Train & 45min & First-person ($280\times150$) & 2 cont., 1 bin. & \num{4800}\\
            Minecraft Dungeons & Arch Haven & 8h & Isometric ($1280\times720$) & 4 cont., 11 bin. & \num{3000}\\
            \bottomrule
        \end{tabular}
\end{table*}

\section{Video Game Tasks}
\label{sec:games}
We train and evaluate BC models with all visual encoders in three different games, Minecraft, CS:GO, and Minecraft Dungeons illustrated in \Cref{fig:game_screenshots}. \Cref{tab:env_summary} provides a summary of the tasks and datasets in each of the games. We highlight that (1) all games represent complex and popular video games, (2) the games cover varying image perspectives (isometric and first-person) and varying degrees of realism (as shown in \Cref{fig:game_screenshots}), (3) all games require agents to take long sequences of actions with Minecraft, CS:GO, and Minecraft Dungeons requiring up to \num{1200}, \num{4800}, and \num{3000} actions, respectively, per rollout, and (4) all games have complex combinatorial action spaces including multiple continuous actions (mouse or joystick movement) and discrete binary actions (button presses) with up to fifteen actions in Minecraft Dungeons (as shown in \Cref{tab:env_summary}). Below, we provide a summary of the key features of each of the considered tasks. For full details about each game, including the task, action space, dataset, and online evaluation, we refer to \Cref{app:games}.

In Minecraft, we evaluate in the established MineRL ``Treechop" task~\citep{guss2019minerl} and train on filtered human gameplay demonstrations from the VPT dataset~\citep{baker2022vpt}. We consider any rollout a success if the agent chops a tree within 1 minute of gameplay. Collecting a log from a chopped tree is the first step to craft many of the items in Minecraft, and has been previously used to benchmark reinforcement learning algorithms~\citep{guss2019minerl}. To succeed at this task, the agent needs to navigate a rich and diverse visual environment and coordinate its movement, camera movement, and ``interacting'' action required to chop a tree. Images in Minecraft features highly stylised visuals in the first-person perspective. 

In CS:GO, we evaluate in the ``Clean Aim Train" task and use an existing dataset of human expert demonstrations~\citep{pearce2022counter}. The agent is located in a fixed position and controls the camera to aim and the trigger to shoot. To succeed at the task, the agent needs to look around, identify enemies moving towards it and shoot them. We measure the kills per minute during rollouts as a performance metric. Images in CS:GO are in first-person and have a more realistic style compared to the other two games, so this benchmark tests if our findings generalise across video games with distinct visual styles. We further highlight that images in CS:GO appear more akin to real-world images most pre-trained visual encoders are trained on, but also exhibit few colours with little contrast between objects of interest and the grey background.

Minecraft Dungeons constitutes a novel benchmark for imitation learning which has not been evaluated in before. In this game, we train agents on demonstrations of human gameplay in the ``Arch Haven" level that requires precise navigation through a level with multiple combat scenarios against randomised enemies. In particular the stochastic combat of the game represents a major challenge for imitation learning since the agents will encounter states not seen within the training data, and, thus, have to learn policies that generalise across similar states. Providing the policy with robust representations of the image inputs represents one approach to facilitate such generalisation. Additionally, successful rollouts in this task require the agent to take thousands of accurate steps. Images in Minecraft Dungeons feature a similar style to Minecraft but are always centred on the agent character and have an isometric perspective looking down on the character. These properties make Minecraft Dungeons a highly challenging and suitable evaluation task to study the efficacy of varying visual encoders for imitation learning. For more details on the ``Arch Haven" level, including a 2D visualisation of the level, we refer to \Cref{app:dungeons_archhaven}.

\begin{figure*}[t]
    \centering
    \begin{subfigure}{0.355\textwidth}
        \includegraphics[width=\textwidth]{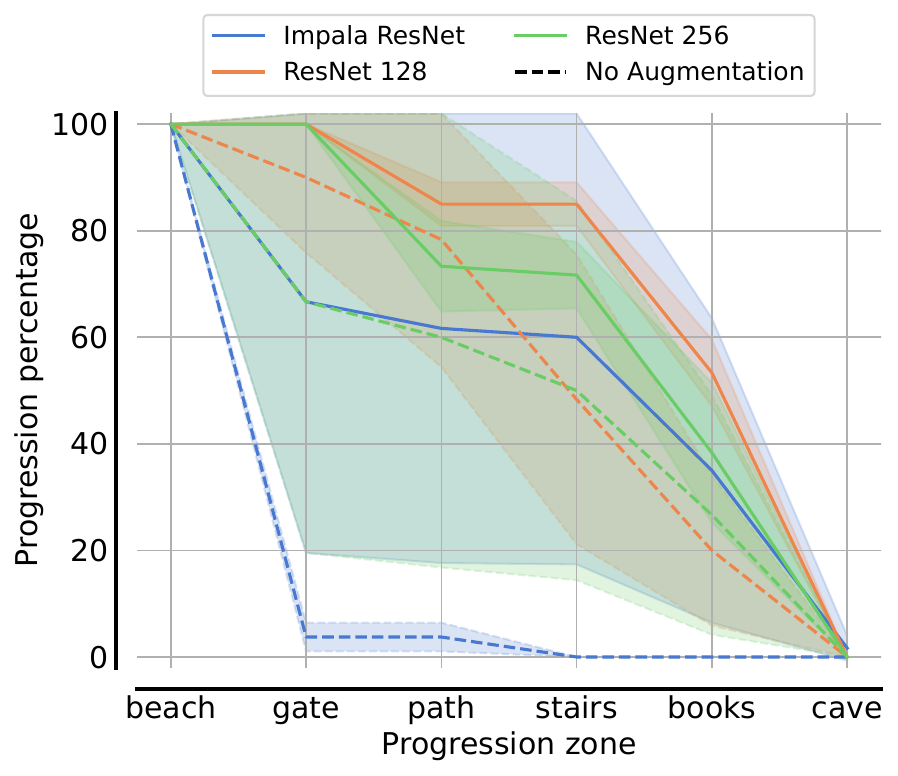}
        \caption{End-to-end ResNets}
        \label{fig:end_to_end_resnets_online_evaluation}
    \end{subfigure}
    \hfill
    \begin{subfigure}{0.31\textwidth}
        \includegraphics[width=\textwidth]{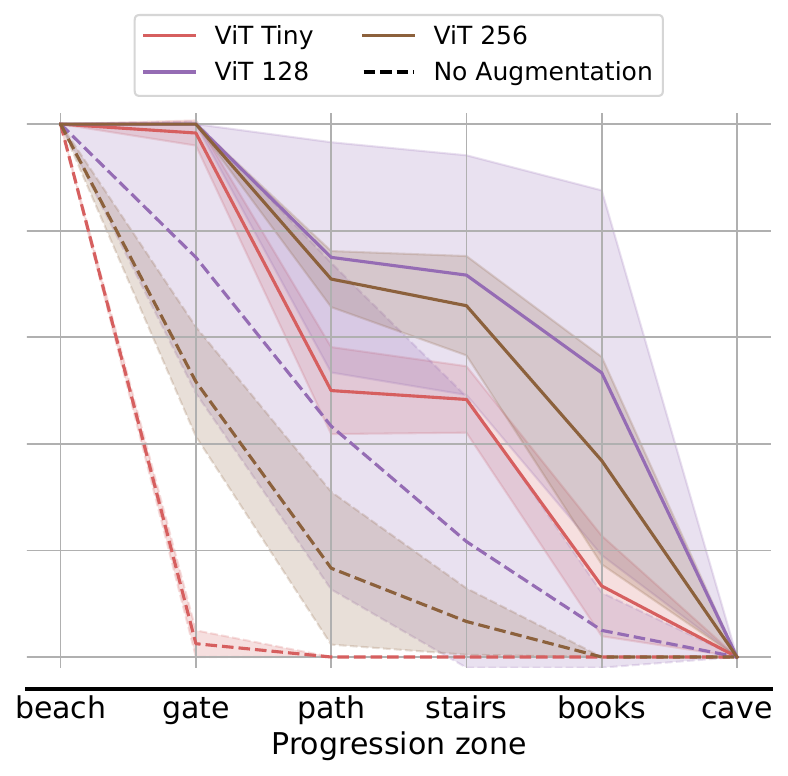}
        \caption{End-to-end ViTs}
        \label{fig:end_to_end_vits_online_evaluation}
    \end{subfigure}
    \hfill
    \begin{subfigure}{0.31\textwidth}
        \includegraphics[width=\textwidth]{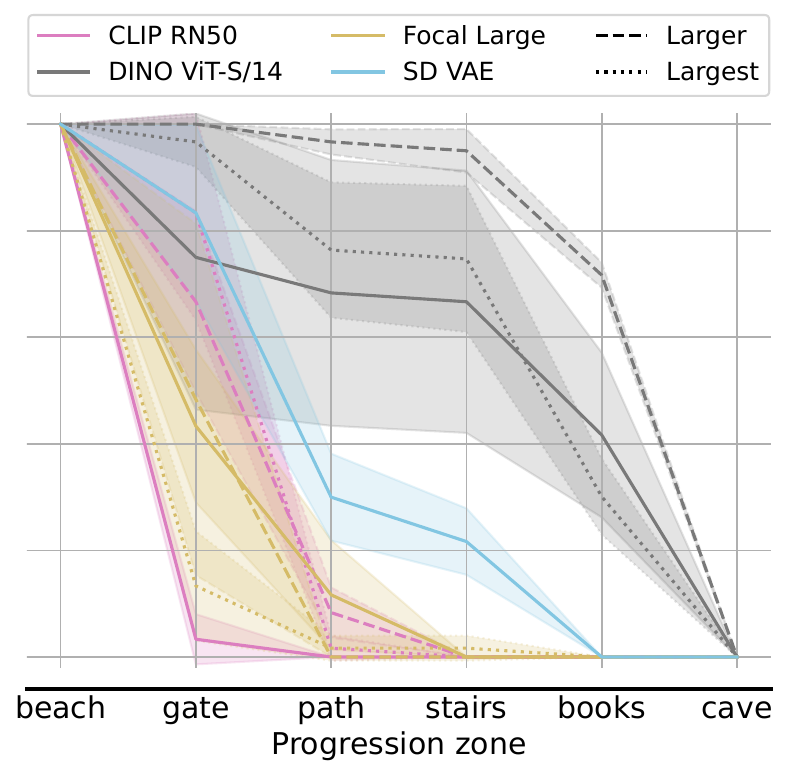}
        \caption{Pre-trained encoders}
        \label{fig:pretrained_online_evaluation}
    \end{subfigure}
    \caption{Online evaluation for BC agents in Minecraft Dungeons with (a) end-to-end ResNets, (b) end-to-end ViTs, and (c) pre-trained visual encoders. We visualise the mean and standard deviation, computed over three training seeds, of the percentage of rollouts progressing to objectives within the ``Arch Haven'' level. Results for the ViT Tiny model are only aggregated over two seeds, as one seed resulted in an invalid checkpoint.
    }
    \label{fig:online_evaluation}
\end{figure*}

\begin{table}[t]
    \centering
    \caption{Online evaluation performance in Minecraft (MC) and CS:GO, measured by the rate of chopping a single tree and kills-per-minute (KPM), respectively. The best end-to-end and pre-trained model in each group is highlighted in bold. We report mean and one standard deviation computed over three training seeds and indicate the number of valid seeds with stars (*) whenever training resulted in invalid checkpoints. For VPT and human baselines, no standard deviation is reported as only one model is available.}
    \label{tab:minecraft_csgo_online_evaluation}
    \small
    \centering
    \begin{tabular}{l S S}
        \toprule
        Model name & \text{MC success (\%)} & \text{CS:GO KPM}\\
        \midrule
        Impala ResNet & 4.00(4.00)\textsuperscript{**} & 6.78(7.06)\\
        ResNet 128 & 12.67(3.86) & 0.38(0.29)\\
        ResNet 256 & 10.00(2.45) & 1.18(1.27)\\
        ViT Tiny & 23.33(4.19) & 0.09(0.14)\\
        ViT 128 & 19.00(2.94) & 2.56(2.08)\\
        ViT 256 & \bf 24.33(0.94) & 0.24(0.62)\\
        \midrule
        Impala ResNet +Aug & 14.00 \ \hspace{-2.7em} $\pm$~0.00\textsuperscript{*} & 11.73(8.42)\\
        ResNet 128 +Aug & 10.00(1.41) & \bf 17.40(2.08)\\
        ResNet 256 +Aug & 6.67(1.70) & 11.40(4.93)\\
        ViT Tiny +Aug & 20.00(5.66) & 6.67(2.25)\\
        ViT 128 +Aug & 20.33(8.06) & 7.29(1.32)\\
        ViT 256 +Aug & 13.67(2.62) & 5.36(1.74)\\
        \midrule
        CLIP ResNet50 & 19.33(8.65) & 5.56(0.84)\\
        CLIP ViT-B/16 & 11.33(1.25) & \bf 6.02(1.01)\\
        CLIP ViT-L/14 & 11.33(3.30) & 2.87(1.85)\\
        DINOv2 ViT-S/14 & 22.33(2.49) & 3.53(1.68)\\
        DINOv2 ViT-B/14 & 25.33(2.05) & 2.98(1.92)\\
        DINOv2 ViT-L/14 & \bf 32.00(1.63) & 5.45(1.34)\\
        FocalNet Large & 16.00(5.66) &  1.93(1.33)\\
        FocalNet XLarge & 15.33(4.03) & 2.44(0.58)\\
        FocalNet Huge & 13.00(1.41) & 1.36(0.87)\\
        Stable Diffusion VAE & 20.00(5.89) & 0.87(0.49)\\
        \midrule
    \end{tabular}
    \begin{tabular}{p{8.6em} w{c}{6.25em} w{c}{5.05em}}
        VPT 71M & $54.00$ & \text{---}\\
        VPT 248M & $55.00$ & \text{---}\\
        VPT 500M & $25.00$ & \text{---}\\
        VPT 248M FT & $48.00$ & \text{---}\\
        VPT 500M FT & $33.00$ & \text{---}\\
        \midrule
        Human (Non-gamer) & \text{---} & $14.32$ \\
        Human (Casual) & \text{---} & $26.21$ \\
        Human (Strong) & \text{---} & $33.21$ \\
        \bottomrule
    \end{tabular}

    \vspace{-.3cm}
\end{table}

\section{Evaluation}
\label{sec:evaluation}
In our evaluation, we focus on the guiding question of how to encode images for data-efficient imitation learning in modern video games. The evaluation is structured in three experiments studying (1) which end-to-end visual encoder is most effective, (2) which pre-trained encoder is most effective, and (3) how do the best end-to-end and pre-trained visual encoders compare under reduced amounts of training data. For each experiment, we train each model with three different seeds and report aggregated training and online evaluation metrics. Lastly, we visually inspect the visual encoders with respect to the information they attend to during action selection, and present the computational cost of training a model with different visual encoders. An outline of the computational resources used for training and evaluation can be found in \Cref{app:hardware_specification}.

To further contextualise the evaluation performance of our models, \Cref{tab:minecraft_csgo_online_evaluation} shows the online evaluation performance of VPT~\citep{baker2022vpt} models in Minecraft Treechop and human performance in the aim assist task of CS:GO, as reported by \citet{pearce2022counter}. VPT models represent transformers models trained on human gameplay in Minecraft. However, these models were trained using behaviour cloning with mangitudes more data and compute compared to our models. In addition to the base models of VPT, we also evaluate VPT models fine-tuned on early-game data (denoted with FT) which, among other tasks, contains significant amounts of demonstrations of players chopping trees as required by the task.

\subsection{End-To-End Visual Encoders}
\label{sec:evaluation_end_to_end}
To identify which end-to-end visual encoder is the most effective, we train all six end-to-end visual encoder architectures (listed in \Cref{tab:encoder_overview}) with and without image augmentations using the BC loss. \Cref{fig:end_to_end_resnets_online_evaluation,fig:end_to_end_vits_online_evaluation} visualise the online evaluation performance for all models with end-to-end ResNet and ViT visual encoders, respectively, in Minecraft Dungeons. For Minecraft and CS:GO, online evaluation performance with end-to-end visual encoders are shown in \Cref{tab:minecraft_csgo_online_evaluation} (left half). Results for Minecraft Dungeons and CS:GO exhibit consistent trends, with image augmentation improving online performance for almost all encoders, ResNet image encoders slightly outperforming ViT models, but not always by significant margins, and image encoders trained on $128\times128$ images performing similar or better than their counterparts with $256\times256$ image inputs. In CS:GO, the best-performing ResNet 128 +Aug as well as the following ResNet 256 +Aug models outperform all other visual encoders but the Impala ResNet + Aug model with high-variance performance by statistically significant margins (double-tailed Welch's test, $p < 0.05$), and exhibits performance comparable to a non-gamer human player. In Minecraft, we also observe that the input image size has no significant effect on the results. However, ViT 256 and ViT Tiny outperform most ResNets by statistically significant margins, and image augmentations harm online evaluation performance in several cases. These results suggest two main findings: (1) small images of $128\times128$ are sufficient to train agents in complex modern video games, and (2) both image augmentation and architecture choice (ResNet or ViT) have the potential to significantly improve performance but are game-specific.

\subsection{Pre-Trained Visual Encoders}
\label{sec:evaluation_pretrained}
To identify which pre-trained visual encoders enable effective decision making in video games, we compare BC agents trained with the representations of 10 pre-trained encoders. The encoders are frozen during training. 

In MineCraft (\Cref{tab:minecraft_csgo_online_evaluation}, right half) and Minecraft Dungeons (\Cref{fig:pretrained_online_evaluation}), we find that BC models with DINOv2 visual encoders generally outperform other models. In Minecraft, the largest DINOv2 ViT-L/14, significantly ($p < 0.05$) outperforms all but the noisiest models (Tiny ViT, ViT 128 +Aug, Stable Diffusion and CLIP ResNet 50) and reaches performance comparable to the largest VPT models despite having magnitudes less parameters and training budget. While smaller DINOv2 models appear better than FocalNet or CLIP, their results are not significantly different from ViT-B/14 and ViT-S/14 DINOv2 models. Stable diffusion VAE appears similarly effective to smaller DINOv2 models in Minecraft. Lastly, we observe that there is no clear correlation between the model size of pre-trained encoders and online performance. While larger DINOv2 models perform best in Minecraft, the same trend does not hold for CLIP and FocalNet where encoders with fewer parameters perform better. In Minecraft Dungeons, the BC models trained with DINOv2 ViT-B/14 pre-trained encoder outperforms all other models, including any end-to-end trained visual encoder. The stable diffusion encoder still outperforms FocalNet and CLIP visual encoders, but performs notably worse than all DINOv2 models.

\looseness=-1
Interestingly, we observe different trends in CS:GO. The CLIP ResNet50 and ViT-B/16 encoders and the largest DINOv2 ViT-L/14 model outperform all other pre-trained encoders by statistically significant margins. However, while the best-performing pre-trained encoders in Minecraft Dungeons and Minecraft outperformed the best-performing end-to-end trained encoders, the opposite holds in CS:GO with pre-trained encoders performing significantly worse than the best-performing end-to-end visual encoders. This trend is consistent across all pre-trained encoders, and might occur due to the 
different visual style of CS:GO compared to the other evaluation games. The ``Clean Aim Train" map in CS:GO is a comparably visually monotone environment with little contrast between background and target enemies, which might lead to representations of pre-trained encoders that lack important information.%

\looseness=-1
Overall, these results suggest that the efficacy of pre-trained visual encoder is game-specific, but DINOv2 represent a strong starting point for BC agents in visually complex video games. These results complement recent findings in robotics~\citep{di2024dinobot} that suggest that DINO~\citep{caron2021emerging} can yield high-quality universal representations for data-efficient imitation learning.

\begin{figure}[t]
    \centering
    \includegraphics[width=0.8\linewidth]{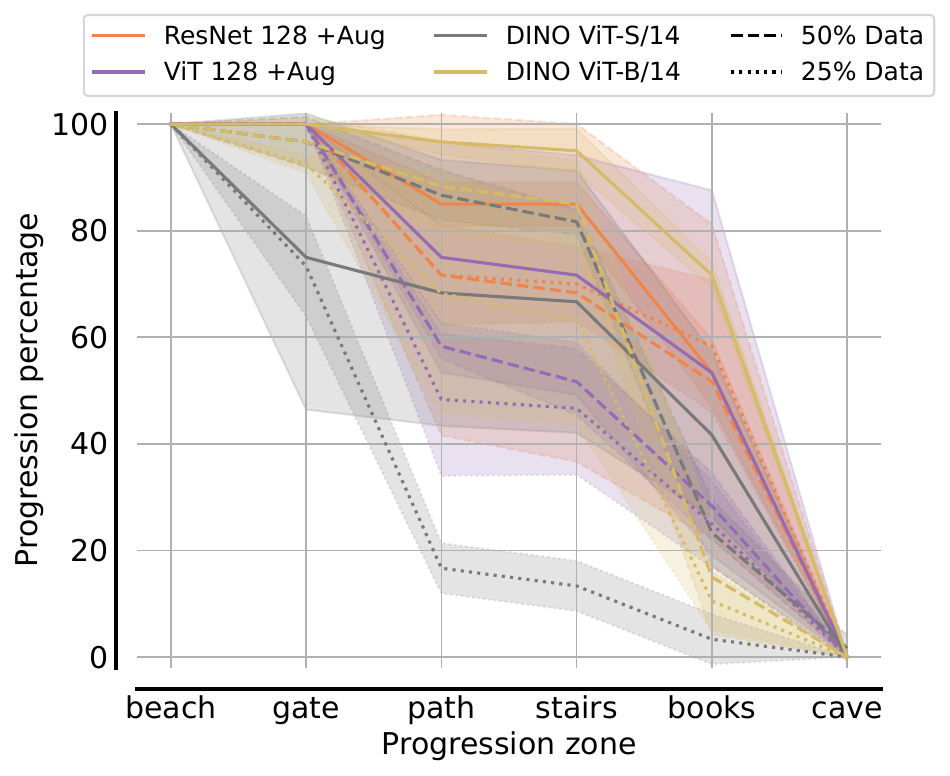}
    \caption{Online evaluation progression for the best-performing BC agents in Minecraft Dungeons with the full dataset (solid line) and subparts of the dataset.}
    \label{fig:reduced_dataset_online_evaluation}
\end{figure}

\begin{table}[t]
    \centering
    \caption{Online evaluation performance for the best-performing BC agents in Minecraft with the full dataset and 10\% of the dataset.}
    \label{tab:minecraft_reduced_dataset_online_evaluation}
    \robustify\bf
    \begin{tabular}{l S}
        \toprule
        Model name & \text{Success rate (\%)}\\
        \midrule
        ViT 256 (Full) & 24.33(0.94)\\
        ViT 256 (10\%) & 10.33(1.70)\\
        \midrule            
        ViT Tiny (Full) & 23.33(4.19)\\
        ViT Tiny (10\%) & 16.50(1.50)\\
        \midrule
        DINOv2 ViT-L/14 (Full) & 32.00(1.63)\\
        DINOv2 ViT-L/14 (10\%) & 15.00(2.16)\\
        \midrule
        DINOv2 ViT-B/14 (Full) & 25.33(2.05)\\
        DINOv2 ViT-B/14 (10\%) & 17.00(1.41)\\
        \bottomrule
    \end{tabular}

    \vspace{-.3cm}
\end{table}

\begin{figure*}[t]
    \centering
    \begin{adjustbox}{valign=t}  
        \begin{minipage}{0.16\linewidth} 
            \centering
            \includegraphics[width=\textwidth,height=\textwidth]{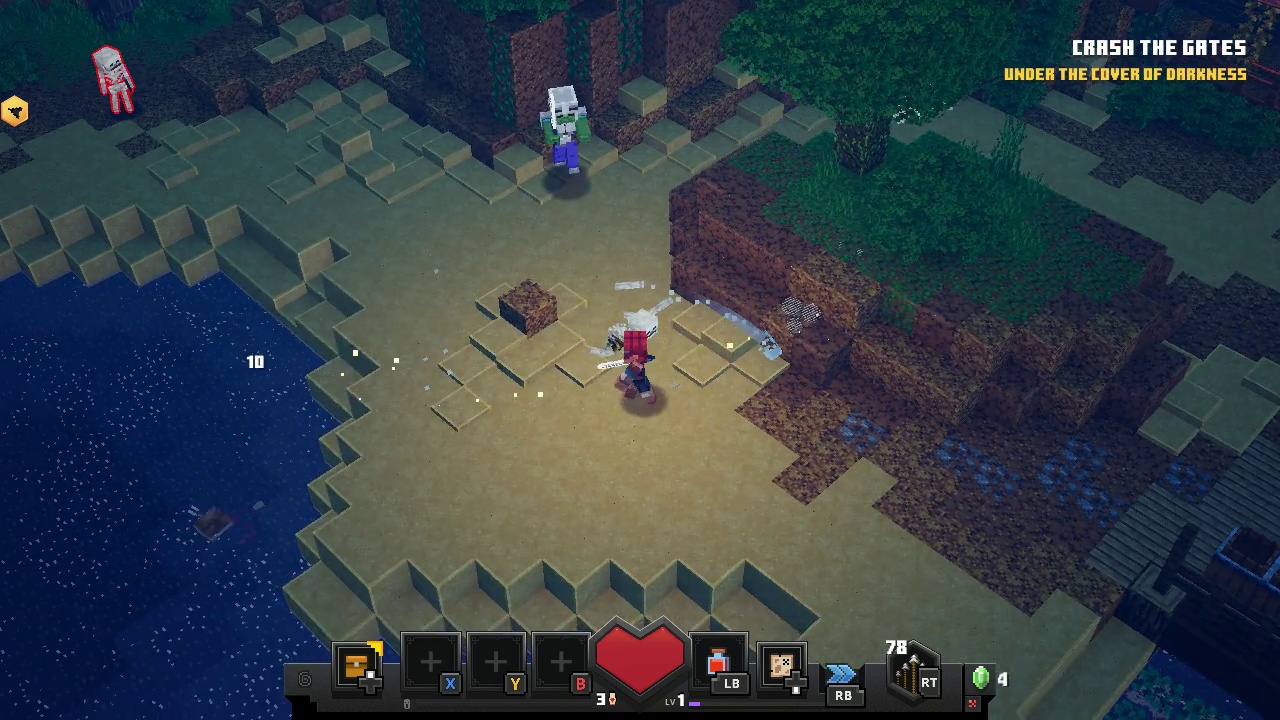}
            \subcaption{Original}
        \end{minipage}
    \end{adjustbox}
     \begin{adjustbox}{valign=t}  
        \begin{minipage}{0.16\linewidth}
            \includegraphics[width=\textwidth,trim={0.5em 0.5em 0.5em 0.5em},clip]{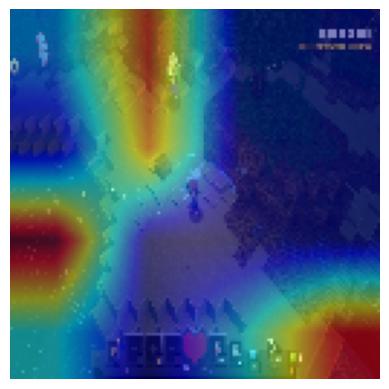}
            \subcaption{RN 128 +Aug}
        \end{minipage}
    \end{adjustbox}
    \begin{adjustbox}{valign=t}  
        \begin{minipage}{0.16\linewidth}
            \includegraphics[width=\textwidth,trim={0.5em 0.5em 0.5em 0.5em},clip]{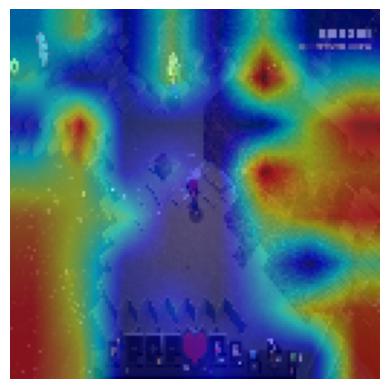}
            \subcaption{ViT 128 +Aug}
        \end{minipage}
    \end{adjustbox}
    \begin{adjustbox}{valign=t}  
        \begin{minipage}{0.16\linewidth}
            \includegraphics[width=\textwidth,trim={0.5em 0.5em 0.5em 0.5em},clip]{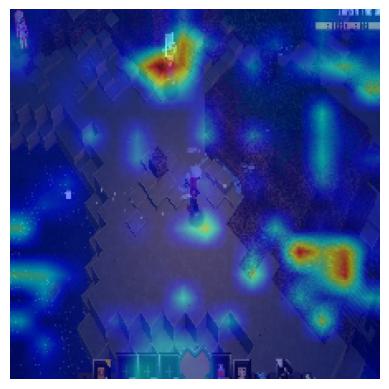}
            \subcaption{DINOv2 ViT-S}
        \end{minipage}
    \end{adjustbox}
    \begin{adjustbox}{valign=t}
        \begin{minipage}{0.16\linewidth}
            \includegraphics[width=\textwidth,trim={0.5em 0.5em 0.5em 0.5em},clip]{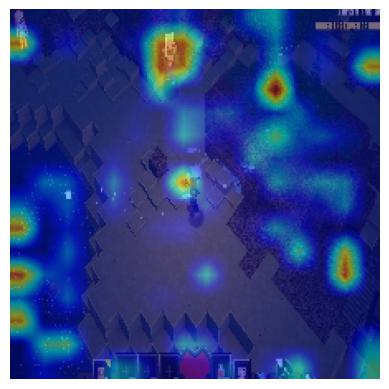}
            \subcaption{DINOv2 ViT-B}
        \end{minipage}
    \end{adjustbox}
    
    \begin{adjustbox}{valign=t}  
        \begin{minipage}{0.16\linewidth} 
            \centering
            \includegraphics[width=\textwidth,height=\textwidth]{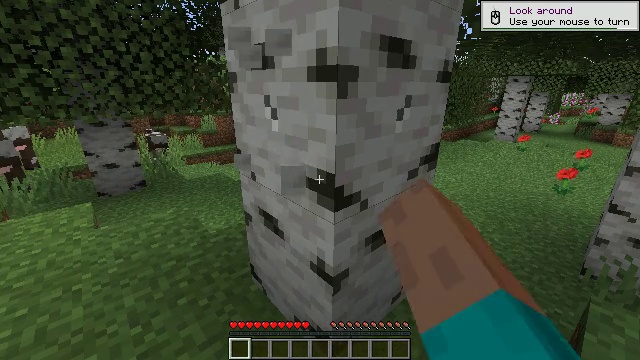}
            \subcaption{Original}
        \end{minipage}
    \end{adjustbox}
     \begin{adjustbox}{valign=t}  
        \begin{minipage}{0.16\linewidth}
            \includegraphics[width=\textwidth,trim={0.5em 0.5em 0.5em 0.5em},clip]{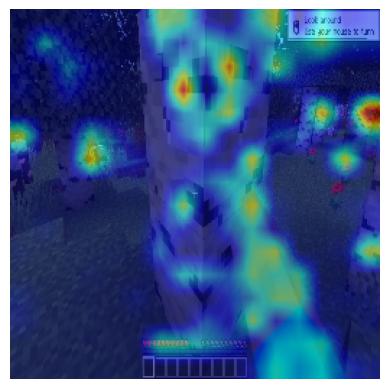}
            \subcaption{ViT 256}
        \end{minipage}
    \end{adjustbox}
    \begin{adjustbox}{valign=t}  
        \begin{minipage}{0.16\linewidth}
            \includegraphics[width=\textwidth,trim={0.5em 0.5em 0.5em 0.5em},clip]{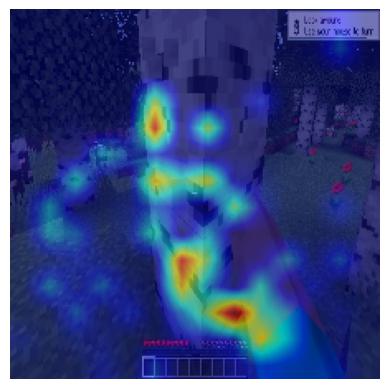}
            \subcaption{ViT Tiny}
        \end{minipage}
    \end{adjustbox}
    \begin{adjustbox}{valign=t}  
        \begin{minipage}{0.16\linewidth}
            \includegraphics[width=\textwidth,trim={0.5em 0.5em 0.5em 0.5em},clip]{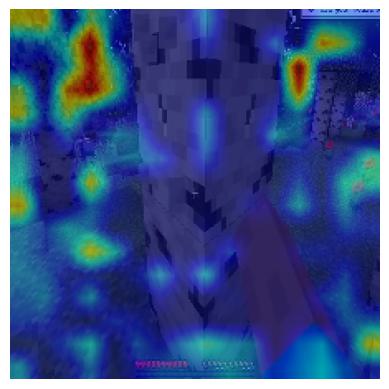}
            \subcaption{DINOv2 ViT-L}
        \end{minipage}
    \end{adjustbox}
    \begin{adjustbox}{valign=t}  
        \begin{minipage}{0.16\linewidth}
            \includegraphics[width=\textwidth,trim={0.5em 0.5em 0.5em 0.5em},clip]{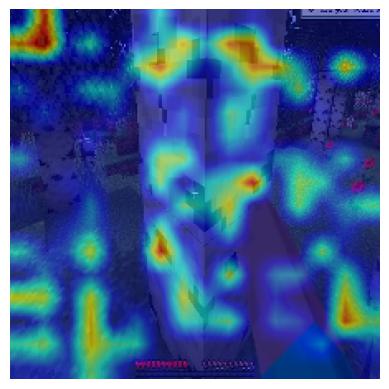}
            \subcaption{DINOv2 ViT-B}
        \end{minipage}
    \end{adjustbox}

    \begin{adjustbox}{valign=t}  
        \begin{minipage}{0.16\linewidth} 
            \centering
            \includegraphics[width=\textwidth,height=\textwidth]{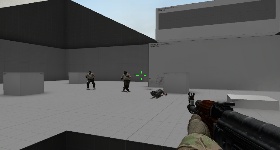}
            \subcaption{Original}
        \end{minipage}
    \end{adjustbox}
     \begin{adjustbox}{valign=t}  
        \begin{minipage}{0.16\linewidth}
            \includegraphics[width=\textwidth,trim={0.5em 0.5em 0.5em 0.5em},clip]{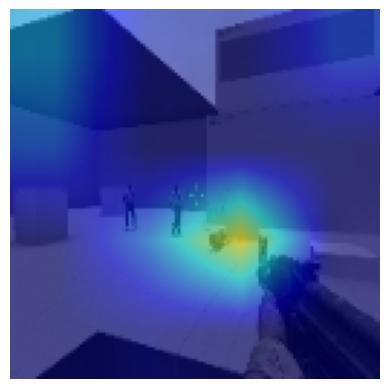}
            \subcaption{RN 128 +Aug}
        \end{minipage}
    \end{adjustbox}
    \begin{adjustbox}{valign=t}  
        \begin{minipage}{0.16\linewidth}
            \includegraphics[width=\textwidth,trim={0.5em 0.5em 0.5em 0.5em},clip]{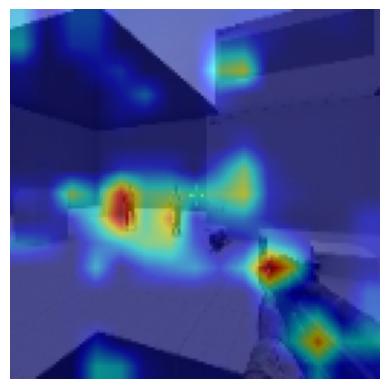}
            \subcaption{Imp RN +Aug}
        \end{minipage}
    \end{adjustbox}
    \begin{adjustbox}{valign=t}  
        \begin{minipage}{0.16\linewidth}
            \includegraphics[width=\textwidth,trim={0.5em 0.5em 0.5em 0.5em},clip]{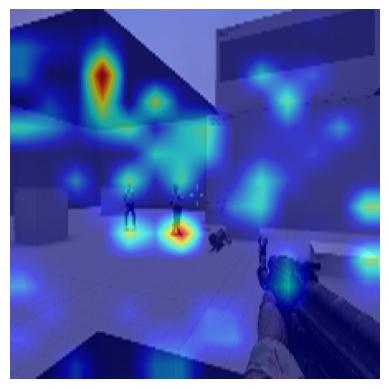}
            \subcaption{CLIP ViT-B}
        \end{minipage}
    \end{adjustbox}
    \begin{adjustbox}{valign=t}  
        \begin{minipage}{0.16\linewidth}
            \includegraphics[width=\textwidth,trim={0.5em 0.5em 0.5em 0.5em},clip]{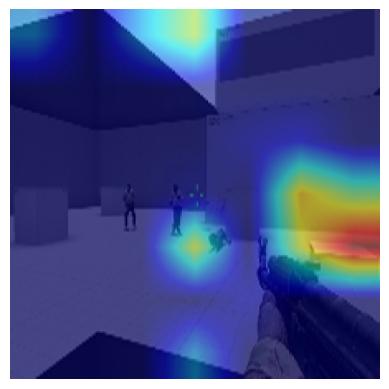}
            \subcaption{CLIP RN50}
        \end{minipage}
    \end{adjustbox}

    \caption{Grad-CAM visualisation of the activation of the best-performing visual encoders for Minecraft Dungeons (top), Minecraft (middle), and CS:GO (bottom) with action logits of a BC policy serving as targets. Red areas represent the parts of the image the visual encoders focus on the most.}
    \label{fig:grad_cam_highlights}
\end{figure*}

\subsection{How Much Data Do You Need?}
\label{sec:data_reduction}
A significant advantage of utilising pre-trained visual encoders is their independence from additional training, potentially resulting in more reliable performance with limited data. In contrast, visual encoders specifically trained for a particular task may be less generalisable but have the potential to outperform general-purpose pre-trained encoders. To test this hypothesis, we examine how the top-performing end-to-end and pre-trained visual encoders (based on online evaluation performance) compare when trained on reduced datasets in Minecraft and Minecraft Dungeons.

In Minecraft, we select the DINOv2 ViT-B/14 and ViT-L/14 models, and the ViT 256 and ViT Tiny models as best-performing pre-trained and end-to-end encoders and train them on a reduced dataset. The reduced dataset contains only 10\% ($\sim$ 3.5 minutes, 14 demonstrations, 4269 total steps of data) of the original gameplay data sampled uniformly at random. \Cref{tab:minecraft_reduced_dataset_online_evaluation} shows the rollout success rate of all models trained on the full and 10\% datasets. For the reduced dataset, the success rate of all models drops by half, but they still achieve reasonable success rate and perform better than some encoders trained on the full dataset. This is surprising considering how little data these models have been trained on and how visually diverse the environment in Minecraft can be across different rollouts. Perhaps surprisingly and contrary to our initial hypothesis, models with pre-trained visual encoders are similarly affected by the reduction in training data than the end-to-end visual encoders. This finding suggests that both pre-trained and end-to-end trained encoders can work reasonably well with less than 5 minutes of high-quality demonstration data.  %

\begin{table*}[t]
    \centering
    \caption{Training cost in training time (minutes) and VRAM (GB) usage for \num{2500} training steps on Minecraft Treechop training dataset for end-to-end trained (left) and pre-trained (right) visual encoders. Numbers are the average across three runs, measured on a machine with a single A100 GPU (80GB VRAM). Runs marked with $\dagger$ ran out of VRAM.}
    \label{tab:training_cost}

    \begin{minipage}[t]{.39\textwidth}
        \begin{tabular}{l c c}
             \toprule
             Encoder & \text{Time (min)} & \text{VRAM (GB)} \\
             \midrule
             Impala ResNet & \num{17.84} & \num{21.37} \\
             ResNet 128 & \num{10.58} & \num{5.74} \\
             ResNet 256 & \num{19.72} & \num{14.30} \\
             ViT Tiny & $\dagger$ & $\dagger$ \\
             ViT 128 & \num{16.38} & \num{21.20} \\
             ViT 256 & $\dagger$ & $\dagger$ \\
             \midrule
             Impala ResNet + Aug & \num{23.23} & \num{20.01} \\
             ResNet 128 + Aug & \num{15.73} & \num{17.97} \\
             ResNet 256 + Aug & $\dagger$ & $\dagger$ \\
             ViT Tiny + Aug & $\dagger$ & $\dagger$ \\
             ViT 128 + Aug & \num{21.85} & \num{26.88} \\
             ViT 256 + Aug & $\dagger$ & $\dagger$ \\
             \bottomrule
        \end{tabular}
    \end{minipage}
    \begin{minipage}[t]{.59\textwidth}
        \begin{tabular}{l c c c c}
             \toprule
             & \multicolumn{2}{c}{\text{No pre-computation}} & \multicolumn{2}{c}{\text{Pre-computed embeddings}}\\
             Encoder & \text{Time (min)} & \text{VRAM (GB)} & \text{Time (min)} & \text{VRAM (GB)}\\
             \midrule
             CLIP ResNet50 & \num{43.69} & \num{30.62} & \num{4.12} & \num{1.44}\\
             CLIP ViT-B/16 & \num{62.38} & \num{29.72} & \num{3.35} & \num{1.53}\\
             CLIP ViT-L/14 & \num{227.47} & \num{49.21} & \num{4.47} & \num{2.12}\\
             \midrule
             DINOv2 ViT-S/14 & \num{49.12} & \num{21.36} & \num{3.90} & \num{1.27}\\
             DINOv2 ViT-B/14 & \num{99.90} & \num{41.65} & \num{3.34} & \num{1.52}\\
             DINOv2 ViT-L/14 & \num{280.48} & \num{54.38} & \num{4.58} & \num{2.47}\\
             \midrule
             FocalNet Large & $\dagger$ & $\dagger$ & \num{4.43} & \num{2.19}\\
             FocalNet XLarge & $\dagger$ & $\dagger$ & \num{4.10} & \num{2.80}\\
             FocalNet Huge & $\dagger$ & $\dagger$ & \num{6.45} & \num{4.07}\\
             \midrule
             Stable Diffusion VAE & $\dagger$ & $\dagger$ & \num{4.92} & \num{1.46}\\
             \bottomrule
        \end{tabular}
    \end{minipage}
\end{table*}

In Minecraft Dungeons, we select the DINOv2 ViT-S/14, ViT-B/14 models, as well as the ResNet and ViT architectures on $128\times128$ images with image augmentation as the best-performing pre-trained and end-to-end trained encoders, respectively. We generate two reduced datasets with 50\% ($\sim 4$ hours) and 25\% ($\sim 2$ hours) of the training data by sampling trajectories uniformly at random. Each of the selected models is then trained on the 50\% and 25\% training datasets for 500 and 250 thousand gradient updates, respectively. \Cref{fig:reduced_dataset_online_evaluation} shows the online evaluation performance of all models. As expected, we can see that the performance of all models gradually deteriorates as the training data is reduced. For pre-trained models, the larger DINOv2 ViT-B/14 outperforms the smaller ViT-S/14 when dealing with smaller datasets. %
Regarding end-to-end trained models, the ViT model's performance declines more rapidly with smaller data quantities compared to the ResNet. However, similarly to results in Minecraft, both end-to-end trained visual encoders yield performance comparable to pre-trained models in lower data regimes. These findings might indicate that the rollout performance of models in this low-data regime is mostly bottlenecked by the policy rather than the visual encoder.

\subsection{Grad-Cam Inspection of Visual Encoders}
To understand what information is captured by the trained visual encoders, we use gradient-weighted class activation mappings (Grad-CAM)~\citep{selvaraju2017grad} to inspect each visual encoder. We visualise the Grad-CAM activations of visual encoders for images in all three games with action logits of trained BC policies serving as the targets. The resulting heatmap visualisations on top of game images can be interpreted as which parts of the image are most relevant for the visual encoder during action selection.

\Cref{fig:grad_cam_highlights} shows the Grad-CAM activations for the best-performing visual encoders in all three games. In Minecraft, most visual encoders tend to focus on parts indicative of nearby terrain, wood, and the progress of chopping a tree, which aligns with the task objective. Similarly, in CS:GO, we observe that many visual encoders focus on areas around enemies and close to edges of obstacles behind which enemies might hide and appear from. However, these correlations are less clear for most of the pre-trained encoders. In Minecraft Dungeons, many visual encoders tend to focus on image segments containing the player character and enemy units. We hypothesise that other activations might correspond to way points the models focus on to navigate through the level. These observations suggest that both end-to-end trained and pre-trained visual encoders capture relevant information for the training tasks, but we emphasise that visualisations vary greatly between images and visual encoders, which makes it difficult to draw general conclusions from these visualisations. For more details on the Grad-CAM visualisations, plots for more game screenshots in both games and all visual encoders, see \Cref{app:grad_cam}.

\subsection{Computational Budget for Training}
\label{sec:training_cost}
Pre-trained visual encoders typically represent comparably large models, leading to costly training and inference of large batches of images, as required during training with frozen pre-trained encoders. To reduce this computational cost, we pre-compute embeddings for all images in our training dataset. We note that this pre-computation is only possible because we freeze the parameters of pre-trained visual encoders during training and we have comparably small datasets. After this pre-computation, we can train the BC policy on the pre-computed embeddings without requiring the pre-trained encoder. This process significantly reduces the cost of training in both time and memory requirements compared to both training with pre-trained encoders without pre-computed embeddings and training with end-to-end visual encoders.

To quantify the computational cost of training with different visual encoders, \Cref{tab:training_cost} presents the training time and VRAM usage of training a BC agent with all visual encoders on the Minecraft dataset. For pre-trained encoders, we present both the computational cost when training with or without pre-computed embeddings. We note that pre-computing embeddings is not possible for end-to-end trained encoders since the encoder parameters change throughout training. As expected, training with pre-computed embeddings is significantly faster and requires less memory than training with the pre-trained encoder. For example for DINOv2 ViT-L/14, training time and VRAM usage is reduced by more than 98\% and 95\%, respectively, compared to computing embeddings during training. With pre-computed embeddings, training with pre-trained encoders is even faster and requires notably less memory than training with end-to-end visual encoders.

\section{Conclusion}
In this study, we systematically evaluated the effectiveness of imitation learning in modern video games by comparing the conventional end-to-end training of task-specific visual encoders with the use of publicly available pre-trained encoders. Our findings revealed that training visual encoders end-to-end on relatively small images can yield strong performance when using high-quality human demonstrations, even in low-data regimes of just a few hours or minutes. DINOv2, trained with self-supervised objectives on diverse data, outperformed other pre-trained visual encoders in two out of three games, indicating its generality and suitability for video games. However, the efficacy of pre-trained encoders varied across games, with all pre-trained encoders performing worse than end-to-end trained encoders in CS:GO. Lastly, we highlighted the significant reduction in computation cost when pre-computing the embeddings with pre-trained encoders. Combined with their strong performance in the majority of games, this suggests that pre-trained encoders are an accessible and strong starting point for training agents in video games.

In order to maintain focus, our study concentrated on a specific visual encoders that allowed for a range of comparisons across architectures and pre-training paradigms. Nevertheless, our study could be complemented by exploring additional supervised-trained pre-trained encoder architectures and additional scenarios within the examined or other video games. Although our study focused on settings with available training data for the evaluation task, future work could explore the potential benefits of pre-trained visual encoders when agents need to generalise across diverse levels or maps with variable visuals.

\bibliographystyle{ACM-Reference-Format} 
\bibliography{references}

%%% -*-BibTeX-*-
%%% Do NOT edit. File created by BibTeX with style
%%% ACM-Reference-Format-Journals [18-Jan-2012].

\begin{thebibliography}{41}

%%% ====================================================================
%%% NOTE TO THE USER: you can override these defaults by providing
%%% customized versions of any of these macros before the \bibliography
%%% command.  Each of them MUST provide its own final punctuation,
%%% except for \shownote{}, \showDOI{}, and \showURL{}.  The latter two
%%% do not use final punctuation, in order to avoid confusing it with
%%% the Web address.
%%%
%%% To suppress output of a particular field, define its macro to expand
%%% to an empty string, or better, \unskip, like this:
%%%
%%% \newcommand{\showDOI}[1]{\unskip}   % LaTeX syntax
%%%
%%% \def \showDOI #1{\unskip}           % plain TeX syntax
%%%
%%% ====================================================================

\ifx \showCODEN    \undefined \def \showCODEN     #1{\unskip}     \fi
\ifx \showDOI      \undefined \def \showDOI       #1{#1}\fi
\ifx \showISBNx    \undefined \def \showISBNx     #1{\unskip}     \fi
\ifx \showISBNxiii \undefined \def \showISBNxiii  #1{\unskip}     \fi
\ifx \showISSN     \undefined \def \showISSN      #1{\unskip}     \fi
\ifx \showLCCN     \undefined \def \showLCCN      #1{\unskip}     \fi
\ifx \shownote     \undefined \def \shownote      #1{#1}          \fi
\ifx \showarticletitle \undefined \def \showarticletitle #1{#1}   \fi
\ifx \showURL      \undefined \def \showURL       {\relax}        \fi
% The following commands are used for tagged output and should be
% invisible to TeX
\providecommand\bibfield[2]{#2}
\providecommand\bibinfo[2]{#2}
\providecommand\natexlab[1]{#1}
\providecommand\showeprint[2][]{arXiv:#2}

\bibitem[\protect\citeauthoryear{Baker, Akkaya, Zhokov, Huizinga, Tang,
  Ecoffet, Houghton, Sampedro, and Clune}{Baker et~al\mbox{.}}{2022}]%
        {baker2022vpt}
\bibfield{author}{\bibinfo{person}{Bowen Baker}, \bibinfo{person}{Ilge Akkaya},
  \bibinfo{person}{Peter Zhokov}, \bibinfo{person}{Joost Huizinga},
  \bibinfo{person}{Jie Tang}, \bibinfo{person}{Adrien Ecoffet},
  \bibinfo{person}{Brandon Houghton}, \bibinfo{person}{Raul Sampedro}, {and}
  \bibinfo{person}{Jeff Clune}.} \bibinfo{year}{2022}\natexlab{}.
\newblock \showarticletitle{Video pretraining (vpt): Learning to act by
  watching unlabeled online videos}. In \bibinfo{booktitle}{\emph{Advances in
  Neural Information Processing Systems}}.
\newblock


\bibitem[\protect\citeauthoryear{Berner, Brockman, Chan, Cheung, Debiak,
  Dennison, Farhi, Fischer, Hashme, Hesse, et~al\mbox{.}}{Berner
  et~al\mbox{.}}{2019}]%
        {berner2019dota}
\bibfield{author}{\bibinfo{person}{Christopher Berner}, \bibinfo{person}{Greg
  Brockman}, \bibinfo{person}{Brooke Chan}, \bibinfo{person}{Vicki Cheung},
  \bibinfo{person}{Przemyslaw Debiak}, \bibinfo{person}{Christy Dennison},
  \bibinfo{person}{David Farhi}, \bibinfo{person}{Quirin Fischer},
  \bibinfo{person}{Shariq Hashme}, \bibinfo{person}{Chris Hesse},
  {et~al\mbox{.}}} \bibinfo{year}{2019}\natexlab{}.
\newblock \showarticletitle{Dota 2 with large scale deep reinforcement
  learning}.
\newblock \bibinfo{journal}{\emph{ArXiv}}  \bibinfo{volume}{abs/1912.06680}
  (\bibinfo{year}{2019}).
\newblock


\bibitem[\protect\citeauthoryear{Cai, Wang, Ma, Liu, and Liang}{Cai
  et~al\mbox{.}}{2023a}]%
        {cai2023open}
\bibfield{author}{\bibinfo{person}{Shaofei Cai}, \bibinfo{person}{Zihao Wang},
  \bibinfo{person}{Xiaojian Ma}, \bibinfo{person}{Anji Liu}, {and}
  \bibinfo{person}{Yitao Liang}.} \bibinfo{year}{2023}\natexlab{a}.
\newblock \showarticletitle{Open-world multi-task control through goal-aware
  representation learning and adaptive horizon prediction}. In
  \bibinfo{booktitle}{\emph{Proceedings of the Conference on Computer Vision
  and Pattern Recognition}}. \bibinfo{pages}{13734--13744}.
\newblock


\bibitem[\protect\citeauthoryear{Cai, Zhang, Wang, Ma, Liu, and Liang}{Cai
  et~al\mbox{.}}{2023b}]%
        {cai2023groot}
\bibfield{author}{\bibinfo{person}{Shaofei Cai}, \bibinfo{person}{Bowei Zhang},
  \bibinfo{person}{Zihao Wang}, \bibinfo{person}{Xiaojian Ma},
  \bibinfo{person}{Anji Liu}, {and} \bibinfo{person}{Yitao Liang}.}
  \bibinfo{year}{2023}\natexlab{b}.
\newblock \showarticletitle{GROOT: Learning to Follow Instructions by Watching
  Gameplay Videos}.
\newblock \bibinfo{journal}{\emph{ArXiv}}  \bibinfo{volume}{abs/2310.08235}
  (\bibinfo{year}{2023}).
\newblock


\bibitem[\protect\citeauthoryear{Caron, Touvron, Misra, J\'egou, Mairal,
  Bojanowski, and Joulin}{Caron et~al\mbox{.}}{2021}]%
        {caron2021emerging}
\bibfield{author}{\bibinfo{person}{Mathilde Caron}, \bibinfo{person}{Hugo
  Touvron}, \bibinfo{person}{Ishan Misra}, \bibinfo{person}{Herv\'e J\'egou},
  \bibinfo{person}{Julien Mairal}, \bibinfo{person}{Piotr Bojanowski}, {and}
  \bibinfo{person}{Armand Joulin}.} \bibinfo{year}{2021}\natexlab{}.
\newblock \showarticletitle{Emerging Properties in Self-Supervised Vision
  Transformers}. In \bibinfo{booktitle}{\emph{Proceedings of the International
  Conference on Computer Vision}}. \bibinfo{pages}{9650--9660}.
\newblock


\bibitem[\protect\citeauthoryear{Di~Palo and Johns}{Di~Palo and Johns}{2024}]%
        {di2024dinobot}
\bibfield{author}{\bibinfo{person}{Norman Di~Palo} {and}
  \bibinfo{person}{Edward Johns}.} \bibinfo{year}{2024}\natexlab{}.
\newblock \showarticletitle{DINOBot: Robot Manipulation via Retrieval and
  Alignment with Vision Foundation Models}.
\newblock \bibinfo{journal}{\emph{ArXiv}}  \bibinfo{volume}{abs/2402.13181}
  (\bibinfo{year}{2024}).
\newblock


\bibitem[\protect\citeauthoryear{Dosovitskiy, Beyer, Kolesnikov, Weissenborn,
  Zhai, Unterthiner, Dehghani, Minderer, Heigold, Gelly,
  et~al\mbox{.}}{Dosovitskiy et~al\mbox{.}}{2021}]%
        {dosovitskiy2021vit}
\bibfield{author}{\bibinfo{person}{Alexey Dosovitskiy}, \bibinfo{person}{Lucas
  Beyer}, \bibinfo{person}{Alexander Kolesnikov}, \bibinfo{person}{Dirk
  Weissenborn}, \bibinfo{person}{Xiaohua Zhai}, \bibinfo{person}{Thomas
  Unterthiner}, \bibinfo{person}{Mostafa Dehghani}, \bibinfo{person}{Matthias
  Minderer}, \bibinfo{person}{Georg Heigold}, \bibinfo{person}{Sylvain Gelly},
  {et~al\mbox{.}}} \bibinfo{year}{2021}\natexlab{}.
\newblock \showarticletitle{An image is worth 16x16 words: Transformers for
  image recognition at scale}. In \bibinfo{booktitle}{\emph{International
  Conference on Learning Representations}}.
\newblock


\bibitem[\protect\citeauthoryear{Espeholt, Soyer, Munos, Simonyan, Mnih, Ward,
  Doron, Firoiu, Harley, Dunning, et~al\mbox{.}}{Espeholt
  et~al\mbox{.}}{2018}]%
        {espeholt2018impala}
\bibfield{author}{\bibinfo{person}{Lasse Espeholt}, \bibinfo{person}{Hubert
  Soyer}, \bibinfo{person}{Remi Munos}, \bibinfo{person}{Karen Simonyan},
  \bibinfo{person}{Vlad Mnih}, \bibinfo{person}{Tom Ward},
  \bibinfo{person}{Yotam Doron}, \bibinfo{person}{Vlad Firoiu},
  \bibinfo{person}{Tim Harley}, \bibinfo{person}{Iain Dunning},
  {et~al\mbox{.}}} \bibinfo{year}{2018}\natexlab{}.
\newblock \showarticletitle{Impala: Scalable distributed deep-rl with
  importance weighted actor-learner architectures}. In
  \bibinfo{booktitle}{\emph{International conference on machine learning}}.
  PMLR, \bibinfo{pages}{1407--1416}.
\newblock


\bibitem[\protect\citeauthoryear{Fan, Wang, Jiang, Mandlekar, Yang, Zhu, Tang,
  Huang, Zhu, and Anandkumar}{Fan et~al\mbox{.}}{2022}]%
        {fan2022minedojo}
\bibfield{author}{\bibinfo{person}{Linxi Fan}, \bibinfo{person}{Guanzhi Wang},
  \bibinfo{person}{Yunfan Jiang}, \bibinfo{person}{Ajay Mandlekar},
  \bibinfo{person}{Yuncong Yang}, \bibinfo{person}{Haoyi Zhu},
  \bibinfo{person}{Andrew Tang}, \bibinfo{person}{De-An Huang},
  \bibinfo{person}{Yuke Zhu}, {and} \bibinfo{person}{Anima Anandkumar}.}
  \bibinfo{year}{2022}\natexlab{}.
\newblock \showarticletitle{Minedojo: Building open-ended embodied agents with
  internet-scale knowledge}. In \bibinfo{booktitle}{\emph{Advances in Neural
  Information Processing Systems}}.
\newblock


\bibitem[\protect\citeauthoryear{Gillberg, Bergdahl, Sestini, Eakins, and
  Gisslen}{Gillberg et~al\mbox{.}}{2023}]%
        {gillberg2023technical}
\bibfield{author}{\bibinfo{person}{Jonas Gillberg}, \bibinfo{person}{Joakim
  Bergdahl}, \bibinfo{person}{Alessandro Sestini}, \bibinfo{person}{Andrew
  Eakins}, {and} \bibinfo{person}{Linus Gisslen}.}
  \bibinfo{year}{2023}\natexlab{}.
\newblock \showarticletitle{Technical Challenges of Deploying Reinforcement
  Learning Agents for Game Testing in AAA Games}.
\newblock \bibinfo{journal}{\emph{ArXiv}}  \bibinfo{volume}{abs/2307.11105}
  (\bibinfo{year}{2023}).
\newblock


\bibitem[\protect\citeauthoryear{Guss, Houghton, Topin, Wang, Codel, Veloso,
  and Salakhutdinov}{Guss et~al\mbox{.}}{2019}]%
        {guss2019minerl}
\bibfield{author}{\bibinfo{person}{William~H Guss}, \bibinfo{person}{Brandon
  Houghton}, \bibinfo{person}{Nicholay Topin}, \bibinfo{person}{Phillip Wang},
  \bibinfo{person}{Cayden Codel}, \bibinfo{person}{Manuela Veloso}, {and}
  \bibinfo{person}{Ruslan Salakhutdinov}.} \bibinfo{year}{2019}\natexlab{}.
\newblock \showarticletitle{Minerl: A large-scale dataset of minecraft
  demonstrations}.
\newblock \bibinfo{journal}{\emph{ArXiv}}  \bibinfo{volume}{abs/1907.13440}
  (\bibinfo{year}{2019}).
\newblock


\bibitem[\protect\citeauthoryear{He, Zhang, Ren, and Sun}{He
  et~al\mbox{.}}{2016}]%
        {he2016deep}
\bibfield{author}{\bibinfo{person}{Kaiming He}, \bibinfo{person}{Xiangyu
  Zhang}, \bibinfo{person}{Shaoqing Ren}, {and} \bibinfo{person}{Jian Sun}.}
  \bibinfo{year}{2016}\natexlab{}.
\newblock \showarticletitle{Deep residual learning for image recognition}. In
  \bibinfo{booktitle}{\emph{IEEE conference on computer vision and pattern
  recognition}}.
\newblock


\bibitem[\protect\citeauthoryear{Hochreiter and Schmidhuber}{Hochreiter and
  Schmidhuber}{1997}]%
        {hochreiter1997long}
\bibfield{author}{\bibinfo{person}{Sepp Hochreiter} {and}
  \bibinfo{person}{J{\"u}rgen Schmidhuber}.} \bibinfo{year}{1997}\natexlab{}.
\newblock \showarticletitle{Long short-term memory}.
\newblock \bibinfo{journal}{\emph{Neural computation}} \bibinfo{volume}{9},
  \bibinfo{number}{8} (\bibinfo{year}{1997}), \bibinfo{pages}{1735--1780}.
\newblock


\bibitem[\protect\citeauthoryear{Jacob, Devlin, and Hofmann}{Jacob
  et~al\mbox{.}}{2020}]%
        {jacob2020s}
\bibfield{author}{\bibinfo{person}{Mikhail Jacob}, \bibinfo{person}{Sam
  Devlin}, {and} \bibinfo{person}{Katja Hofmann}.}
  \bibinfo{year}{2020}\natexlab{}.
\newblock \showarticletitle{``It’s unwieldy and it takes a lot of time'' —
  Challenges and Opportunities for creating agents in commercial games}. In
  \bibinfo{booktitle}{\emph{Proceedings of the AAAI Conference on Artificial
  Intelligence and Interactive Digital Entertainment}},
  Vol.~\bibinfo{volume}{16}. \bibinfo{pages}{88--94}.
\newblock


\bibitem[\protect\citeauthoryear{Johnson, Hofmann, Hutton, and Bignell}{Johnson
  et~al\mbox{.}}{2016}]%
        {johnson2016malmo}
\bibfield{author}{\bibinfo{person}{Matthew Johnson}, \bibinfo{person}{Katja
  Hofmann}, \bibinfo{person}{Tim Hutton}, {and} \bibinfo{person}{David
  Bignell}.} \bibinfo{year}{2016}\natexlab{}.
\newblock \showarticletitle{The Malmo Platform for Artificial Intelligence
  Experimentation.}. In \bibinfo{booktitle}{\emph{Ijcai}}.
  \bibinfo{pages}{4246--4247}.
\newblock


\bibitem[\protect\citeauthoryear{Kanervisto, Pussinen, and
  Hautam{\"a}ki}{Kanervisto et~al\mbox{.}}{2020}]%
        {kanervisto2020benchmarking}
\bibfield{author}{\bibinfo{person}{Anssi Kanervisto}, \bibinfo{person}{Joonas
  Pussinen}, {and} \bibinfo{person}{Ville Hautam{\"a}ki}.}
  \bibinfo{year}{2020}\natexlab{}.
\newblock \showarticletitle{Benchmarking end-to-end behavioural cloning on
  video games}. In \bibinfo{booktitle}{\emph{IEEE conference on games}}. IEEE.
\newblock


\bibitem[\protect\citeauthoryear{Kingma and Ba}{Kingma and Ba}{2014}]%
        {kingma2014adam}
\bibfield{author}{\bibinfo{person}{Diederik~P Kingma} {and}
  \bibinfo{person}{Jimmy Ba}.} \bibinfo{year}{2014}\natexlab{}.
\newblock \showarticletitle{Adam: A method for stochastic optimization}.
\newblock \bibinfo{journal}{\emph{ArXiv}}  \bibinfo{volume}{abs/1412.6980}
  (\bibinfo{year}{2014}).
\newblock


\bibitem[\protect\citeauthoryear{Kingma and Welling}{Kingma and
  Welling}{2013}]%
        {kingma2013auto}
\bibfield{author}{\bibinfo{person}{Diederik~P Kingma} {and}
  \bibinfo{person}{Max Welling}.} \bibinfo{year}{2013}\natexlab{}.
\newblock \showarticletitle{Auto-encoding variational bayes}.
\newblock \bibinfo{journal}{\emph{ArXiv}}  \bibinfo{volume}{abs/1312.6114}
  (\bibinfo{year}{2013}).
\newblock


\bibitem[\protect\citeauthoryear{Lifshitz, Paster, Chan, Ba, and
  McIlraith}{Lifshitz et~al\mbox{.}}{2023}]%
        {lifshitz2023steve}
\bibfield{author}{\bibinfo{person}{Shalev Lifshitz}, \bibinfo{person}{Keiran
  Paster}, \bibinfo{person}{Harris Chan}, \bibinfo{person}{Jimmy Ba}, {and}
  \bibinfo{person}{Sheila McIlraith}.} \bibinfo{year}{2023}\natexlab{}.
\newblock \showarticletitle{STEVE-1: A Generative Model for Text-to-Behavior in
  Minecraft}.
\newblock \bibinfo{journal}{\emph{ArXiv}}  \bibinfo{volume}{abs/2306.00937}
  (\bibinfo{year}{2023}).
\newblock


\bibitem[\protect\citeauthoryear{Liu, Mao, Wu, Feichtenhofer, Darrell, and
  Xie}{Liu et~al\mbox{.}}{2022}]%
        {liu2022convnext}
\bibfield{author}{\bibinfo{person}{Zhuang Liu}, \bibinfo{person}{Hanzi Mao},
  \bibinfo{person}{Chao-Yuan Wu}, \bibinfo{person}{Christoph Feichtenhofer},
  \bibinfo{person}{Trevor Darrell}, {and} \bibinfo{person}{Saining Xie}.}
  \bibinfo{year}{2022}\natexlab{}.
\newblock \showarticletitle{A convnet for the 2020s}. In
  \bibinfo{booktitle}{\emph{IEEE/CVF conference on computer vision and pattern
  recognition}}.
\newblock


\bibitem[\protect\citeauthoryear{Loshchilov and Hutter}{Loshchilov and
  Hutter}{2019}]%
        {loshchilov2019adamw}
\bibfield{author}{\bibinfo{person}{Ilya Loshchilov} {and}
  \bibinfo{person}{Frank Hutter}.} \bibinfo{year}{2019}\natexlab{}.
\newblock \showarticletitle{Decoupled weight decay regularization}. In
  \bibinfo{booktitle}{\emph{International Conference on Learning
  Representations}}.
\newblock


\bibitem[\protect\citeauthoryear{Nair, Rajeswaran, Kumar, Finn, and Gupta}{Nair
  et~al\mbox{.}}{2022}]%
        {nair2022r3m}
\bibfield{author}{\bibinfo{person}{Suraj Nair}, \bibinfo{person}{Aravind
  Rajeswaran}, \bibinfo{person}{Vikash Kumar}, \bibinfo{person}{Chelsea Finn},
  {and} \bibinfo{person}{Abhinav Gupta}.} \bibinfo{year}{2022}\natexlab{}.
\newblock \showarticletitle{R3m: A universal visual representation for robot
  manipulation}. In \bibinfo{booktitle}{\emph{Conference on Robot Learning}}.
\newblock


\bibitem[\protect\citeauthoryear{Oquab, Darcet, Moutakanni, Vo, Szafraniec,
  Khalidov, Fernandez, Haziza, Massa, El-Nouby, et~al\mbox{.}}{Oquab
  et~al\mbox{.}}{2024}]%
        {oquab2023dinov2}
\bibfield{author}{\bibinfo{person}{Maxime Oquab}, \bibinfo{person}{Timoth{\'e}e
  Darcet}, \bibinfo{person}{Th{\'e}o Moutakanni}, \bibinfo{person}{Huy Vo},
  \bibinfo{person}{Marc Szafraniec}, \bibinfo{person}{Vasil Khalidov},
  \bibinfo{person}{Pierre Fernandez}, \bibinfo{person}{Daniel Haziza},
  \bibinfo{person}{Francisco Massa}, \bibinfo{person}{Alaaeldin El-Nouby},
  {et~al\mbox{.}}} \bibinfo{year}{2024}\natexlab{}.
\newblock \showarticletitle{DINOv2: Learning robust visual features without
  supervision}.
\newblock \bibinfo{journal}{\emph{Transactions on Machine Learning Research}}
  (\bibinfo{year}{2024}).
\newblock


\bibitem[\protect\citeauthoryear{Parisi, Rajeswaran, Purushwalkam, and
  Gupta}{Parisi et~al\mbox{.}}{2022}]%
        {parisi2022unsurprising}
\bibfield{author}{\bibinfo{person}{Simone Parisi}, \bibinfo{person}{Aravind
  Rajeswaran}, \bibinfo{person}{Senthil Purushwalkam}, {and}
  \bibinfo{person}{Abhinav Gupta}.} \bibinfo{year}{2022}\natexlab{}.
\newblock \showarticletitle{The (Un)Surprising Effectiveness of Pre-Trained
  Vision Models for Control}. In \bibinfo{booktitle}{\emph{International
  Conference on Machine Learning}}.
\newblock


\bibitem[\protect\citeauthoryear{Pearce, Rashid, Kanervisto, Bignell, Sun,
  Georgescu, Macua, Tan, Momennejad, Hofmann, et~al\mbox{.}}{Pearce
  et~al\mbox{.}}{2023}]%
        {pearce2023imitating}
\bibfield{author}{\bibinfo{person}{Tim Pearce}, \bibinfo{person}{Tabish
  Rashid}, \bibinfo{person}{Anssi Kanervisto}, \bibinfo{person}{Dave Bignell},
  \bibinfo{person}{Mingfei Sun}, \bibinfo{person}{Raluca Georgescu},
  \bibinfo{person}{Sergio~Valcarcel Macua}, \bibinfo{person}{Shan~Zheng Tan},
  \bibinfo{person}{Ida Momennejad}, \bibinfo{person}{Katja Hofmann},
  {et~al\mbox{.}}} \bibinfo{year}{2023}\natexlab{}.
\newblock \showarticletitle{Imitating human behaviour with diffusion models}.
  In \bibinfo{booktitle}{\emph{International Conference on Learning
  Representations}}.
\newblock


\bibitem[\protect\citeauthoryear{Pearce and Zhu}{Pearce and Zhu}{2022}]%
        {pearce2022counter}
\bibfield{author}{\bibinfo{person}{Tim Pearce} {and} \bibinfo{person}{Jun
  Zhu}.} \bibinfo{year}{2022}\natexlab{}.
\newblock \showarticletitle{Counter-strike deathmatch with large-scale
  behavioural cloning}. In \bibinfo{booktitle}{\emph{IEEE Conference on
  Games}}. IEEE, \bibinfo{pages}{104--111}.
\newblock


\bibitem[\protect\citeauthoryear{Radford, Kim, Hallacy, Ramesh, Goh, Agarwal,
  Sastry, Askell, Mishkin, Clark, et~al\mbox{.}}{Radford et~al\mbox{.}}{2021}]%
        {radford2021clip}
\bibfield{author}{\bibinfo{person}{Alec Radford}, \bibinfo{person}{Jong~Wook
  Kim}, \bibinfo{person}{Chris Hallacy}, \bibinfo{person}{Aditya Ramesh},
  \bibinfo{person}{Gabriel Goh}, \bibinfo{person}{Sandhini Agarwal},
  \bibinfo{person}{Girish Sastry}, \bibinfo{person}{Amanda Askell},
  \bibinfo{person}{Pamela Mishkin}, \bibinfo{person}{Jack Clark},
  {et~al\mbox{.}}} \bibinfo{year}{2021}\natexlab{}.
\newblock \showarticletitle{Learning transferable visual models from natural
  language supervision}. In \bibinfo{booktitle}{\emph{International conference
  on machine learning}}. PMLR, \bibinfo{pages}{8748--8763}.
\newblock


\bibitem[\protect\citeauthoryear{Reed, Zolna, Parisotto, Colmenarejo, Novikov,
  Barth-Maron, Gimenez, Sulsky, Kay, Springenberg, et~al\mbox{.}}{Reed
  et~al\mbox{.}}{2022}]%
        {reed2022gato}
\bibfield{author}{\bibinfo{person}{Scott Reed}, \bibinfo{person}{Konrad Zolna},
  \bibinfo{person}{Emilio Parisotto}, \bibinfo{person}{Sergio~Gomez
  Colmenarejo}, \bibinfo{person}{Alexander Novikov}, \bibinfo{person}{Gabriel
  Barth-Maron}, \bibinfo{person}{Mai Gimenez}, \bibinfo{person}{Yury Sulsky},
  \bibinfo{person}{Jackie Kay}, \bibinfo{person}{Jost~Tobias Springenberg},
  {et~al\mbox{.}}} \bibinfo{year}{2022}\natexlab{}.
\newblock \showarticletitle{A generalist agent}.
\newblock \bibinfo{journal}{\emph{Transactions on Machine Learning Research}}
  (\bibinfo{year}{2022}).
\newblock


\bibitem[\protect\citeauthoryear{Rombach, Blattmann, Lorenz, Esser, and
  Ommer}{Rombach et~al\mbox{.}}{2022}]%
        {rombach2022stablediffusion}
\bibfield{author}{\bibinfo{person}{Robin Rombach}, \bibinfo{person}{Andreas
  Blattmann}, \bibinfo{person}{Dominik Lorenz}, \bibinfo{person}{Patrick
  Esser}, {and} \bibinfo{person}{Bj{\"o}rn Ommer}.}
  \bibinfo{year}{2022}\natexlab{}.
\newblock \showarticletitle{High-resolution image synthesis with latent
  diffusion models. 2022 IEEE}. In \bibinfo{booktitle}{\emph{CVF Conference on
  Computer Vision and Pattern Recognition}}.
\newblock


\bibitem[\protect\citeauthoryear{Schneider, Krug, Vaskevicius, Palmieri, and
  Boedecker}{Schneider et~al\mbox{.}}{2024}]%
        {schneider2024surprising}
\bibfield{author}{\bibinfo{person}{Moritz Schneider}, \bibinfo{person}{Robert
  Krug}, \bibinfo{person}{Narunas Vaskevicius}, \bibinfo{person}{Luigi
  Palmieri}, {and} \bibinfo{person}{Joschka Boedecker}.}
  \bibinfo{year}{2024}\natexlab{}.
\newblock \showarticletitle{The Surprising Ineffectiveness of Pre-Trained
  Visual Representations for Model-Based Reinforcement Learning}. In
  \bibinfo{booktitle}{\emph{Advances in Neural Information Processing
  Systems}}.
\newblock


\bibitem[\protect\citeauthoryear{Selvaraju, Cogswell, Das, Vedantam, Parikh,
  and Batra}{Selvaraju et~al\mbox{.}}{2017}]%
        {selvaraju2017grad}
\bibfield{author}{\bibinfo{person}{Ramprasaath~R Selvaraju},
  \bibinfo{person}{Michael Cogswell}, \bibinfo{person}{Abhishek Das},
  \bibinfo{person}{Ramakrishna Vedantam}, \bibinfo{person}{Devi Parikh}, {and}
  \bibinfo{person}{Dhruv Batra}.} \bibinfo{year}{2017}\natexlab{}.
\newblock \showarticletitle{Grad-cam: Visual explanations from deep networks
  via gradient-based localization}. In \bibinfo{booktitle}{\emph{Proceedings of
  the IEEE international conference on computer vision}}.
  \bibinfo{pages}{618--626}.
\newblock


\bibitem[\protect\citeauthoryear{Sestini, Bergdahl, Tollmar, Bagdanov, and
  Gissl{\'e}n}{Sestini et~al\mbox{.}}{2022}]%
        {sestini2022towards}
\bibfield{author}{\bibinfo{person}{Alessandro Sestini}, \bibinfo{person}{Joakim
  Bergdahl}, \bibinfo{person}{Konrad Tollmar}, \bibinfo{person}{Andrew~D
  Bagdanov}, {and} \bibinfo{person}{Linus Gissl{\'e}n}.}
  \bibinfo{year}{2022}\natexlab{}.
\newblock \showarticletitle{Towards Informed Design and Validation Assistance
  in Computer Games Using Imitation Learning}. In
  \bibinfo{booktitle}{\emph{Human in the Loop Learning Workshop at the
  Conference on Neural Information Processing Systems}}.
\newblock


\bibitem[\protect\citeauthoryear{Steiner, Kolesnikov, Zhai, Wightman,
  Uszkoreit, and Beyer}{Steiner et~al\mbox{.}}{2022}]%
        {steiner2022train}
\bibfield{author}{\bibinfo{person}{Andreas Steiner}, \bibinfo{person}{Alexander
  Kolesnikov}, \bibinfo{person}{Xiaohua Zhai}, \bibinfo{person}{Ross Wightman},
  \bibinfo{person}{Jakob Uszkoreit}, {and} \bibinfo{person}{Lucas Beyer}.}
  \bibinfo{year}{2022}\natexlab{}.
\newblock \showarticletitle{How to train your vit? data, augmentation, and
  regularization in vision transformers}.
\newblock \bibinfo{journal}{\emph{Transactions on Machine Learning Research}}
  (\bibinfo{year}{2022}).
\newblock


\bibitem[\protect\citeauthoryear{Team, Raad, Ahuja, Barros, Besse, Bolt,
  Bolton, Brownfield, Buttimore, Cant, Chakera, Chan, Clune, Collister,
  Copeman, Cullum, Dasgupta, de~Cesare, Trapani, Donchev, Dunleavy, Engelcke,
  Faulkner, Garcia, Gbadamosi, Gong, Gonzales, Gregor, Hallingstad, Harley,
  Haves, Hill, Hirst, Hudson, Hughes-Fitt, Rezende, Jasarevic, Kampis, Ke,
  Keck, Kim, Knagg, Kopparapu, Lampinen, Legg, Lerchner, Limont, Liu,
  Loks-Thompson, Marino, Cussons, Matthey, Mcloughlin, Mendolicchio, Merzic,
  Mitenkova, Moufarek, and Oliveira}{Team et~al\mbox{.}}{2024}]%
        {Team2024ScalingIA}
\bibfield{author}{\bibinfo{person}{SIMA Team}, \bibinfo{person}{Maria~Abi
  Raad}, \bibinfo{person}{Arun Ahuja}, \bibinfo{person}{Catarina Barros},
  \bibinfo{person}{Frederic Besse}, \bibinfo{person}{Andrew Bolt},
  \bibinfo{person}{Adrian Bolton}, \bibinfo{person}{Bethanie Brownfield},
  \bibinfo{person}{Gavin Buttimore}, \bibinfo{person}{Max Cant},
  \bibinfo{person}{Sarah Chakera}, \bibinfo{person}{Stephanie C.~Y. Chan},
  \bibinfo{person}{Jeff Clune}, \bibinfo{person}{Adrian Collister},
  \bibinfo{person}{Vikki Copeman}, \bibinfo{person}{Alex Cullum},
  \bibinfo{person}{Ishita Dasgupta}, \bibinfo{person}{Dario de Cesare},
  \bibinfo{person}{Julia~Di Trapani}, \bibinfo{person}{Yani Donchev},
  \bibinfo{person}{Emma Dunleavy}, \bibinfo{person}{Martin Engelcke},
  \bibinfo{person}{Ryan Faulkner}, \bibinfo{person}{Frankie Garcia},
  \bibinfo{person}{Charles Takashi~Toyin Gbadamosi}, \bibinfo{person}{Zhitao
  Gong}, \bibinfo{person}{Lucy Gonzales}, \bibinfo{person}{Karol Gregor},
  \bibinfo{person}{Arne~Olav Hallingstad}, \bibinfo{person}{Tim Harley},
  \bibinfo{person}{Sam Haves}, \bibinfo{person}{Felix Hill},
  \bibinfo{person}{Ed Hirst}, \bibinfo{person}{Drew~A. Hudson},
  \bibinfo{person}{Steph Hughes-Fitt}, \bibinfo{person}{Danilo~J. Rezende},
  \bibinfo{person}{Mimi Jasarevic}, \bibinfo{person}{Laura Kampis},
  \bibinfo{person}{Rosemary Ke}, \bibinfo{person}{Thomas~Albert Keck},
  \bibinfo{person}{Junkyung Kim}, \bibinfo{person}{Oscar Knagg},
  \bibinfo{person}{Kavya Kopparapu}, \bibinfo{person}{Andrew~Kyle Lampinen},
  \bibinfo{person}{Shane Legg}, \bibinfo{person}{Alexander Lerchner},
  \bibinfo{person}{Marjorie Limont}, \bibinfo{person}{Yulan Liu},
  \bibinfo{person}{Maria Loks-Thompson}, \bibinfo{person}{Joseph Marino},
  \bibinfo{person}{Kathryn~Martin Cussons}, \bibinfo{person}{Lo{\"i}c Matthey},
  \bibinfo{person}{Siobhan Mcloughlin}, \bibinfo{person}{Piermaria
  Mendolicchio}, \bibinfo{person}{Hamza Merzic}, \bibinfo{person}{Anna
  Mitenkova}, \bibinfo{person}{Alexandre Moufarek}, {and}
  \bibinfo{person}{Valeria Oliveira}.} \bibinfo{year}{2024}\natexlab{}.
\newblock \showarticletitle{Scaling Instructable Agents Across Many Simulated
  Worlds}.
\newblock \bibinfo{journal}{\emph{ArXiv}}  \bibinfo{volume}{abs/2404.10179}
  (\bibinfo{year}{2024}).
\newblock


\bibitem[\protect\citeauthoryear{Trivedi, Makantasis, Liapis, and
  Yannakakis}{Trivedi et~al\mbox{.}}{2022}]%
        {trivedi2022learning}
\bibfield{author}{\bibinfo{person}{Chintan Trivedi},
  \bibinfo{person}{Konstantinos Makantasis}, \bibinfo{person}{Antonios Liapis},
  {and} \bibinfo{person}{Georgios~N Yannakakis}.}
  \bibinfo{year}{2022}\natexlab{}.
\newblock \showarticletitle{Learning Task-Independent Game State
  Representations from Unlabeled Images}. In \bibinfo{booktitle}{\emph{2022
  IEEE Conference on Games}}.
\newblock


\bibitem[\protect\citeauthoryear{Trivedi, Makantasis, Liapis, and
  Yannakakis}{Trivedi et~al\mbox{.}}{2023}]%
        {trivedi2023towards}
\bibfield{author}{\bibinfo{person}{Chintan Trivedi},
  \bibinfo{person}{Konstantinos Makantasis}, \bibinfo{person}{Antonios Liapis},
  {and} \bibinfo{person}{Georgios~N Yannakakis}.}
  \bibinfo{year}{2023}\natexlab{}.
\newblock \showarticletitle{Towards General Game Representations: Decomposing
  Games Pixels into Content and Style}.
\newblock \bibinfo{journal}{\emph{ArXiv}}  \bibinfo{volume}{abs/2307.11141}
  (\bibinfo{year}{2023}).
\newblock


\bibitem[\protect\citeauthoryear{Vinyals, Babuschkin, Czarnecki, Mathieu,
  Dudzik, Chung, Choi, Powell, Ewalds, Georgiev, et~al\mbox{.}}{Vinyals
  et~al\mbox{.}}{2019}]%
        {vinyals2019grandmaster}
\bibfield{author}{\bibinfo{person}{Oriol Vinyals}, \bibinfo{person}{Igor
  Babuschkin}, \bibinfo{person}{Wojciech~M Czarnecki},
  \bibinfo{person}{Micha{\"e}l Mathieu}, \bibinfo{person}{Andrew Dudzik},
  \bibinfo{person}{Junyoung Chung}, \bibinfo{person}{David~H Choi},
  \bibinfo{person}{Richard Powell}, \bibinfo{person}{Timo Ewalds},
  \bibinfo{person}{Petko Georgiev}, {et~al\mbox{.}}}
  \bibinfo{year}{2019}\natexlab{}.
\newblock \showarticletitle{Grandmaster level in StarCraft II using multi-agent
  reinforcement learning}.
\newblock \bibinfo{journal}{\emph{Nature}} \bibinfo{volume}{575},
  \bibinfo{number}{7782} (\bibinfo{year}{2019}), \bibinfo{pages}{350--354}.
\newblock


\bibitem[\protect\citeauthoryear{Wang, Xie, Jiang, Mandlekar, Xiao, Zhu, Fan,
  and Anandkumar}{Wang et~al\mbox{.}}{2023}]%
        {wang2023voyager}
\bibfield{author}{\bibinfo{person}{Guanzhi Wang}, \bibinfo{person}{Yuqi Xie},
  \bibinfo{person}{Yunfan Jiang}, \bibinfo{person}{Ajay Mandlekar},
  \bibinfo{person}{Chaowei Xiao}, \bibinfo{person}{Yuke Zhu},
  \bibinfo{person}{Linxi Fan}, {and} \bibinfo{person}{Anima Anandkumar}.}
  \bibinfo{year}{2023}\natexlab{}.
\newblock \showarticletitle{Voyager: An open-ended embodied agent with large
  language models}.
\newblock \bibinfo{journal}{\emph{ArXiv}}  \bibinfo{volume}{abs/2305.16291}
  (\bibinfo{year}{2023}).
\newblock


\bibitem[\protect\citeauthoryear{Wurman, Barrett, Kawamoto, MacGlashan,
  Subramanian, Walsh, Capobianco, Devlic, Eckert, Fuchs, et~al\mbox{.}}{Wurman
  et~al\mbox{.}}{2022}]%
        {wurman2022outracing}
\bibfield{author}{\bibinfo{person}{Peter~R Wurman}, \bibinfo{person}{Samuel
  Barrett}, \bibinfo{person}{Kenta Kawamoto}, \bibinfo{person}{James
  MacGlashan}, \bibinfo{person}{Kaushik Subramanian}, \bibinfo{person}{Thomas~J
  Walsh}, \bibinfo{person}{Roberto Capobianco}, \bibinfo{person}{Alisa Devlic},
  \bibinfo{person}{Franziska Eckert}, \bibinfo{person}{Florian Fuchs},
  {et~al\mbox{.}}} \bibinfo{year}{2022}\natexlab{}.
\newblock \showarticletitle{Outracing champion Gran Turismo drivers with deep
  reinforcement learning}.
\newblock \bibinfo{journal}{\emph{Nature}} \bibinfo{volume}{602},
  \bibinfo{number}{7896} (\bibinfo{year}{2022}), \bibinfo{pages}{223--228}.
\newblock


\bibitem[\protect\citeauthoryear{Yang, Li, Dai, and Gao}{Yang
  et~al\mbox{.}}{2022}]%
        {yang2022focal}
\bibfield{author}{\bibinfo{person}{Jianwei Yang}, \bibinfo{person}{Chunyuan
  Li}, \bibinfo{person}{Xiyang Dai}, {and} \bibinfo{person}{Jianfeng Gao}.}
  \bibinfo{year}{2022}\natexlab{}.
\newblock \showarticletitle{Focal modulation networks}. In
  \bibinfo{booktitle}{\emph{Advances in Neural Information Processing
  Systems}}.
\newblock


\bibitem[\protect\citeauthoryear{Yuan, Xue, Yuan, Wang, Wu, Gao, and Xu}{Yuan
  et~al\mbox{.}}{2022}]%
        {yuan2022pre}
\bibfield{author}{\bibinfo{person}{Zhecheng Yuan}, \bibinfo{person}{Zhengrong
  Xue}, \bibinfo{person}{Bo Yuan}, \bibinfo{person}{Xueqian Wang},
  \bibinfo{person}{Yi Wu}, \bibinfo{person}{Yang Gao}, {and}
  \bibinfo{person}{Huazhe Xu}.} \bibinfo{year}{2022}\natexlab{}.
\newblock \showarticletitle{Pre-trained image encoder for generalizable visual
  reinforcement learning}. In \bibinfo{booktitle}{\emph{Advances in Neural
  Information Processing Systems}}.
\newblock


\end{thebibliography}

\clearpage
\appendix

\begin{figure*}[t]
    \centering
    \includegraphics[width=\textwidth]{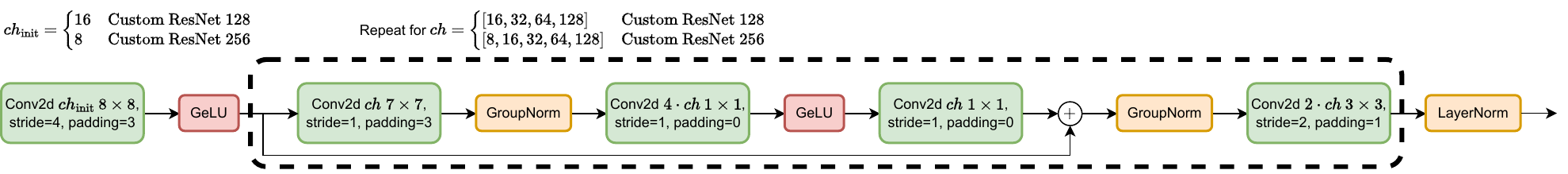}
    \caption{Illustration of the architecture of our custom ResNet visual encoders for $128\times128$ and $256\times256$ images.}
    \label{fig:custom_resnet_architecture}
\end{figure*}

\section{Visual Encoders}
\label{app:visual_encoders}
In this section, we will describe the architectures of all end-to-end visual encoders, the image augmentations applied for end-to-end visual encoders, and detail the sources of the pre-trained encoders used in our study.

\subsection{End-To-End Visual Encoders}
\label{app:end_to_end_encoders}

\paragraph{Impala ResNet} The Impala ResNet architecture faithfully implements the visual encoder of the "large architecture" outlined by \citet{espeholt2018impala} consisting of a $3\times3$ convolution with stride 1, max pooling with $3\times3$ kernels and stride 2 followed by two residual blocks of two $3\times3$ convolutions with stride 1. This joint block is repeated three times with 16, 32, and 32 channels, respectively.

\paragraph{Custom ResNet} The architecture for our custom ResNet models is modelled after \citet{liu2022convnext} and illustrated in detail in \Cref{fig:custom_resnet_architecture}.

\paragraph{ViT} Our ViT architectures are all based on the reference implementation at \url{https://github.com/lucidrains/vit-pytorch/blob/main/vit_pytorch/vit.py}. For all models, we use no dropout, and the following configurations are used across the considered ViT visual encoders:

\begin{table}[h]
    \centering
    \resizebox{\columnwidth}{!}{
        \begin{tabular}{l | c c c c c}
            \toprule
             Model name & Patch size & Num layers & Width & MLP dim & Num heads \\
             \midrule
             ViT Tiny & 16 & 12 & 192 & 768 & 3\\
             Custom ViT & 16 & 4 & 512 & 512 & 12\\
             \bottomrule
        \end{tabular}
    }
    \caption{Configurations of end-to-end ViT models.}
    \label{tab:vit_architecture_details}
\end{table}

The ViT Tiny architecture follows the suggested architecture of \citet{steiner2022train}. In contrast, both custom ViT for $128\times128$ and $256\times256$ have notably fewer layers, wider dimensions of the attention layers and no increase of dimensions in the MLP projections. In our experiments, we found that such an architecture resulted in better online evaluation performance in CS:GO and Minecraft Dungeons.

\paragraph{Image augmentations} If image augmentations are applied during training, we randomly augment images after the down-scaling process. We implement all augmentations with the \texttt{torchvision} library and randomly sample augmentations during training. We apply the following augmentations as described by \citet{baker2022vpt}:
\begin{itemize}
    \item Change colour hue by a random factor between -0.2 and 0.2
    \item Change colour saturation with a random factor between 0.8 and 1.2
    \item Change brightness with a random factor between 0.8 and 1.2
    \item Change colour contrast with a random factor between 0.8 and 1.2
    \item Randomly rotate the image between -2 and 2 degrees
    \item Scale the image with a random factor between 0.98 and 1.02 in each dimension
    \item Apply a random shear to the image between -2 and 2 degrees
    \item Randomly translate the image between -2 and 2 pixels in both the x- and y-direction
\end{itemize}

\subsection{Pre-Trained Visual Encoders}
\label{app:pretrained_encoders}
In this section, we will detail the sources for all pre-trained visual encoders considered in our evaluation.

\paragraph{OpenAI CLIP} For the visual encoders of OpenAI's CLIP models~\citep{radford2021clip}, we use the official interface at \url{https://github.com/openai/CLIP}. We use the following models from this repository: "RN50" (ResNet 50), "ViT-B/16", and "ViT-L/14". In preliminary experiments, we found the available larger ResNet models to provide no significant improvements in online evaluation performance and the ViT model with a larger patch size of 32 to perform worse than the chosen ViT models with patch sizes of 16 and 14. 

\paragraph{DINOv2} For the DINOv2 pre-trained visual encoders~\citep{oquab2023dinov2}, we use the official interface at \url{https://github.com/facebookresearch/dinov2}. Due to the computational cost, we do not evaluate the non-distilled ViT-G/14 checkpoint with 1.1 billion parameters.

\paragraph{FocalNet} For the FocalNet pre-trained visual encoders~\citep{yang2022focal}, we used the Hugging Face \textit{timm} library (\url{https://huggingface.co/docs/timm/index}) to load the pre-trained models for its ease of use. We use the FocalNet models pre-trained on ImageNet-22K classification with 4 focal layers: "focalnet\_large\_fl4", "focalnet\_xlarge\_fl4", and "focalnet\_huge\_fl4".

\paragraph{Stable Diffusion} For the pre-trained stable diffusion 2.1 VAE encoder, we use the Hugging Face checkpoint of the model available at \url{https://huggingface.co/stabilityai/sdxl-vae}. This model can be accessed with the \textit{diffusers} library. In contrast to other encoders, the VAE outputs a Gaussian distribution of embeddings rather than an individual embedding for a given image. We use the mode of the distribution of a given image as its embedding since (1) we want to keep the embeddings of the frozen encoder for a given image deterministic, and (2) we find the standard deviation to be neglectable for most inputs.   

\section{Details for Video Games}
\label{app:games}
Below, we provide more details for each video game we evaluate in, including details about the task, action space, dataset, and online evaluations.

\subsection{Minecraft}
\label{sec:games_minerl}
Minecraft is a game that lets players create and explore a world made of breakable cubes. Players can gather resources, craft items and fight enemies in this open-world sandbox game. Minecraft is also a useful platform for AI research, where different learning algorithms can be tested and compared~\citep{johnson2016malmo}. We use the MineRL~\citep{guss2019minerl,baker2022vpt} environment, which connects Minecraft with Python and allows us to control the agents and the environment. We use MineRL version 1.0.2, which has been used for large-scale imitation learning experiments before~\citep{baker2022vpt}, and which offers simpler mouse and keyboard input than previous MineRL versions~\citep{guss2019minerl}.

\textbf{Action space.} In Minecraft, agents have two continuous actions corresponding to the x- and y-movement of the mouse to control the camera, and eight binary buttons. The buttons control the movement in four directions (forward, backward, rotate left, rotate right), interaction with items or objects, attacking (also used to destroy blocks needed to harvest wood), sprinting, and jumping.

\textbf{Dataset.}
We use the Minecraft dataset released with the OpenAI VPT model~\citep{baker2022vpt} to select demonstrations of tree chopping. We choose the 6.13 version of the dataset and filter it to 40 minutes of human demonstrations that start from a fresh world and chop a tree within 1 minute. We also remove any erroneous files that remain after the filtering. The demonstrations include the image pixels seen by the human player at $640\times360$ resolution and the keyboard and mouse state at the same time, recorded at 20Hz. We also run the models at 20Hz.

\textbf{Online evaluation.}
To evaluate our BC models, we use the ``Treechop" task; after spawning to a new, randomly generated world, the player has to chop a single log of a tree within 1 minute. This is the first step to craft many of the items in Minecraft, and has been previously used to benchmark reinforcement learning algorithms~\citep{guss2019minerl}. See \Cref{fig:minecraft_screenshot} for a screenshot of the starting state. The agent observes the shown image pixels in first-person perspective, can move the player around and attack to chop trees. For reporting the performance of trained models, we rollout each model for 100 episodes with the same world seeds, and record the number of trees the player chopped. If the player chopped at least one tree within the first minute, the episode is counted as a success, otherwise it is counted as a failure (the timeout is set to 1 minute).

\subsection{Counter-Strike: Global Offensive}
\label{sec:games_csgo}
CS:GO is a first-person shooter game designed for competitive, five versus five games. The core skill of the game is accurate aiming and handling the weapon recoil/sway as the weapon is fired. Previous work has used CS:GO as a benchmark to train and test behavioural cloning models~\citep{pearce2023imitating}, with best models able to outperform easier bots~\citep{pearce2022counter}. We incorporate experiments using CS:GO, as it offers visuals more similar to the real-world images that most pre-trained visual encoders were trained on, in contrast to our primary evaluation in Minecraft Dungeons and Minecraft (see \Cref{fig:csgo_screenshot}). %

\textbf{Action space.}
We represent the mouse movement in x- and y-direction as two continuous values, and the left mouse click as a binary button.

\textbf{Dataset.}
Following \citet{pearce2022counter}, we use the ``Clean aim train" dataset and setup. The controlled player is placed in the middle of an arena, and random enemies are spawned around them who try to reach the player. %
The player can not move; they can only aim and shoot (\Cref{fig:csgo_screenshot}). The dataset contains 45 minutes of expert-human gameplay from one player, recorded at 16Hz. 

\textbf{Online evaluation.}
To evaluate models, we run each model for three rollouts of five minutes each in the ``Clean aim train" environment at 16Hz, and report the average and standard deviation of the kills-per-minute. 

\subsection{Minecraft Dungeons}
\label{sec:games_dungeons}
Minecraft Dungeons is an action-adventure role-playing video game with isometric camera view centered on the player. The player controls the movement and actions (including dodge roll, attack, use health potion, use items) of a single character which is kept in the center of the video frame (as seen in \Cref{fig:dungeons_screenshot}). The player has to complete diverse levels by following and completing several objectives. In our evaluation, we focus on the ``Arch Haven'' level of Minecraft Dungeons which contains fighting against several types of enemies and navigation across visually diverse terrain.

\textbf{Action space.} Agents have access to all effective controls in Minecraft Dungeons, including the x- and y-positions of both joysticks as four continuous values in the range $[-1, 1]$, the right trigger position (for shooting the bow), and ten buttons as binary actions. The most frequently used buttons during recordings control sword attacks, bow shooting, healing potions, and forward dodging. 

\textbf{Dataset.} Before data collection, we pre-registered this study with our Institutional Review Board (IRB) who advised on the drafting of our participant instructions to ensure informed consent. After their approval, four players\footnote{120 recordings were collected by one player with the remaining 19 recordings being roughly evenly split across the other three players.} played the ``Arch Haven'' level, and game frames at $1280\times720$ resolution, actions (joystick positions and button presses on a controller), and character position
within the level were captured. The dataset includes a total of 139 recorded trajectories with more than eight hours of gameplay at 30Hz.
Individual demonstrations vary between 160 and 380 seconds which corresponds to 4,800 and 11,400 recorded actions, respectively. We use 80\% of the data for training and reserve 20\% for validation.
Each player was instructed to complete the level using a fixed character equipped with only the starting equipment of a sword and bow, and most players followed the immediate path towards level completion.

\textbf{Online evaluation.} To evaluate the quality of trained BC policies, we rollout the policy in the game with actions being queried at 10Hz (see \Cref{app:dungeons_online_fps} for details).
These actions are then taken in the game using Xbox controller emulation software. Each rollout spawns the agent in the beginning of the ``Arch Haven'' level and queries actions until five minutes passed (3,000 actions) or the agent dies four times resulting in the level being failed. We run 20 rollouts per trained agent and report the progression throughout the level (\Cref{app:dungeons_archhaven}).

\section{Training Loss Curves}
\label{app:evaluation_data}
In this section, we learning curves for the training loss of all models across Minecraft Dungeons and Minecraft.

\begin{figure*}[t]
    \centering
    \begin{subfigure}{.99\textwidth}
        \includegraphics[width=.45\textwidth]{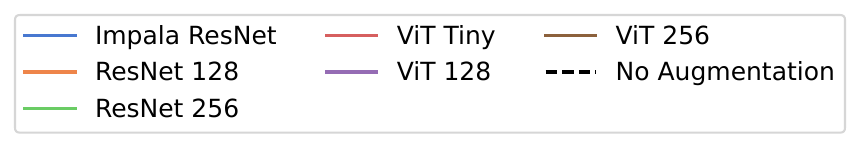}
        \includegraphics[width=.45\textwidth]{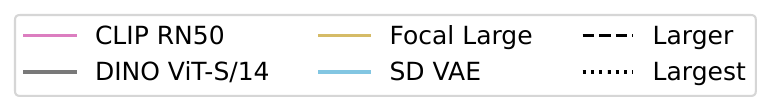}
    \end{subfigure}
    \begin{subfigure}{.99\textwidth}
        \includegraphics[width=.45\textwidth]{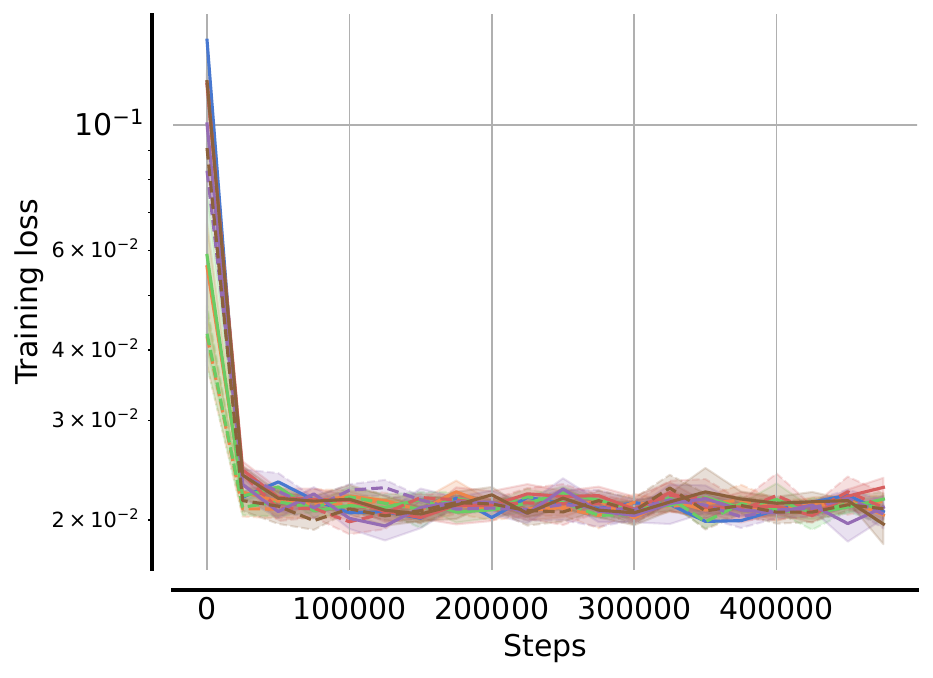}
        \includegraphics[width=.45\textwidth]{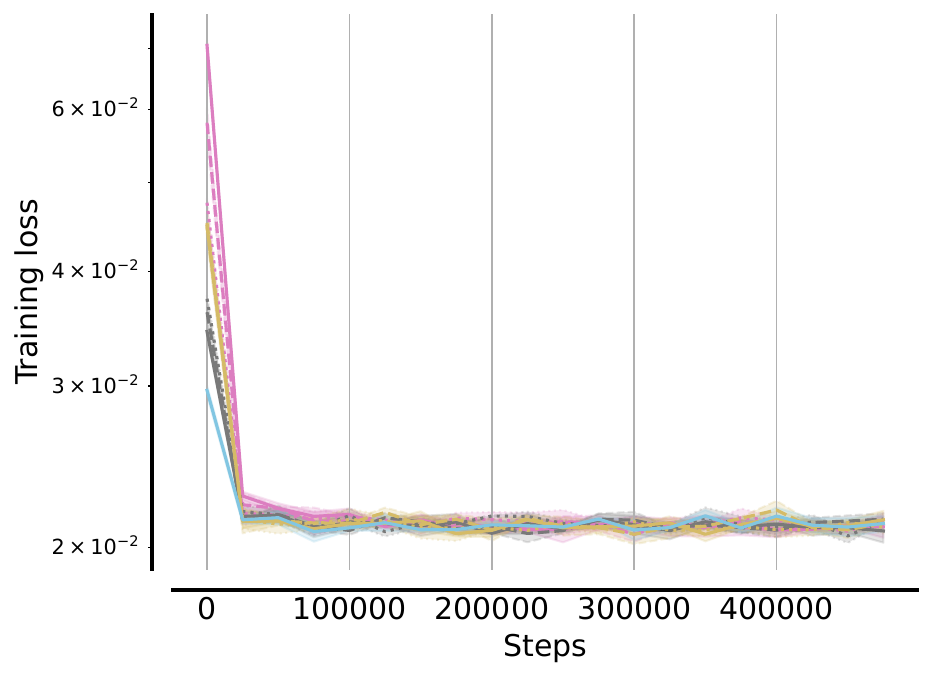}
        \caption{Minecraft}
        \label{fig:minerl_train_loss}
    \end{subfigure}
    \begin{subfigure}{0.99\textwidth}
        \includegraphics[width=.45\textwidth]{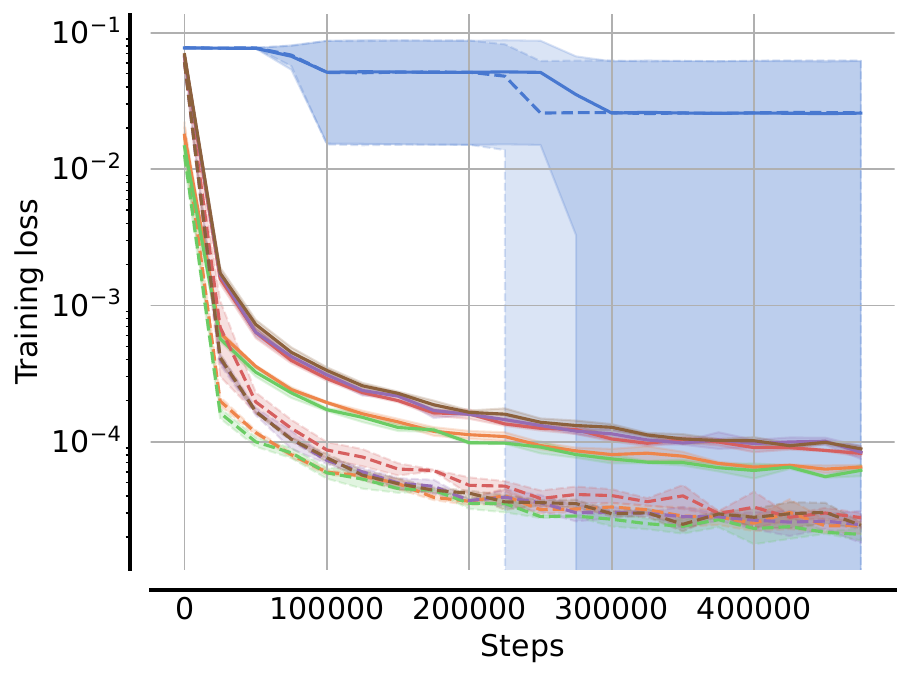}
        \includegraphics[width=.45\textwidth]{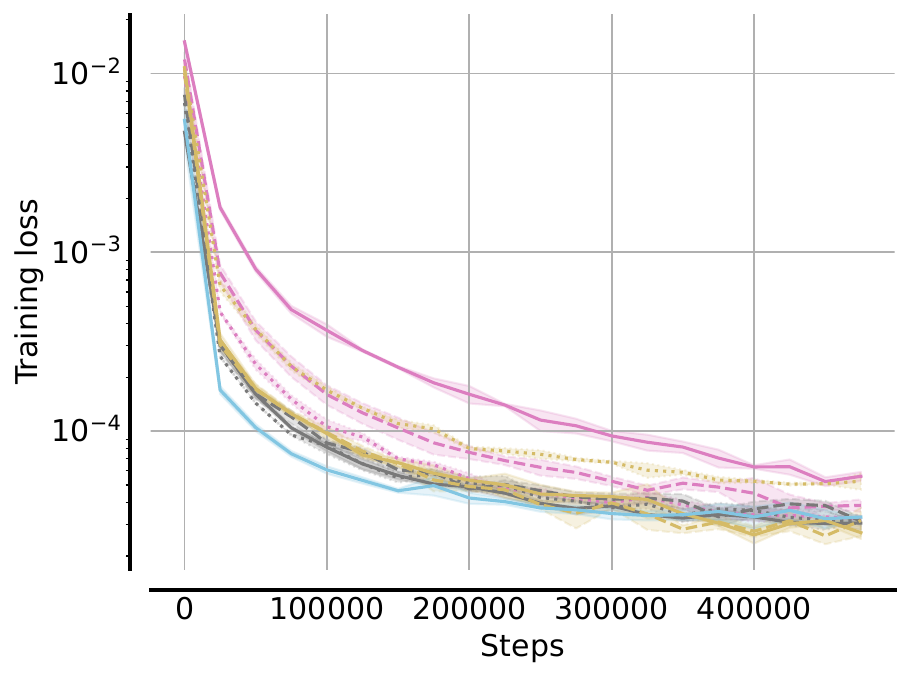}
        \caption{Counter-Strike: Global Offensive}
        \label{fig:csgo_train_loss}
    \end{subfigure}
    \begin{subfigure}{0.99\textwidth}
        \includegraphics[width=.45\textwidth]{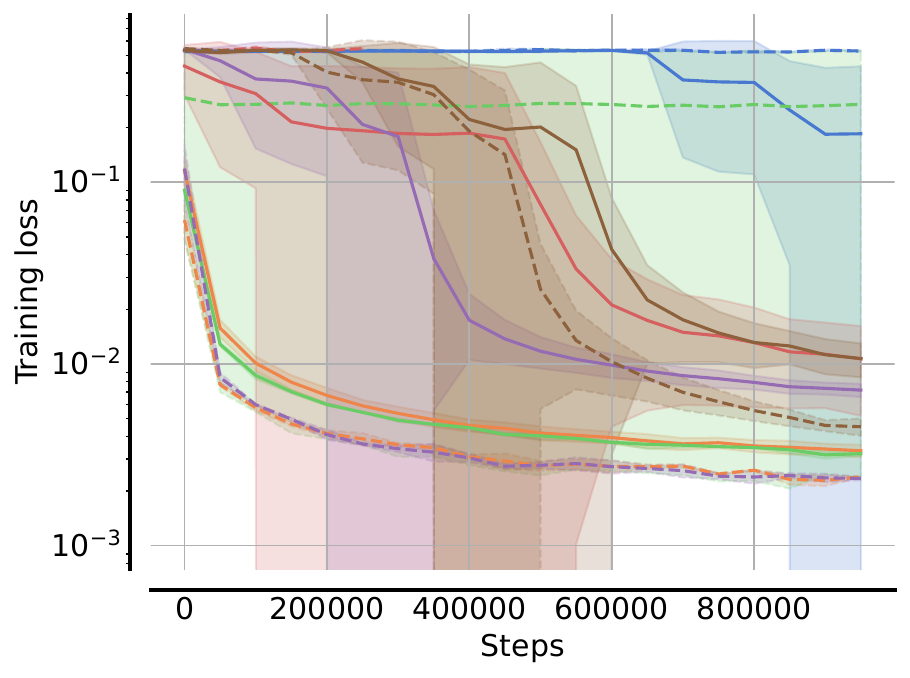}
        \includegraphics[width=.45\textwidth]{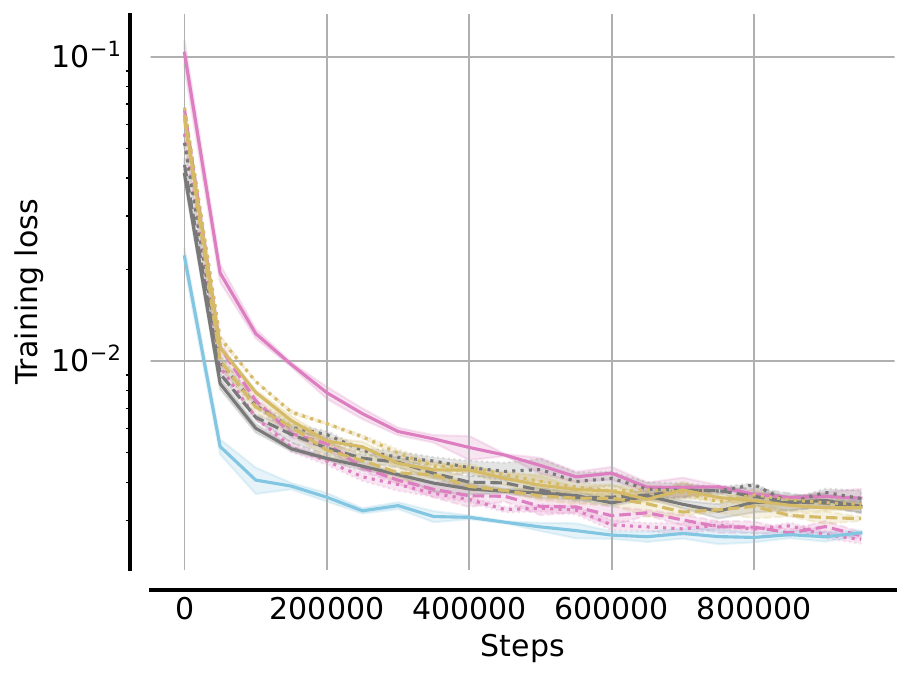}
        \caption{Minecraft Dungeons}
        \label{fig:dungeons_train_loss}
    \end{subfigure}
    \caption{Training loss in log-scale for BC agents in Minecraft (top), Counter-Strike: Global Offensive (middle), and Minecraft Dungeons (bottom) with end-to-end trained (left) and pre-trained (right) visual encoders. We visualise the mean and standard deviation across three seeds after computing window-averaged training losses at twenty regular intervals throughout training.}
    \label{fig:training_loss}
\end{figure*}

\paragraph{Minecraft} \Cref{fig:minerl_train_loss} shows the training loss of end-to-end and pre-trained visual encoders in Minecraft. We find that the training loss for all end-to-end and pre-trained visual encoders in Minecraft plateaus after less than \num{50000} training steps and all encoders appear to converge to a similar training loss. This might indicate that comparably short training might be sufficient given the small dataset in Minecraft, but we have observed that online rollout performance can increase even after training loss stagnates. Furthermore, this indicates that training loss is a poor indicator of online performance in Minecraft given we observed significant differences in online performance as seen in \Cref{tab:minecraft_csgo_online_evaluation}.

\paragraph{Counter-Strike: Global Offensive} \Cref{fig:csgo_train_loss} shows the training loss for all models in CS:GO. In contrast to Minecraft, we can see that the training loss improves all throughout training for all trained models, indicating that the training task in CS:GO might be harder to learn compared to Minecraft. This is somewhat surprising given the comparable amount of demonstrations and lower dimensional action space of CS:GO compared to Minecraft. As expected, we can see that visual encoders trained end-to-end with image augmentations generally have a larger training loss despite improving online performance in most cases as seen in \Cref{tab:minecraft_csgo_online_evaluation}. Furthermore, we observe that the Impala ResNet models exhibit comparably high dispersion across three seeds, leading to large shading and stagnating training loss early in training. We hypothesise that this occurs due to the very large embeddings of the Impala encoders that make training a BC policy challenging. However, while the large dispersion across runs is also observed for the online rollout performance of Impala ResNet, in aggregate, we observe strong online performance of the Impala ResNet models in CS:GO (\Cref{tab:minecraft_csgo_online_evaluation}) despite the high training loss. Among pre-trained encoders, we find most models to converge to similar training loss values that are also comparable to end-to-end trained models.

\paragraph{Minecraft Dungeons} \Cref{fig:dungeons_train_loss} shows the training loss for all models with end-to-end and pre-trained visual encoders in Minecraft Dungeons. Generally, we observe similar trends for Minecraft Dungeons as we observe for CS:GO with slowly converging training losses, image augmentations leading to higher training losses for end-to-end trained models, and the Impala ResNet models exhibiting high training loss that stagnates early in training and high dispersion across random seeds. Similarly, we observe that the training loss for the ResNet with $256\times256$ images trained without image augmentations stagnates early in training and exhibits high dispersion across seeds. Among the models with pre-trained visual encoders, the training loss appears comparable for most models. Only the reconstruction-based stable diffusion encoder and the CLIP ResNet50 models stand out with the lowest and highest training loss throughout training, respectively. Comparing the training loss of models with end-to-end and pre-trained visual encoders further shows that end-to-end encoders trained without image augmentation are capable of reaching lower losses. We hypothesise that this occurs since the end-to-end trained encoders are specialised to perform well on the exact training data the loss is computed over.

\clearpage

\section{Minecraft Dungeons Arch Haven Level}
\label{app:dungeons_archhaven}
\begin{figure*}[t]
    \centering
    \begin{subfigure}{.49\textwidth}
        \includegraphics[width=\textwidth,trim={6em 6em 6em 6em},clip]{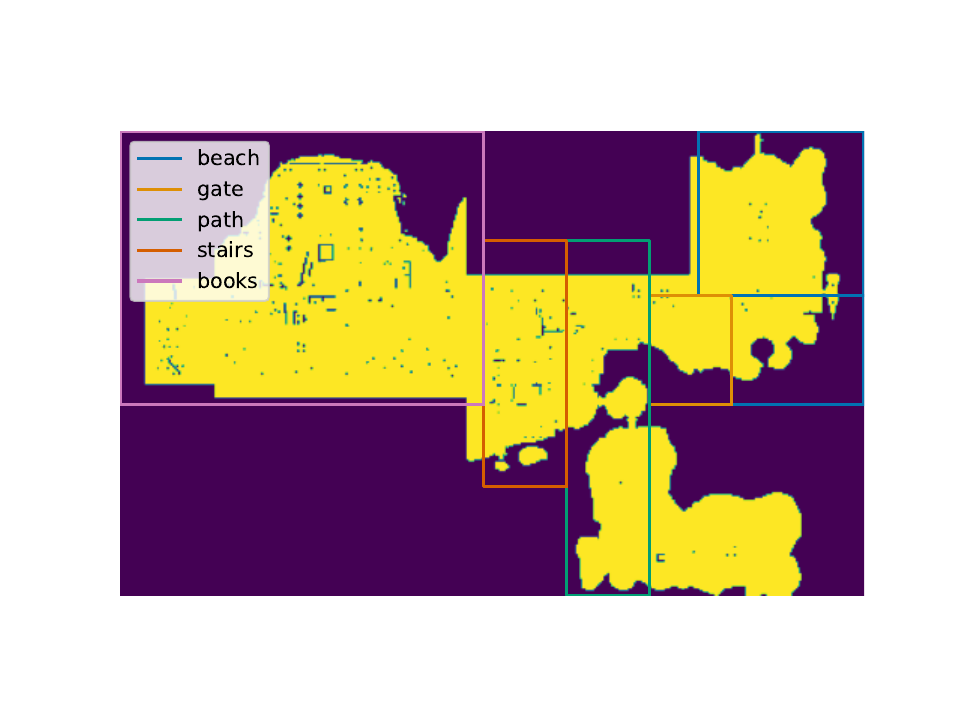}
        \caption{Progression zones}
        \label{fig:archhaven_progression_zones}
    \end{subfigure}
    \begin{subfigure}{.45\textwidth}
        \includegraphics[width=\textwidth,trim={1em 1em 1em 1em},clip]{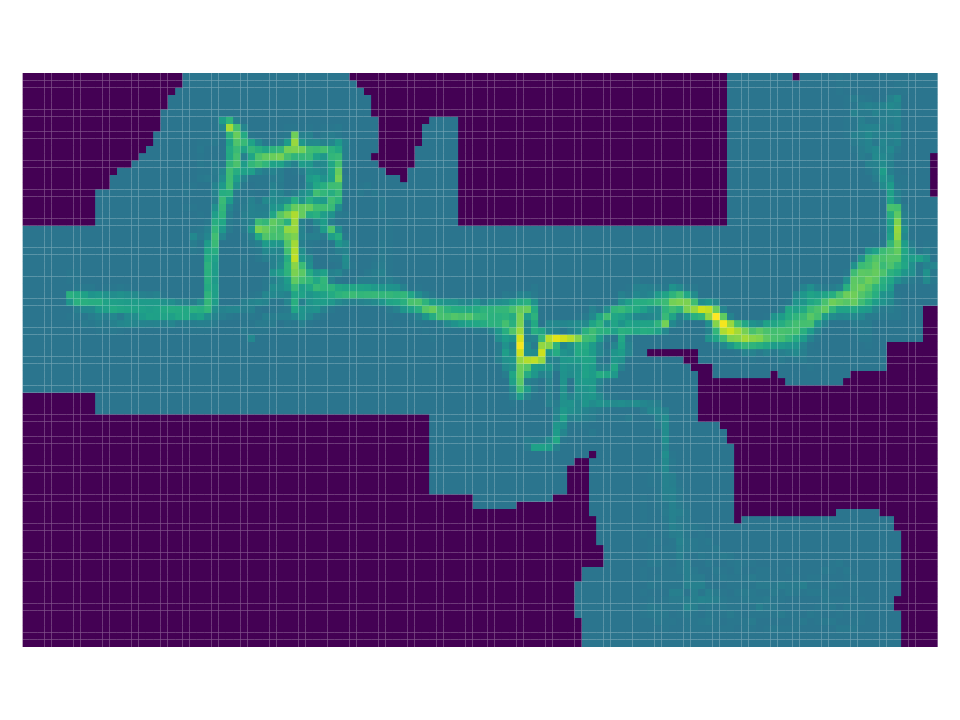}
        \caption{Human dataset heatmap}
        \label{fig:archhaven_location_heatmap}
    \end{subfigure}
    \caption{(a) A visualisation of the boundaries of each progression zone in the "Arch Haven" level in Minecraft Dungeons used for online evaluations. (b) A heatmap visualising the visited locations of the human dataset of demonstrations within the "Arch Haven" level.}
    \label{fig:archhaven}
\end{figure*}

To measure progress for the online evaluation in Minecraft Dungeons, we define boundaries of zones which determine the progression throughout the "Arch Haven" level we evaluate in. These zones and a heatmap showing the visited locations of the human demonstrations used for training are visualised in \Cref{fig:archhaven}. The heatmap also shows the path followed by most demonstrations towards completion of the level.

\section{Minecraft Dungeons Action Frequency in Online Evaluation}
\label{app:dungeons_online_fps}
The visual encoders used in our evaluation have vastly different model sizes (see \Cref{tab:encoder_overview}) and, thus, notably different computational cost at inference time. This is particularly challenging during online evaluation in Minecraft Dungeons, since there exists no programmatic interface to pause or slow down the game like in Minecraft and CS:GO. We attempt to take actions during evaluation at 10Hz to match the action selection frequency of the (processed) training data, in particular due to the recurrent architecture of our policy. However, we are unable to perfectly match this frequency for all visual encoders on the hardware used to conduct the evaluation (see \Cref{app:hardware_specification} for specifications on the hardware used during training and online evaluation) despite using a more powerful GPU for pre-trained visual encoders due to their comparably large size.

\Cref{tab:dungeons_online_fps} lists the average action frequencies of all models during online evaluation in Minecraft Dungeons across all runs conducted as part of our evaluation. We note that most end-to-end trained visual encoders enable fast inference achieving close to 10 Hz action frequency. The ViT Tiny model is the slowest model, likely due to its deeper 12 layers in comparison to the other end-to-end trained ViT models with 4 layers as shown in \Cref{tab:vit_architecture_details}, but we are still able to take actions at more than 8.5Hz. For pre-trained visual encoders, we see comparably fast action frequencies for all CLIP and most DINOv2 models as. The largest DINOv2 and stable diffusion VAE have notably slower action frequencies, but the FocalNet models induced the highest inference cost. However, we highlight that we did not observe behaviour during online evaluation which would suggest that these models were significantly inhibited due to this discrepancy.

\begin{table}[h]
    \centering
    \caption{Average action frequencies during online evaluation in Minecraft Dungeons across 60 runs per model (20 for each seed).}
    \label{tab:dungeons_online_fps}
    \begin{tabular}{l c}
        \toprule
        Model name & \text{Action freq. (Hz)}\\
        \midrule
        Impala ResNet & 9.83\\
        ResNet 128 & 9.90\\
        ResNet 256 & 9.81\\
        ViT Tiny & 8.63\\
        ViT 128 & 9.90\\
        ViT 256 & 9.46\\
        \midrule
        Impala ResNet +Aug & 9.78\\
        ResNet 128 +Aug & 9.67\\
        ResNet 256 +Aug & 9.62\\
        ViT Tiny +Aug & 8.77\\
        ViT 128 +Aug & 9.69\\
        ViT 256 +Aug & 9.63\\
        \midrule
        CLIP ResNet50 & 9.85\\
        CLIP ViT-B/16 & 9.84\\
        CLIP ViT-L/14 & 9.71\\
        \midrule
        DINOv2 ViT-S/14 & 9.81\\
        DINOv2 ViT-B/14 & 9.81\\
        DINOv2 ViT-L/14 & 7.93\\
        \midrule
        FocalNet Large & 8.00\\
        FocalNet XLarge & 6.13\\
        FocalNet Huge & 6.91\\
        \midrule
        Stable Diffusion VAE & 8.77\\
        \bottomrule
    \end{tabular}
\end{table}

\section{Training and Evaluation Hardware}
\label{app:hardware_specification}

All training runs have been completed using Azure compute using a mix of Nvidia 16GB V100s, 32GB V100s and A6000 GPUs.

\paragraph{Minecraft Dungeons} For Minecraft Dungeons, end-to-end training runs for Impala ResNet, custom ResNets (for $128\times128$ and $256\times256$ images) and custom ViT for $128\times128$ images without image augmentation have been done on four 16GB V100s for each run. Training runs for the same models with image augmentation have been run on one A6000 GPU (with 48GB of VRAM) for each run. Training the ViT Tiny and ViT model for $256\times256$ images needed more VRAMs, so these were trained on eight 16GB V100s for each run.

For training runs using pre-trained visual encoders, we computed the embeddings of all images in the Minecraft Dungeons dataset prior to training for more efficient training using A6000 GPUs. After, we were able to train each model using pre-trained visual encoders with four 16GB V100s for a single run.

To train models on half or a quarter of the training data for the third set of experiments, we used four 16GB V100s for a single run of any configuration.

Since the Minecraft Dungeons game is unable to run on Linux servers, we used Azure virtual machines running Windows 10 for the online evaluation. For evaluation of end-to-end trained models, we use a machine with two M60 GPUs, 24 CPU cores and 224GB of RAM. However, we noticed that this configuration was insufficient to evaluate models with larger pre-trained visual encoders at the desired 10Hz. Therefore, we used a configuration with one A10 GPU, 18 CPU cores and 220GB of RAM which was able to run the game and rollout the trained policy close to the desired 10Hz for all models.

\paragraph{Minecraft} The training hardware is similar to Minecraft Dungeons, with A6000s used for embedding/training with pretrained models, and 32GB V100s used to train the end-to-end models. Training pretrained models took considerably less time, with most models training within hours on a single A6000 GPU. 

Minecraft evaluation was performed on remote Linux machines with A6000s, as MineRL is able to run on headless machines with virtual X buffers (\texttt{xvfb}). Each GPU had maximum of three rollouts happening concurrently, with each rollout running at 3-9 frames per second, depending on the model size.

\paragraph{Counter-Strike: Global Offensive} Training was performed on the same hardware as with Minecraft experiments. For evaluation, we ran CS:GO on local Windows machines, equipped with either a GTX 1650Ti or a GTX 980 GPU, as per instructions in the original CS:GO paper~\cite{pearce2022counter}. We ran the game at lower speeds (and adjusted action rate accordingly) to allow models to predict actions in time to match the 16Hz action frequency.

\section{Grad-Cam Visualisations}
\label{app:grad_cam}
To generate Grad-CAM~\citep{selvaraju2017grad} visualisations, we use the library available at \url{https://github.com/jacobgil/pytorch-grad-cam}. We use all actions of the policy trained on the embeddings of each visual encoder as the target concept to analyse, and visualise the average Grad-CAM plot across all actions. Following \url{https://github.com/jacobgil/pytorch-grad-cam#chosing-the-target-layer}, we use the activations of these layers within the visual encoders to compute visualisations for:
\begin{itemize}
    \item ResNet: Activations across the last ResNet block
    \item ViT: Activations across the layer normalisation before the last attention block
    \item FocalNet: Activations across the layer normalisation before the last focal modulation block
    \item SD VAE: Activations across the last ResNet block within the mid-block of the encoder
\end{itemize}

\captionsetup{font=footnotesize}
\begin{figure*}[h]
    \centering
    \begin{adjustbox}{valign=t}  
        \begin{minipage}{0.335\linewidth} 
            \centering
            \includegraphics[width=\textwidth]{images/grad_cam/original_images/frame_300.jpg}
            \subcaption{Original image}
        \end{minipage}
    \end{adjustbox}
    \hspace{3cm}
    \begin{adjustbox}{valign=t}  
        \begin{minipage}{0.2\linewidth}
            \includegraphics[width=\textwidth,trim={0.5em 0.5em 0.5em 0.5em},clip]{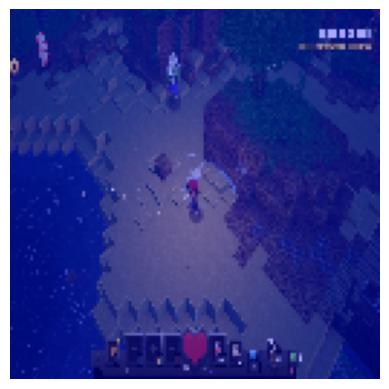}
            \subcaption{Impala ResNet}
        \end{minipage}
    \end{adjustbox}
    \begin{adjustbox}{valign=t}  
        \begin{minipage}{0.2\linewidth}
            \includegraphics[width=\textwidth,trim={0.5em 0.5em 0.5em 0.5em},clip]{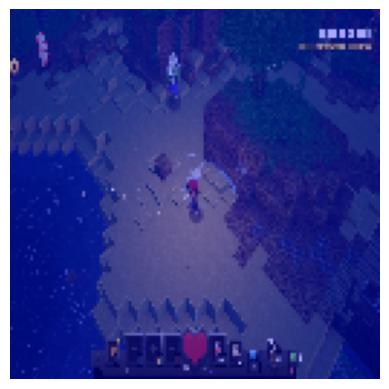}
            \subcaption{Impala ResNet +Aug}
        \end{minipage}
    \end{adjustbox}
    
    \begin{adjustbox}{valign=t}  
        \begin{minipage}{0.19\linewidth}
            \includegraphics[width=\textwidth,trim={0.5em 0.5em 0.5em 0.5em},clip]{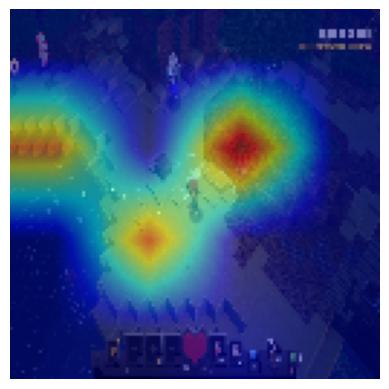}
            \subcaption{ResNet 128}
        \end{minipage}
    \end{adjustbox}
    \begin{adjustbox}{valign=t}  
        \begin{minipage}{0.19\linewidth}
            \includegraphics[width=\textwidth,trim={0.5em 0.5em 0.5em 0.5em},clip]{images/grad_cam/end_to_end_encoders/custom_resnet_128_aug_dungeons_frame_300_actions.jpg}
            \subcaption{ResNet 128 +Aug}
        \end{minipage}
    \end{adjustbox}
    \begin{adjustbox}{valign=t}  
        \begin{minipage}{0.19\linewidth}
            \includegraphics[width=\textwidth,trim={0.5em 0.5em 0.5em 0.5em},clip]{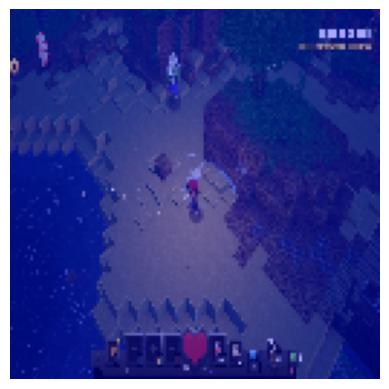}
            \subcaption{ResNet 256}
        \end{minipage}
    \end{adjustbox}
    \begin{adjustbox}{valign=t}  
        \begin{minipage}{0.19\linewidth}
            \includegraphics[width=\textwidth,trim={0.5em 0.5em 0.5em 0.5em},clip]{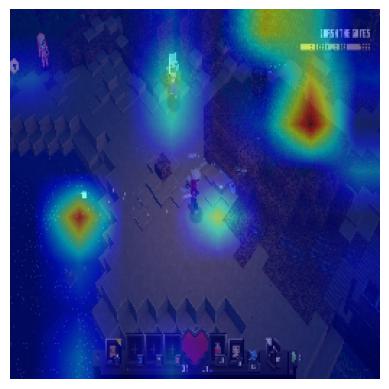}
            \subcaption{ResNet 256 +Aug}
        \end{minipage}
    \end{adjustbox}
    \begin{adjustbox}{valign=t}  
        \begin{minipage}{0.19\linewidth}
            \includegraphics[width=\textwidth,trim={0.5em 0.5em 0.5em 0.5em},clip]{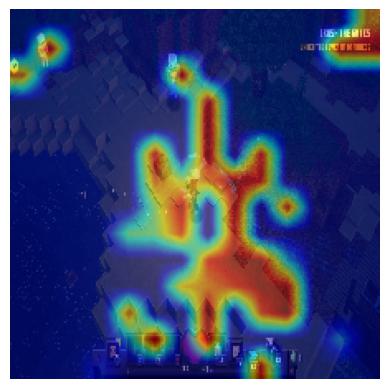}
            \subcaption{ViT Tiny}
        \end{minipage}
    \end{adjustbox}
    
    \begin{adjustbox}{valign=t}  
        \begin{minipage}{0.19\linewidth}
            \includegraphics[width=\textwidth,trim={0.5em 0.5em 0.5em 0.5em},clip]{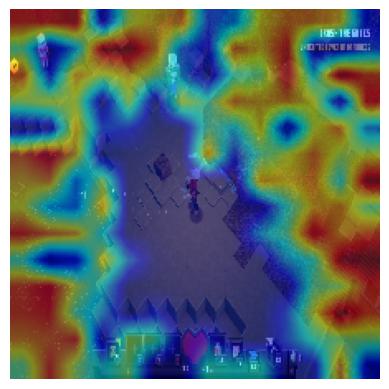}
            \subcaption{ViT Tiny +Aug}
        \end{minipage}
    \end{adjustbox}
    \begin{adjustbox}{valign=t}  
        \begin{minipage}{0.19\linewidth}
            \includegraphics[width=\textwidth,trim={0.5em 0.5em 0.5em 0.5em},clip]{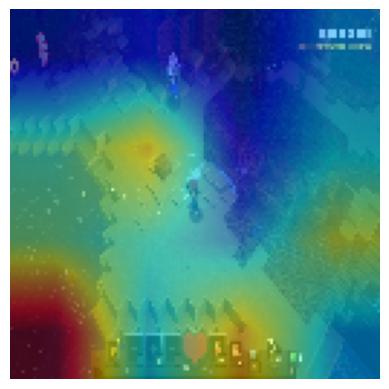}
            \subcaption{ViT 128}
        \end{minipage}
    \end{adjustbox}
    \begin{adjustbox}{valign=t}  
        \begin{minipage}{0.19\linewidth}
            \includegraphics[width=\textwidth,trim={0.5em 0.5em 0.5em 0.5em},clip]{images/grad_cam/end_to_end_encoders/custom_vit_128_aug_dungeons_frame_300_actions.jpg}
            \subcaption{ViT 128 +Aug}
        \end{minipage}
    \end{adjustbox}
    \begin{adjustbox}{valign=t}  
        \begin{minipage}{0.19\linewidth}
            \includegraphics[width=\textwidth,trim={0.5em 0.5em 0.5em 0.5em},clip]{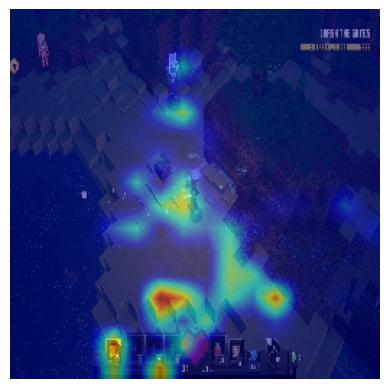}
            \subcaption{ViT 256}
        \end{minipage}
    \end{adjustbox}
    \begin{adjustbox}{valign=t}  
        \begin{minipage}{0.19\linewidth}
            \includegraphics[width=\textwidth,trim={0.5em 0.5em 0.5em 0.5em},clip]{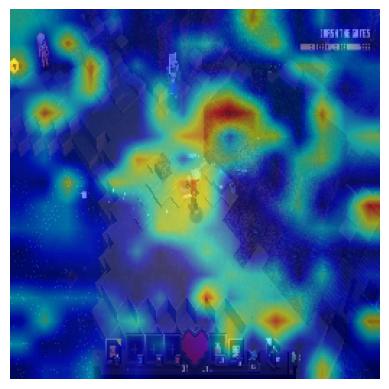}
            \subcaption{ViT 256 +Aug}
        \end{minipage}
    \end{adjustbox}

    \begin{adjustbox}{valign=t}  
        \begin{minipage}{0.19\linewidth}
            \includegraphics[width=\textwidth,trim={0.5em 0.5em 0.5em 0.5em},clip]{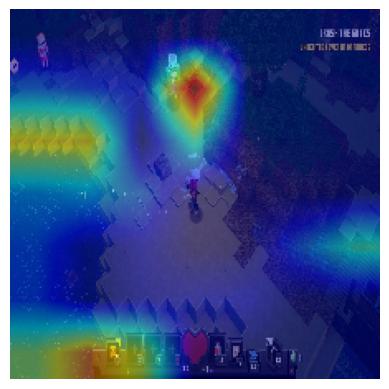}
            \subcaption{CLIP RN50}
        \end{minipage}
    \end{adjustbox}
    \begin{adjustbox}{valign=t}  
        \begin{minipage}{0.19\linewidth}
            \includegraphics[width=\textwidth,trim={0.5em 0.5em 0.5em 0.5em},clip]{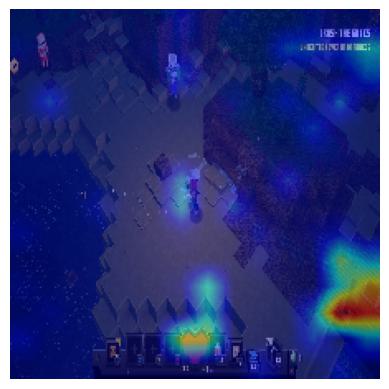}
            \subcaption{CLIP ViT-B/16}
        \end{minipage}
    \end{adjustbox}
    \begin{adjustbox}{valign=t}  
        \begin{minipage}{0.19\linewidth}
            \includegraphics[width=\textwidth,trim={0.5em 0.5em 0.5em 0.5em},clip]{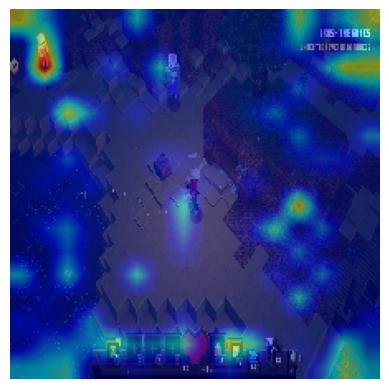}
            \subcaption{CLIP ViT-L/14}
        \end{minipage}
    \end{adjustbox}
    \begin{adjustbox}{valign=t}  
        \begin{minipage}{0.19\linewidth}
            \includegraphics[width=\textwidth,trim={0.5em 0.5em 0.5em 0.5em},clip]{images/grad_cam/pretrained_encoders/dino_vits_dungeons_frame_300_actions.jpg}
            \subcaption{DINOv2 ViT-S/14}
        \end{minipage}
    \end{adjustbox}
    \begin{adjustbox}{valign=t}  
        \begin{minipage}{0.19\linewidth}
            \includegraphics[width=\textwidth,trim={0.5em 0.5em 0.5em 0.5em},clip]{images/grad_cam/pretrained_encoders/dino_vitb_dungeons_frame_300_actions.jpg}
            \subcaption{DINOv2 ViT-B/14}
        \end{minipage}
    \end{adjustbox}

    \begin{adjustbox}{valign=t}  
        \begin{minipage}{0.19\linewidth}
            \includegraphics[width=\textwidth,trim={0.5em 0.5em 0.5em 0.5em},clip]{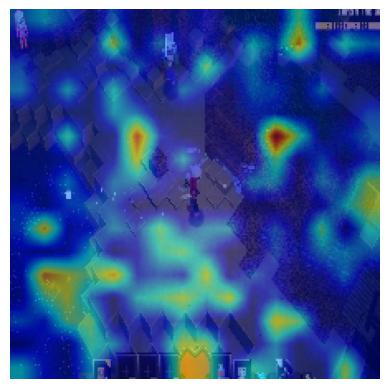}
            \subcaption{DINOv2 ViT-L/14}
        \end{minipage}
    \end{adjustbox}
    \begin{adjustbox}{valign=t}  
        \begin{minipage}{0.19\linewidth}
            \includegraphics[width=\textwidth,trim={0.5em 0.5em 0.5em 0.5em},clip]{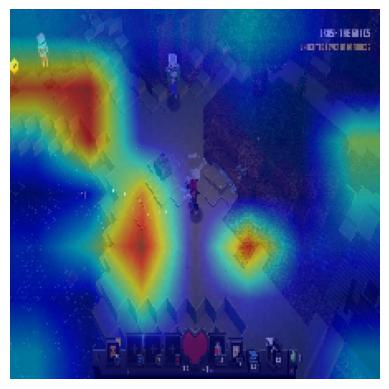}
            \subcaption{Focal Large}
        \end{minipage}
    \end{adjustbox}
    \begin{adjustbox}{valign=t}  
        \begin{minipage}{0.19\linewidth}
            \includegraphics[width=\textwidth,trim={0.5em 0.5em 0.5em 0.5em},clip]{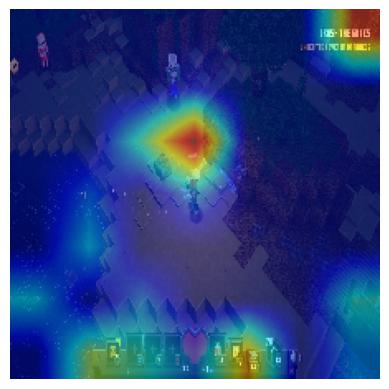}
            \subcaption{Focal XLarge}
        \end{minipage}
    \end{adjustbox}
    \begin{adjustbox}{valign=t}  
        \begin{minipage}{0.19\linewidth}
            \includegraphics[width=\textwidth,trim={0.5em 0.5em 0.5em 0.5em},clip]{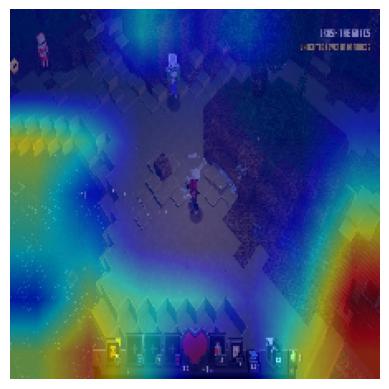}
            \subcaption{Focal Huge}
        \end{minipage}
    \end{adjustbox}
    \begin{adjustbox}{valign=t}  
        \begin{minipage}{0.19\linewidth}
            \includegraphics[width=\textwidth,trim={0.5em 0.5em 0.5em 0.5em},clip]{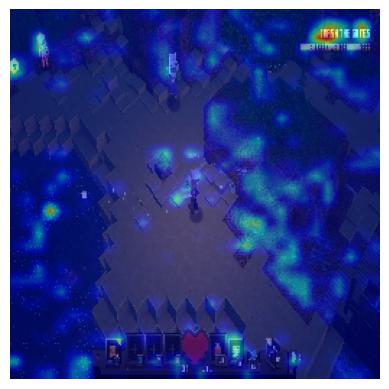}
            \subcaption{SD VAE}
        \end{minipage}
    \end{adjustbox}

    \caption{Grad-Cam visualisations for all encoders (seed 0) in Minecraft Dungeons with policy action logits serving as the targets.}
    \label{fig:grad_cam_dungeons_frame_300_actions}
\end{figure*}

\begin{figure*}[h]
    \centering
    \begin{adjustbox}{valign=t}  
        \begin{minipage}{0.335\linewidth} 
            \centering
            \includegraphics[width=\textwidth]{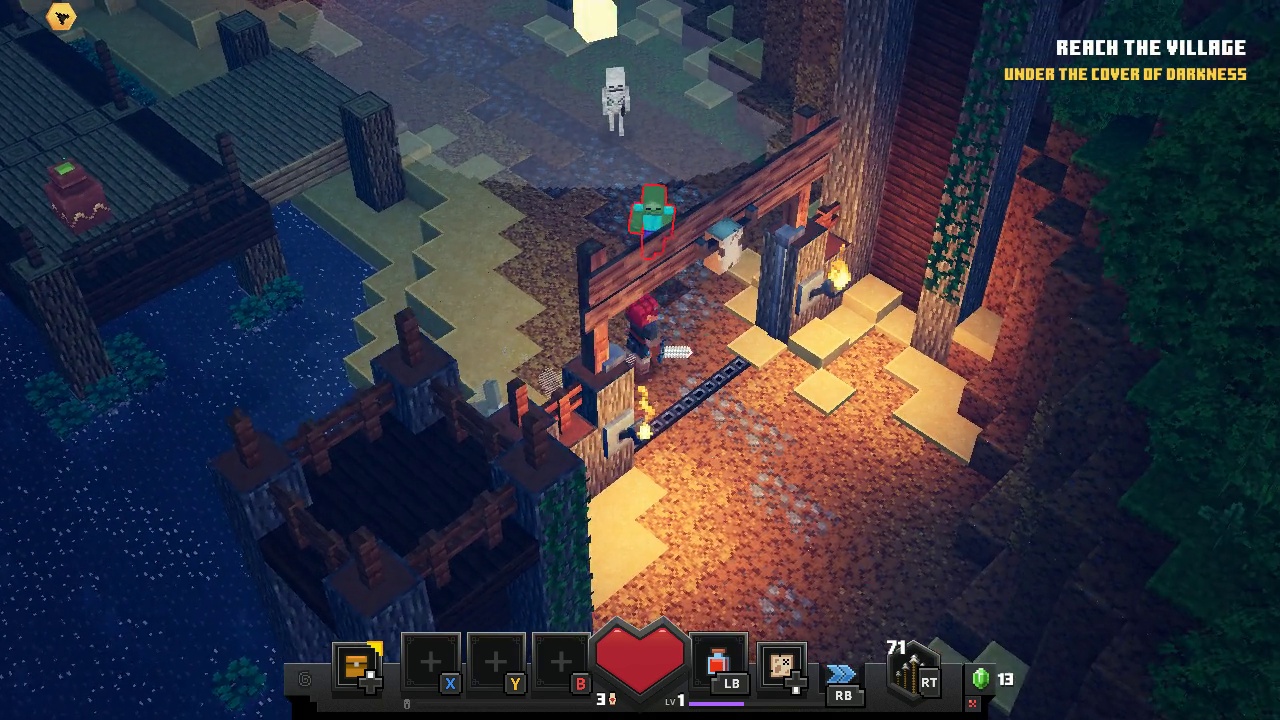}
            \subcaption{Original image}
        \end{minipage}
    \end{adjustbox}
    \hspace{3cm}
    \begin{adjustbox}{valign=t}  
        \begin{minipage}{0.2\linewidth}
            \includegraphics[width=\textwidth,trim={0.5em 0.5em 0.5em 0.5em},clip]{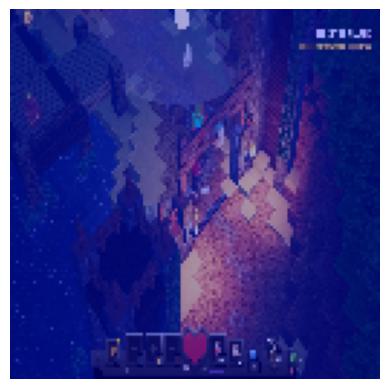}
            \subcaption{Impala ResNet}
        \end{minipage}
    \end{adjustbox}
    \begin{adjustbox}{valign=t}  
        \begin{minipage}{0.2\linewidth}
            \includegraphics[width=\textwidth,trim={0.5em 0.5em 0.5em 0.5em},clip]{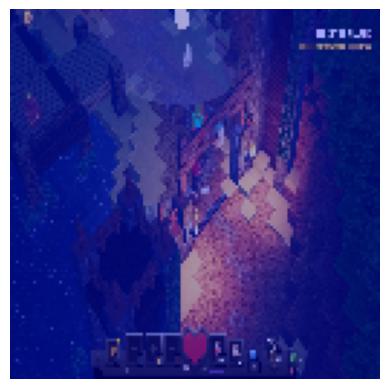}
            \subcaption{Impala ResNet +Aug}
        \end{minipage}
    \end{adjustbox}
    
    \begin{adjustbox}{valign=t}  
        \begin{minipage}{0.19\linewidth}
            \includegraphics[width=\textwidth,trim={0.5em 0.5em 0.5em 0.5em},clip]{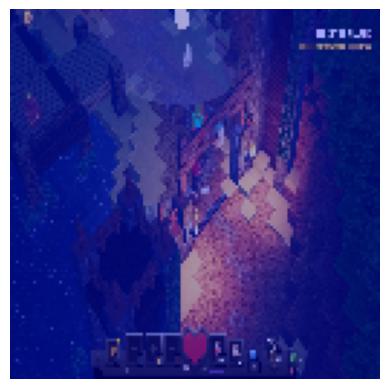}
            \subcaption{ResNet 128}
        \end{minipage}
    \end{adjustbox}
    \begin{adjustbox}{valign=t}  
        \begin{minipage}{0.19\linewidth}
            \includegraphics[width=\textwidth,trim={0.5em 0.5em 0.5em 0.5em},clip]{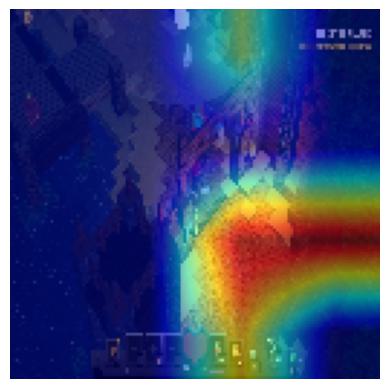}
            \subcaption{ResNet 128 +Aug}
        \end{minipage}
    \end{adjustbox}
    \begin{adjustbox}{valign=t}  
        \begin{minipage}{0.19\linewidth}
            \includegraphics[width=\textwidth,trim={0.5em 0.5em 0.5em 0.5em},clip]{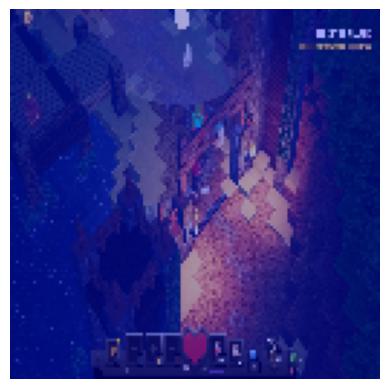}
            \subcaption{ResNet 256}
        \end{minipage}
    \end{adjustbox}
    \begin{adjustbox}{valign=t}  
        \begin{minipage}{0.19\linewidth}
            \includegraphics[width=\textwidth,trim={0.5em 0.5em 0.5em 0.5em},clip]{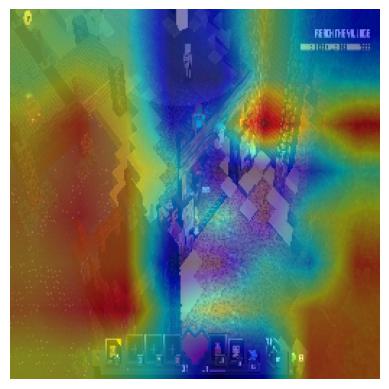}
            \subcaption{ResNet 256 +Aug}
        \end{minipage}
    \end{adjustbox}
    \begin{adjustbox}{valign=t}  
        \begin{minipage}{0.19\linewidth}
            \includegraphics[width=\textwidth,trim={0.5em 0.5em 0.5em 0.5em},clip]{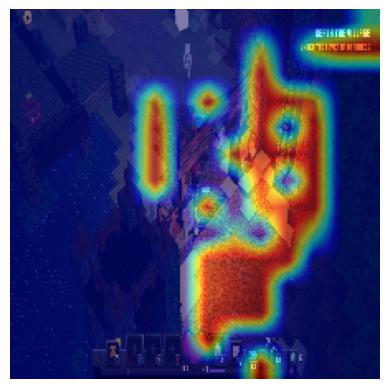}
            \subcaption{ViT Tiny}
        \end{minipage}
    \end{adjustbox}
    
    \begin{adjustbox}{valign=t}  
        \begin{minipage}{0.19\linewidth}
            \includegraphics[width=\textwidth,trim={0.5em 0.5em 0.5em 0.5em},clip]{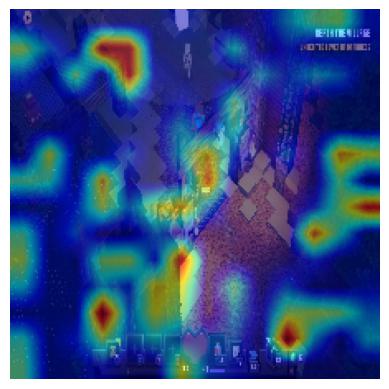}
            \subcaption{ViT Tiny +Aug}
        \end{minipage}
    \end{adjustbox}
    \begin{adjustbox}{valign=t}  
        \begin{minipage}{0.19\linewidth}
            \includegraphics[width=\textwidth,trim={0.5em 0.5em 0.5em 0.5em},clip]{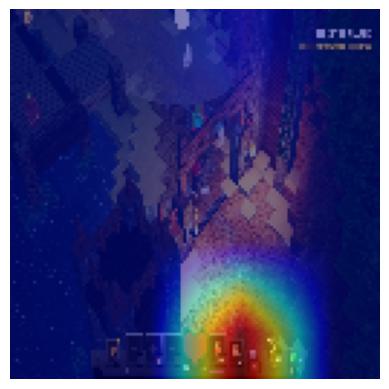}
            \subcaption{ViT 128}
        \end{minipage}
    \end{adjustbox}
    \begin{adjustbox}{valign=t}  
        \begin{minipage}{0.19\linewidth}
            \includegraphics[width=\textwidth,trim={0.5em 0.5em 0.5em 0.5em},clip]{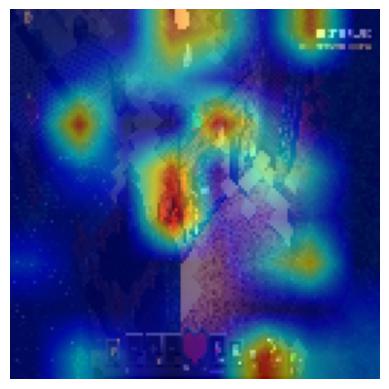}
            \subcaption{ViT 128 +Aug}
        \end{minipage}
    \end{adjustbox}
    \begin{adjustbox}{valign=t}  
        \begin{minipage}{0.19\linewidth}
            \includegraphics[width=\textwidth,trim={0.5em 0.5em 0.5em 0.5em},clip]{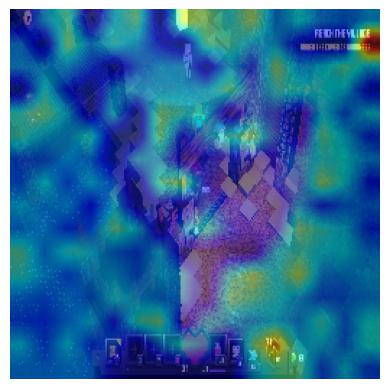}
            \subcaption{ViT 256}
        \end{minipage}
    \end{adjustbox}
    \begin{adjustbox}{valign=t}  
        \begin{minipage}{0.19\linewidth}
            \includegraphics[width=\textwidth,trim={0.5em 0.5em 0.5em 0.5em},clip]{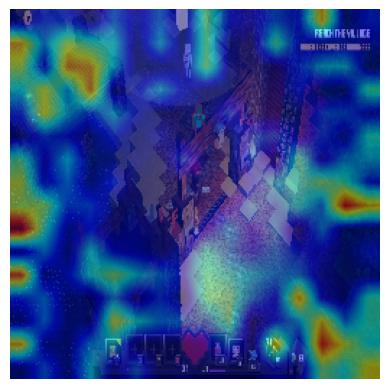}
            \subcaption{ViT 256 +Aug}
        \end{minipage}
    \end{adjustbox}

    \begin{adjustbox}{valign=t}  
        \begin{minipage}{0.19\linewidth}
            \includegraphics[width=\textwidth,trim={0.5em 0.5em 0.5em 0.5em},clip]{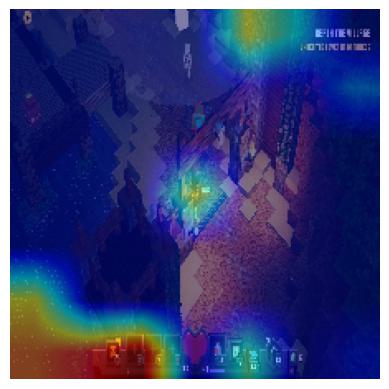}
            \subcaption{CLIP RN50}
        \end{minipage}
    \end{adjustbox}
    \begin{adjustbox}{valign=t}  
        \begin{minipage}{0.19\linewidth}
            \includegraphics[width=\textwidth,trim={0.5em 0.5em 0.5em 0.5em},clip]{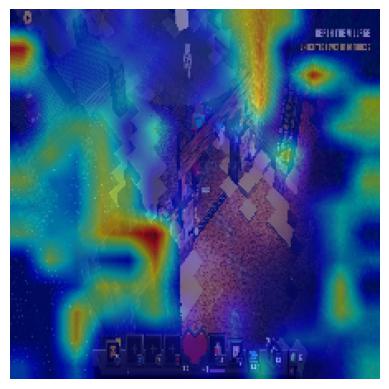}
            \subcaption{CLIP ViT-B/16}
        \end{minipage}
    \end{adjustbox}
    \begin{adjustbox}{valign=t}  
        \begin{minipage}{0.19\linewidth}
            \includegraphics[width=\textwidth,trim={0.5em 0.5em 0.5em 0.5em},clip]{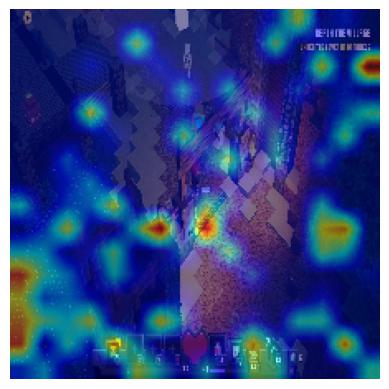}
            \subcaption{CLIP ViT-L/14}
        \end{minipage}
    \end{adjustbox}
    \begin{adjustbox}{valign=t}  
        \begin{minipage}{0.19\linewidth}
            \includegraphics[width=\textwidth,trim={0.5em 0.5em 0.5em 0.5em},clip]{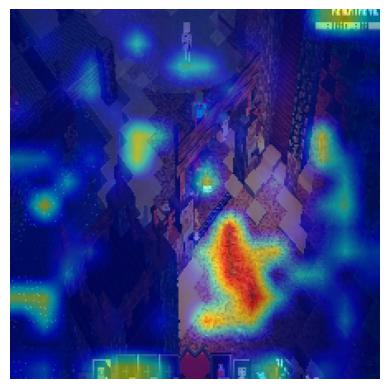}
            \subcaption{DINOv2 ViT-S/14}
        \end{minipage}
    \end{adjustbox}
    \begin{adjustbox}{valign=t}  
        \begin{minipage}{0.19\linewidth}
            \includegraphics[width=\textwidth,trim={0.5em 0.5em 0.5em 0.5em},clip]{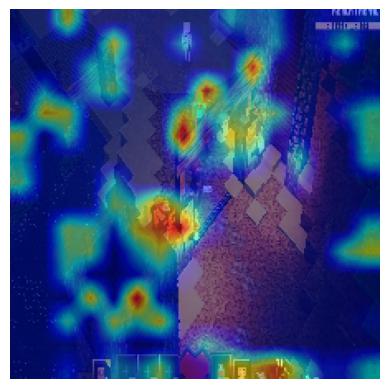}
            \subcaption{DINOv2 ViT-B/14}
        \end{minipage}
    \end{adjustbox}

    \begin{adjustbox}{valign=t}  
        \begin{minipage}{0.19\linewidth}
            \includegraphics[width=\textwidth,trim={0.5em 0.5em 0.5em 0.5em},clip]{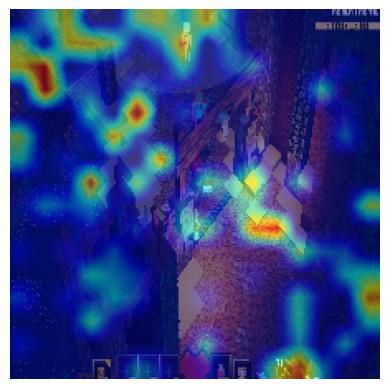}
            \subcaption{DINOv2 ViT-L/14}
        \end{minipage}
    \end{adjustbox}
    \begin{adjustbox}{valign=t}  
        \begin{minipage}{0.19\linewidth}
            \includegraphics[width=\textwidth,trim={0.5em 0.5em 0.5em 0.5em},clip]{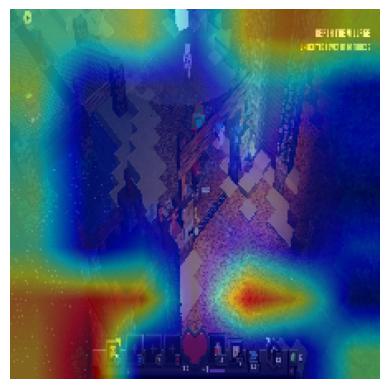}
            \subcaption{Focal Large}
        \end{minipage}
    \end{adjustbox}
    \begin{adjustbox}{valign=t}  
        \begin{minipage}{0.19\linewidth}
            \includegraphics[width=\textwidth,trim={0.5em 0.5em 0.5em 0.5em},clip]{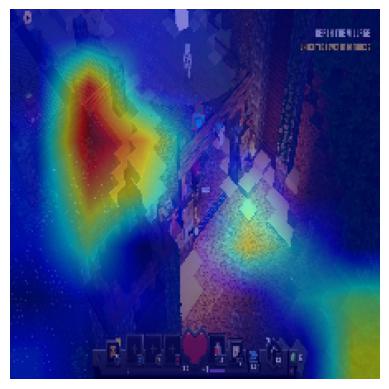}
            \subcaption{Focal XLarge}
        \end{minipage}
    \end{adjustbox}
    \begin{adjustbox}{valign=t}  
        \begin{minipage}{0.19\linewidth}
            \includegraphics[width=\textwidth,trim={0.5em 0.5em 0.5em 0.5em},clip]{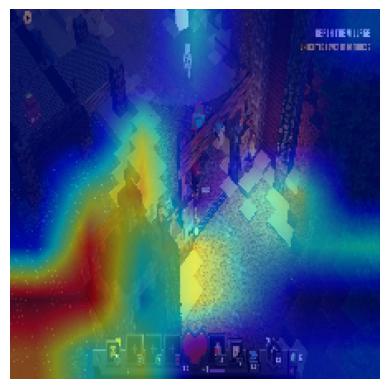}
            \subcaption{Focal Huge}
        \end{minipage}
    \end{adjustbox}
    \begin{adjustbox}{valign=t}  
        \begin{minipage}{0.19\linewidth}
            \includegraphics[width=\textwidth,trim={0.5em 0.5em 0.5em 0.5em},clip]{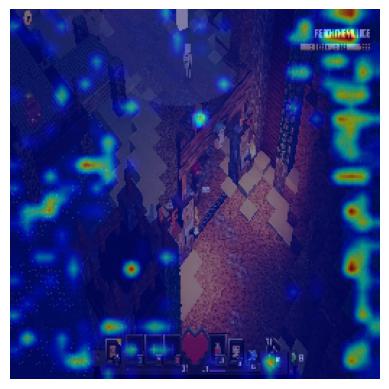}
            \subcaption{SD VAE}
        \end{minipage}
    \end{adjustbox}

    \caption{Grad-Cam visualisations for all encoders (seed 0) in Minecraft Dungeons with policy action logits serving as the targets.}
    \label{fig:grad_cam_dungeons_frame_900_actions}
\end{figure*}

\begin{figure*}[h]
    \centering
    \begin{adjustbox}{valign=t}  
        \begin{minipage}{0.335\linewidth} 
            \centering
            \includegraphics[width=\textwidth]{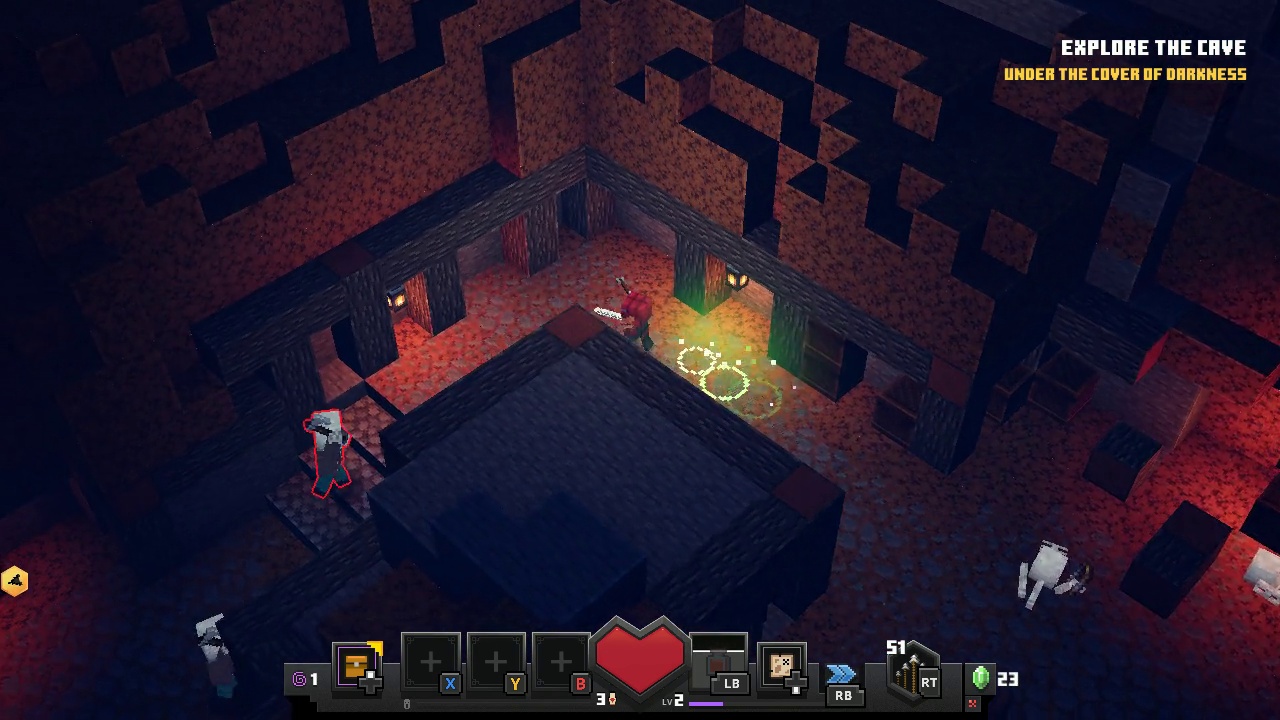}
            \subcaption{Original image}
        \end{minipage}
    \end{adjustbox}
    \hspace{3cm}
    \begin{adjustbox}{valign=t}  
        \begin{minipage}{0.2\linewidth}
            \includegraphics[width=\textwidth,trim={0.5em 0.5em 0.5em 0.5em},clip]{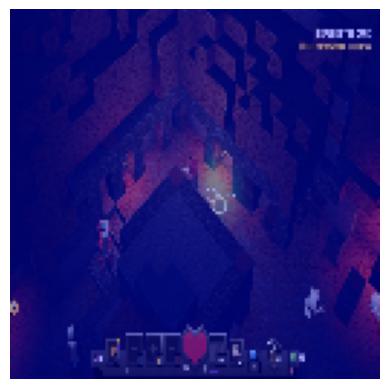}
            \subcaption{Impala ResNet}
        \end{minipage}
    \end{adjustbox}
    \begin{adjustbox}{valign=t}  
        \begin{minipage}{0.2\linewidth}
            \includegraphics[width=\textwidth,trim={0.5em 0.5em 0.5em 0.5em},clip]{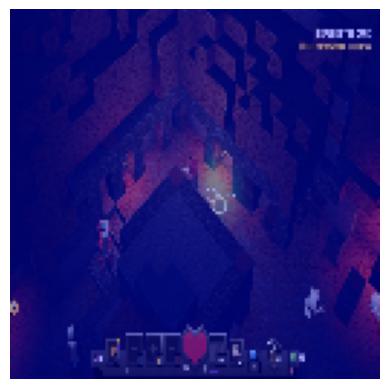}
            \subcaption{Impala ResNet +Aug}
        \end{minipage}
    \end{adjustbox}
    
    \begin{adjustbox}{valign=t}  
        \begin{minipage}{0.19\linewidth}
            \includegraphics[width=\textwidth,trim={0.5em 0.5em 0.5em 0.5em},clip]{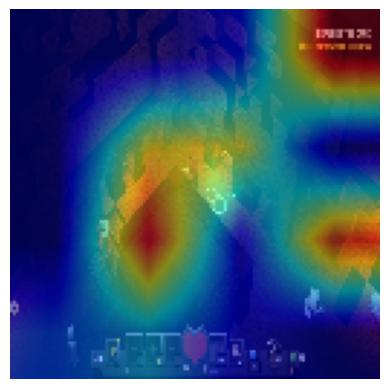}
            \subcaption{ResNet 128}
        \end{minipage}
    \end{adjustbox}
    \begin{adjustbox}{valign=t}  
        \begin{minipage}{0.19\linewidth}
            \includegraphics[width=\textwidth,trim={0.5em 0.5em 0.5em 0.5em},clip]{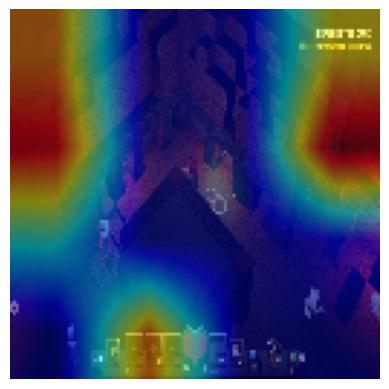}
            \subcaption{ResNet 128 +Aug}
        \end{minipage}
    \end{adjustbox}
    \begin{adjustbox}{valign=t}  
        \begin{minipage}{0.19\linewidth}
            \includegraphics[width=\textwidth,trim={0.5em 0.5em 0.5em 0.5em},clip]{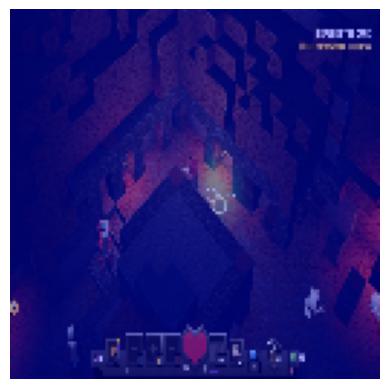}
            \subcaption{ResNet 256}
        \end{minipage}
    \end{adjustbox}
    \begin{adjustbox}{valign=t}  
        \begin{minipage}{0.19\linewidth}
            \includegraphics[width=\textwidth,trim={0.5em 0.5em 0.5em 0.5em},clip]{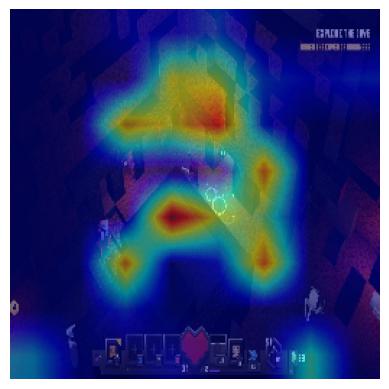}
            \subcaption{ResNet 256 +Aug}
        \end{minipage}
    \end{adjustbox}
    \begin{adjustbox}{valign=t}  
        \begin{minipage}{0.19\linewidth}
            \includegraphics[width=\textwidth,trim={0.5em 0.5em 0.5em 0.5em},clip]{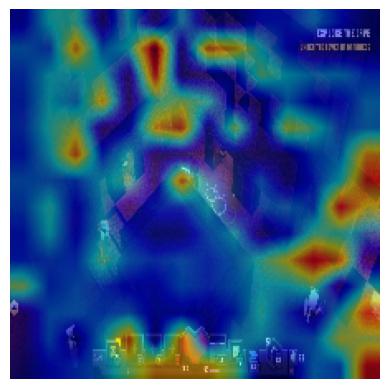}
            \subcaption{ViT Tiny}
        \end{minipage}
    \end{adjustbox}
    
    \begin{adjustbox}{valign=t}  
        \begin{minipage}{0.19\linewidth}
            \includegraphics[width=\textwidth,trim={0.5em 0.5em 0.5em 0.5em},clip]{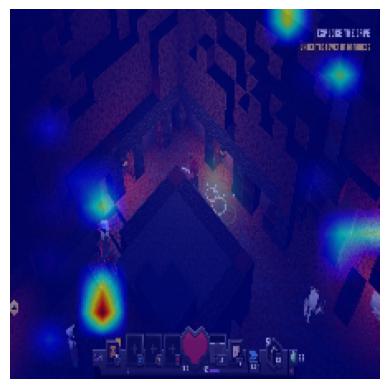}
            \subcaption{ViT Tiny +Aug}
        \end{minipage}
    \end{adjustbox}
    \begin{adjustbox}{valign=t}  
        \begin{minipage}{0.19\linewidth}
            \includegraphics[width=\textwidth,trim={0.5em 0.5em 0.5em 0.5em},clip]{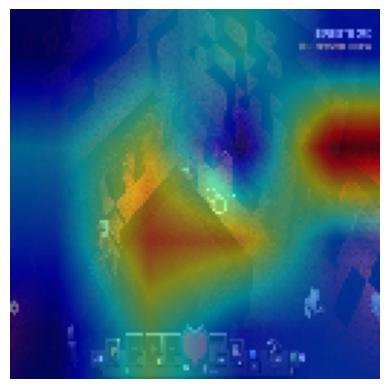}
            \subcaption{ViT 128}
        \end{minipage}
    \end{adjustbox}
    \begin{adjustbox}{valign=t}  
        \begin{minipage}{0.19\linewidth}
            \includegraphics[width=\textwidth,trim={0.5em 0.5em 0.5em 0.5em},clip]{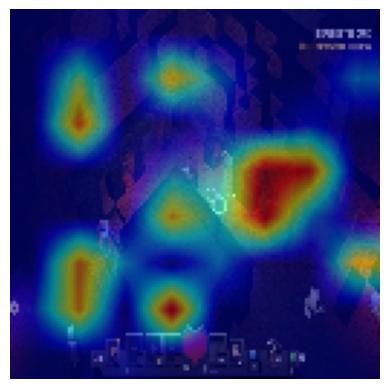}
            \subcaption{ViT 128 +Aug}
        \end{minipage}
    \end{adjustbox}
    \begin{adjustbox}{valign=t}  
        \begin{minipage}{0.19\linewidth}
            \includegraphics[width=\textwidth,trim={0.5em 0.5em 0.5em 0.5em},clip]{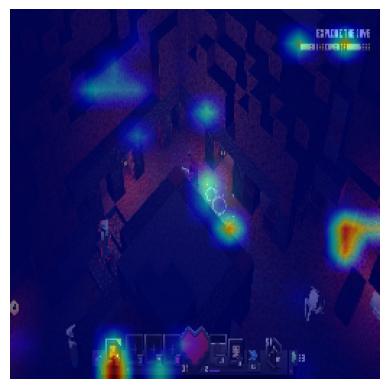}
            \subcaption{ViT 256}
        \end{minipage}
    \end{adjustbox}
    \begin{adjustbox}{valign=t}  
        \begin{minipage}{0.19\linewidth}
            \includegraphics[width=\textwidth,trim={0.5em 0.5em 0.5em 0.5em},clip]{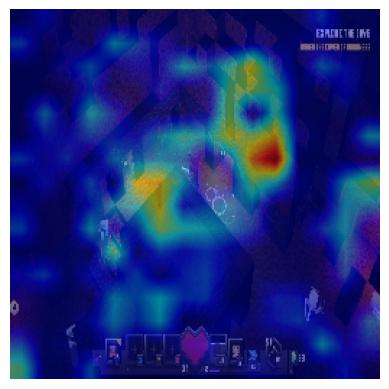}
            \subcaption{ViT 256 +Aug}
        \end{minipage}
    \end{adjustbox}

    \begin{adjustbox}{valign=t}  
        \begin{minipage}{0.19\linewidth}
            \includegraphics[width=\textwidth,trim={0.5em 0.5em 0.5em 0.5em},clip]{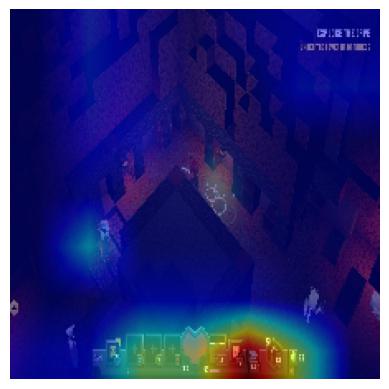}
            \subcaption{CLIP RN50}
        \end{minipage}
    \end{adjustbox}
    \begin{adjustbox}{valign=t}  
        \begin{minipage}{0.19\linewidth}
            \includegraphics[width=\textwidth,trim={0.5em 0.5em 0.5em 0.5em},clip]{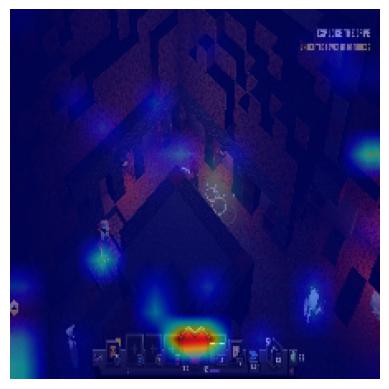}
            \subcaption{CLIP ViT-B/16}
        \end{minipage}
    \end{adjustbox}
    \begin{adjustbox}{valign=t}  
        \begin{minipage}{0.19\linewidth}
            \includegraphics[width=\textwidth,trim={0.5em 0.5em 0.5em 0.5em},clip]{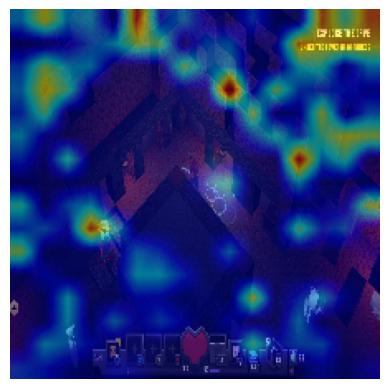}
            \subcaption{CLIP ViT-L/14}
        \end{minipage}
    \end{adjustbox}
    \begin{adjustbox}{valign=t}  
        \begin{minipage}{0.19\linewidth}
            \includegraphics[width=\textwidth,trim={0.5em 0.5em 0.5em 0.5em},clip]{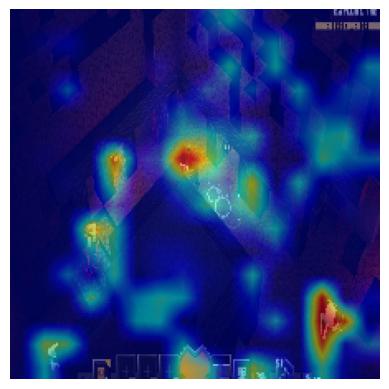}
            \subcaption{DINOv2 ViT-S/14}
        \end{minipage}
    \end{adjustbox}
    \begin{adjustbox}{valign=t}  
        \begin{minipage}{0.19\linewidth}
            \includegraphics[width=\textwidth,trim={0.5em 0.5em 0.5em 0.5em},clip]{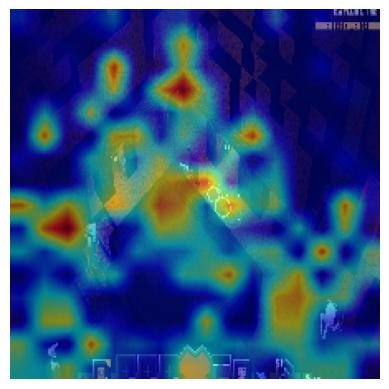}
            \subcaption{DINOv2 ViT-B/14}
        \end{minipage}
    \end{adjustbox}

    \begin{adjustbox}{valign=t}  
        \begin{minipage}{0.19\linewidth}
            \includegraphics[width=\textwidth,trim={0.5em 0.5em 0.5em 0.5em},clip]{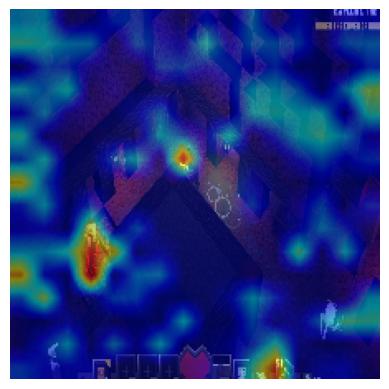}
            \subcaption{DINOv2 ViT-L/14}
        \end{minipage}
    \end{adjustbox}
    \begin{adjustbox}{valign=t}  
        \begin{minipage}{0.19\linewidth}
            \includegraphics[width=\textwidth,trim={0.5em 0.5em 0.5em 0.5em},clip]{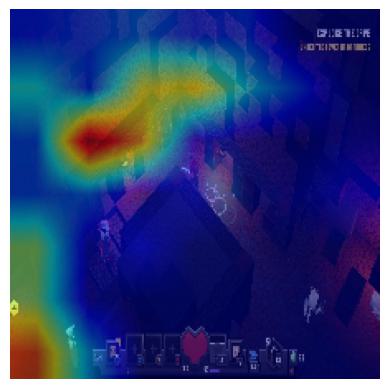}
            \subcaption{Focal Large}
        \end{minipage}
    \end{adjustbox}
    \begin{adjustbox}{valign=t}  
        \begin{minipage}{0.19\linewidth}
            \includegraphics[width=\textwidth,trim={0.5em 0.5em 0.5em 0.5em},clip]{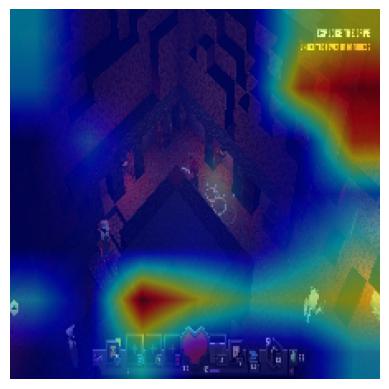}
            \subcaption{Focal XLarge}
        \end{minipage}
    \end{adjustbox}
    \begin{adjustbox}{valign=t}  
        \begin{minipage}{0.19\linewidth}
            \includegraphics[width=\textwidth,trim={0.5em 0.5em 0.5em 0.5em},clip]{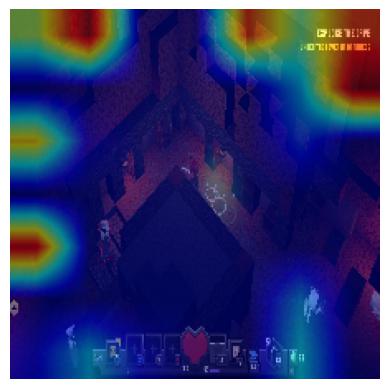}
            \subcaption{Focal Huge}
        \end{minipage}
    \end{adjustbox}
    \begin{adjustbox}{valign=t}  
        \begin{minipage}{0.19\linewidth}
            \includegraphics[width=\textwidth,trim={0.5em 0.5em 0.5em 0.5em},clip]{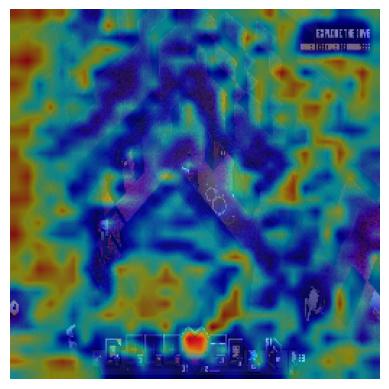}
            \subcaption{SD VAE}
        \end{minipage}
    \end{adjustbox}

    \caption{Grad-Cam visualisations for all encoders (seed 0) in Minecraft Dungeons with policy action logits serving as the targets.}
    \label{fig:grad_cam_dungeons_frame_4200_actions}
\end{figure*}

\begin{figure*}[h]
    \centering
    \begin{adjustbox}{valign=t}  
        \begin{minipage}{0.335\linewidth} 
            \centering
            \includegraphics[width=\textwidth]{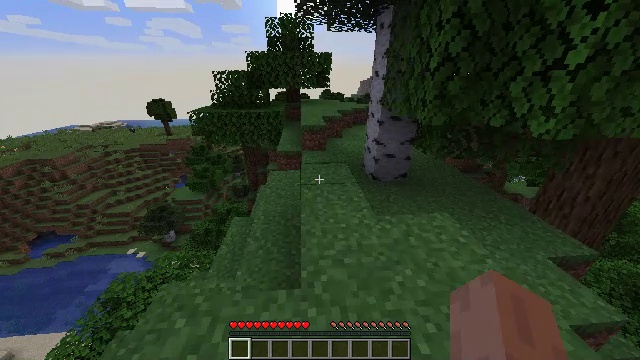}
            \subcaption{Original image}
        \end{minipage}
    \end{adjustbox}
    \hspace{3cm}
    \begin{adjustbox}{valign=t}  
        \begin{minipage}{0.2\linewidth}
            \includegraphics[width=\textwidth,trim={0.5em 0.5em 0.5em 0.5em},clip]{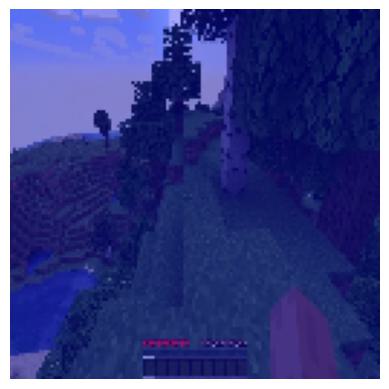}
            \subcaption{Impala ResNet}
        \end{minipage}
    \end{adjustbox}
    \begin{adjustbox}{valign=t}  
        \begin{minipage}{0.2\linewidth}
            \includegraphics[width=\textwidth,trim={0.5em 0.5em 0.5em 0.5em},clip]{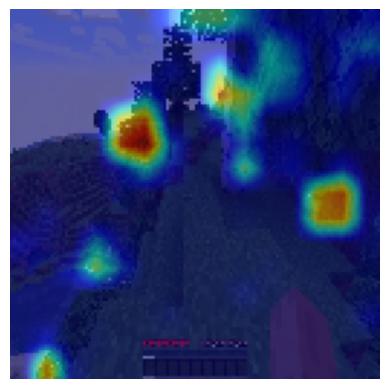}
            \subcaption{Impala ResNet +Aug}
        \end{minipage}
    \end{adjustbox}

    \begin{adjustbox}{valign=t}  
        \begin{minipage}{0.19\linewidth}
            \includegraphics[width=\textwidth,trim={0.5em 0.5em 0.5em 0.5em},clip]{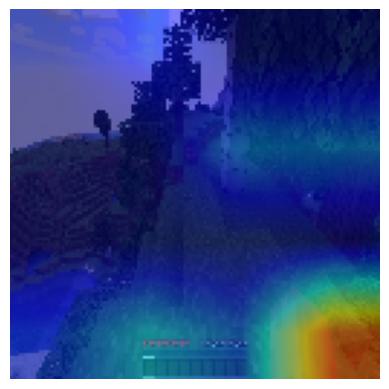}
            \subcaption{ResNet 128}
        \end{minipage}
    \end{adjustbox}
    \begin{adjustbox}{valign=t}  
        \begin{minipage}{0.19\linewidth}
            \includegraphics[width=\textwidth,trim={0.5em 0.5em 0.5em 0.5em},clip]{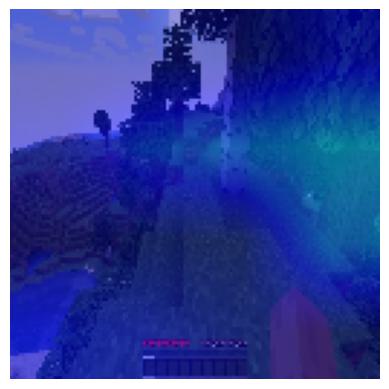}
            \subcaption{ResNet 128 +Aug}
        \end{minipage}
    \end{adjustbox}
    \begin{adjustbox}{valign=t}  
        \begin{minipage}{0.19\linewidth}
            \includegraphics[width=\textwidth,trim={0.5em 0.5em 0.5em 0.5em},clip]{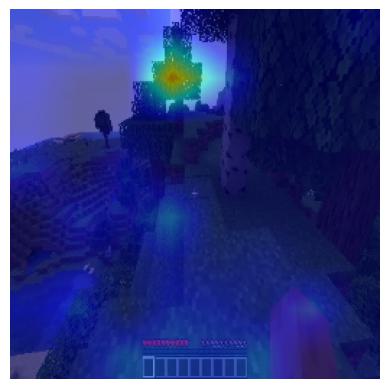}
            \subcaption{ResNet 256}
        \end{minipage}
    \end{adjustbox}
    \begin{adjustbox}{valign=t}  
        \begin{minipage}{0.19\linewidth}
            \includegraphics[width=\textwidth,trim={0.5em 0.5em 0.5em 0.5em},clip]{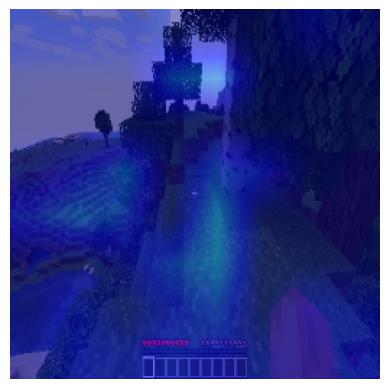}
            \subcaption{ResNet 256 +Aug}
        \end{minipage}
    \end{adjustbox}
    \begin{adjustbox}{valign=t}  
        \begin{minipage}{0.19\linewidth}
            \includegraphics[width=\textwidth,trim={0.5em 0.5em 0.5em 0.5em},clip]{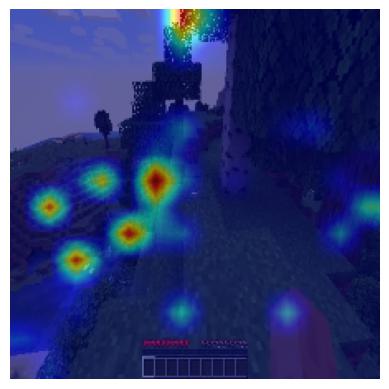}
            \subcaption{ViT Tiny}
        \end{minipage}
    \end{adjustbox}
    
    \begin{adjustbox}{valign=t}  
        \begin{minipage}{0.19\linewidth}
            \includegraphics[width=\textwidth,trim={0.5em 0.5em 0.5em 0.5em},clip]{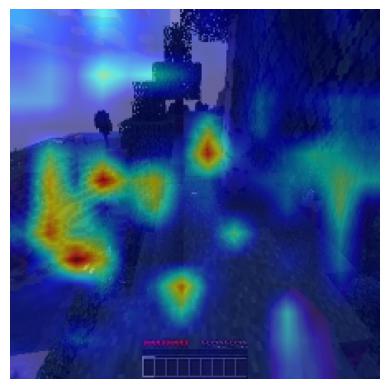}
            \subcaption{ViT Tiny +Aug}
        \end{minipage}
    \end{adjustbox}
    \begin{adjustbox}{valign=t}  
        \begin{minipage}{0.19\linewidth}
            \includegraphics[width=\textwidth,trim={0.5em 0.5em 0.5em 0.5em},clip]{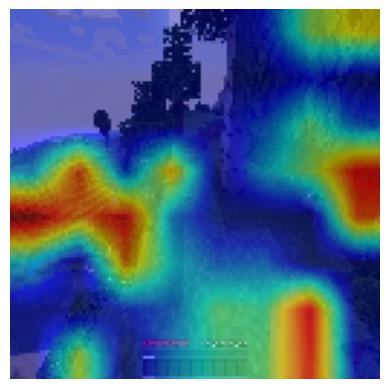}
            \subcaption{ViT 128}
        \end{minipage}
    \end{adjustbox}
    \begin{adjustbox}{valign=t}  
        \begin{minipage}{0.19\linewidth}
            \includegraphics[width=\textwidth,trim={0.5em 0.5em 0.5em 0.5em},clip]{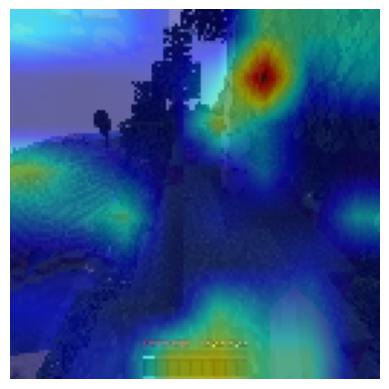}
            \subcaption{ViT 128 +Aug}
        \end{minipage}
    \end{adjustbox}
    \begin{adjustbox}{valign=t}  
        \begin{minipage}{0.19\linewidth}
            \includegraphics[width=\textwidth,trim={0.5em 0.5em 0.5em 0.5em},clip]{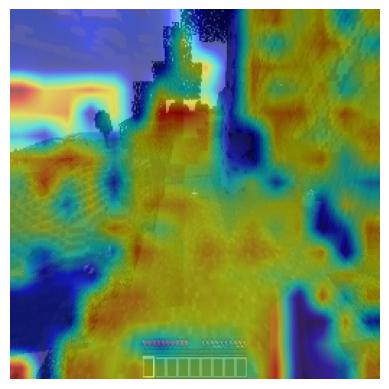}
            \subcaption{ViT 256}
        \end{minipage}
    \end{adjustbox}
    \begin{adjustbox}{valign=t}  
        \begin{minipage}{0.19\linewidth}
            \includegraphics[width=\textwidth,trim={0.5em 0.5em 0.5em 0.5em},clip]{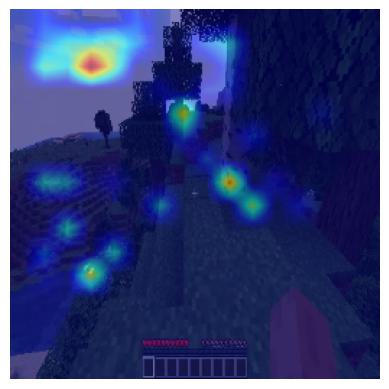}
            \subcaption{ViT 256 +Aug}
        \end{minipage}
    \end{adjustbox}

    \begin{adjustbox}{valign=t}  
        \begin{minipage}{0.19\linewidth}
            \includegraphics[width=\textwidth,trim={0.5em 0.5em 0.5em 0.5em},clip]{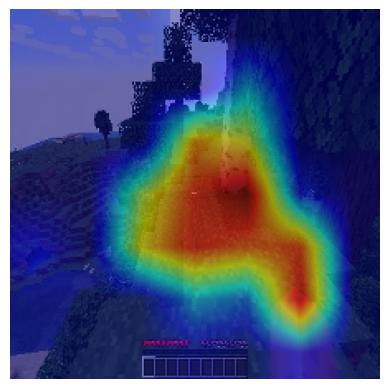}
            \subcaption{CLIP RN50}
        \end{minipage}
    \end{adjustbox}
    \begin{adjustbox}{valign=t}  
        \begin{minipage}{0.19\linewidth}
            \includegraphics[width=\textwidth,trim={0.5em 0.5em 0.5em 0.5em},clip]{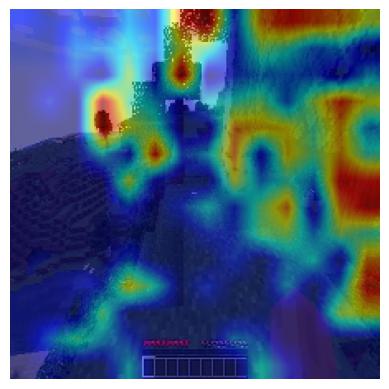}
            \subcaption{CLIP ViT-B/16}
        \end{minipage}
    \end{adjustbox}
    \begin{adjustbox}{valign=t}  
        \begin{minipage}{0.19\linewidth}
            \includegraphics[width=\textwidth,trim={0.5em 0.5em 0.5em 0.5em},clip]{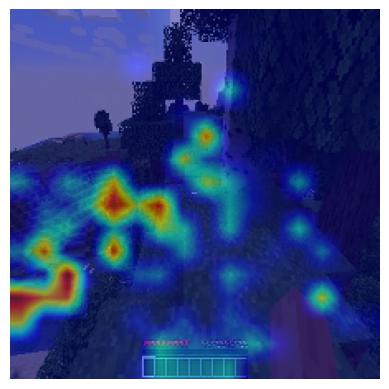}
            \subcaption{CLIP ViT-L/14}
        \end{minipage}
    \end{adjustbox}
    \begin{adjustbox}{valign=t}  
        \begin{minipage}{0.19\linewidth}
            \includegraphics[width=\textwidth,trim={0.5em 0.5em 0.5em 0.5em},clip]{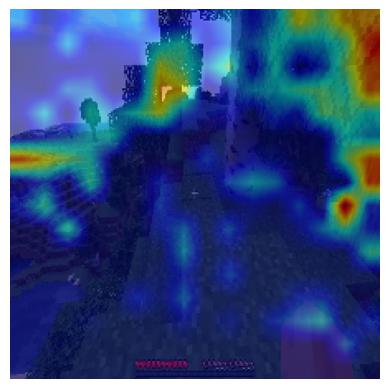}
            \subcaption{DINOv2 ViT-S/14}
        \end{minipage}
    \end{adjustbox}
    \begin{adjustbox}{valign=t}  
        \begin{minipage}{0.19\linewidth}
            \includegraphics[width=\textwidth,trim={0.5em 0.5em 0.5em 0.5em},clip]{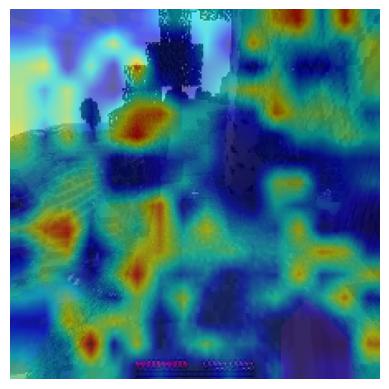}
            \subcaption{DINOv2 ViT-B/14}
        \end{minipage}
    \end{adjustbox}

    \begin{adjustbox}{valign=t}  
        \begin{minipage}{0.19\linewidth}
            \includegraphics[width=\textwidth,trim={0.5em 0.5em 0.5em 0.5em},clip]{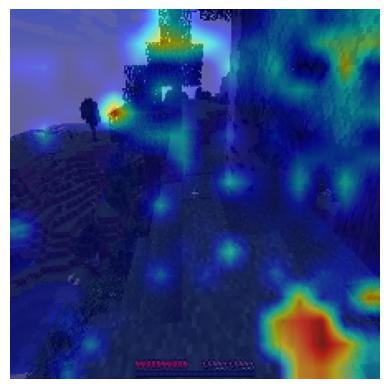}
            \subcaption{DINOv2 ViT-L/14}
        \end{minipage}
    \end{adjustbox}
    \begin{adjustbox}{valign=t}  
        \begin{minipage}{0.19\linewidth}
            \includegraphics[width=\textwidth,trim={0.5em 0.5em 0.5em 0.5em},clip]{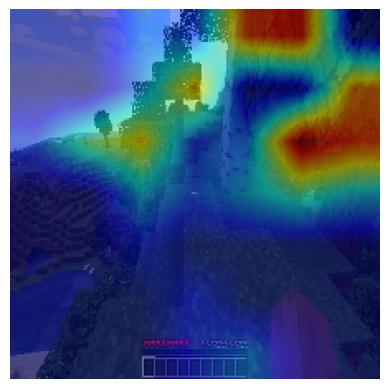}
            \subcaption{Focal Large}
        \end{minipage}
    \end{adjustbox}
    \begin{adjustbox}{valign=t}  
        \begin{minipage}{0.19\linewidth}
            \includegraphics[width=\textwidth,trim={0.5em 0.5em 0.5em 0.5em},clip]{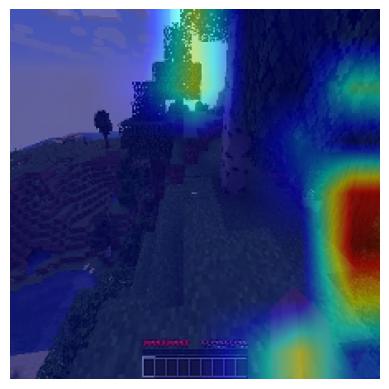}
            \subcaption{Focal XLarge}
        \end{minipage}
    \end{adjustbox}
    \begin{adjustbox}{valign=t}  
        \begin{minipage}{0.19\linewidth}
            \includegraphics[width=\textwidth,trim={0.5em 0.5em 0.5em 0.5em},clip]{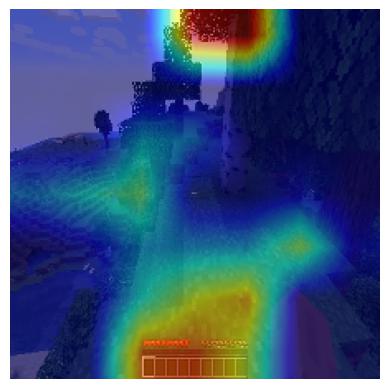}
            \subcaption{Focal Huge}
        \end{minipage}
    \end{adjustbox}
    \begin{adjustbox}{valign=t}  
        \begin{minipage}{0.19\linewidth}
            \includegraphics[width=\textwidth,trim={0.5em 0.5em 0.5em 0.5em},clip]{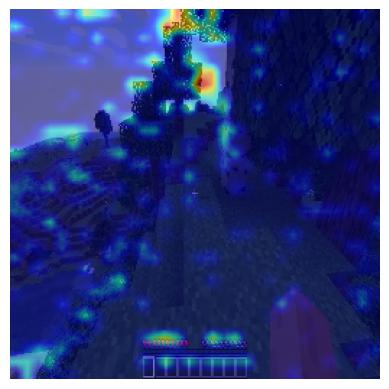}
            \subcaption{SD VAE}
        \end{minipage}
    \end{adjustbox}

    \caption{Grad-Cam visualisations for all encoders (seed 0) in Minecraft with policy action logits serving as the targets.}
    \label{fig:grad_cam_minerl_player112_frame200_actions}
\end{figure*}

\begin{figure*}[h]
    \centering
    \begin{adjustbox}{valign=t}  
        \begin{minipage}{0.335\linewidth} 
            \centering
            \includegraphics[width=\textwidth]{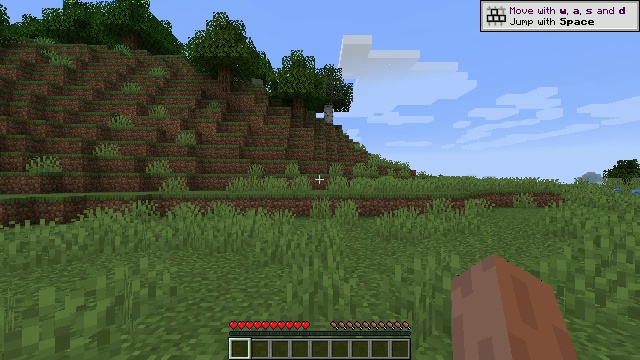}
            \subcaption{Original image}
        \end{minipage}
    \end{adjustbox}
    \hspace{3cm}
    \begin{adjustbox}{valign=t}  
        \begin{minipage}{0.2\linewidth}
            \includegraphics[width=\textwidth,trim={0.5em 0.5em 0.5em 0.5em},clip]{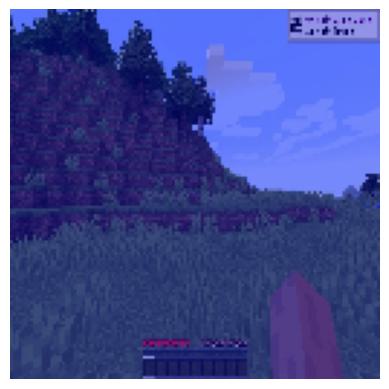}
            \subcaption{Impala ResNet}
        \end{minipage}
    \end{adjustbox}
    \begin{adjustbox}{valign=t}  
        \begin{minipage}{0.2\linewidth}
            \includegraphics[width=\textwidth,trim={0.5em 0.5em 0.5em 0.5em},clip]{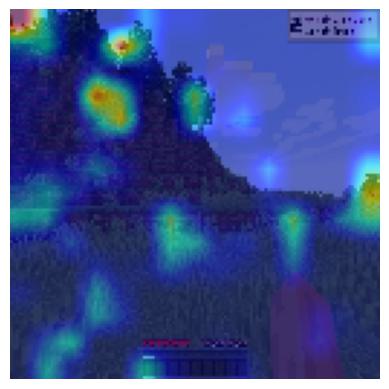}
            \subcaption{Impala ResNet +Aug}
        \end{minipage}
    \end{adjustbox}

    \begin{adjustbox}{valign=t}  
        \begin{minipage}{0.19\linewidth}
            \includegraphics[width=\textwidth,trim={0.5em 0.5em 0.5em 0.5em},clip]{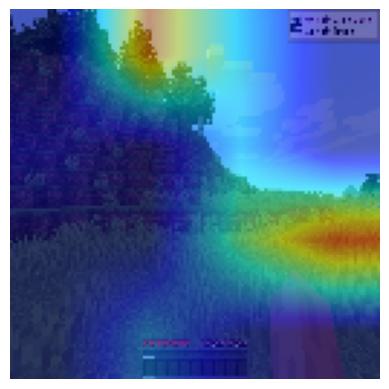}
            \subcaption{ResNet 128}
        \end{minipage}
    \end{adjustbox}
    \begin{adjustbox}{valign=t}  
        \begin{minipage}{0.19\linewidth}
            \includegraphics[width=\textwidth,trim={0.5em 0.5em 0.5em 0.5em},clip]{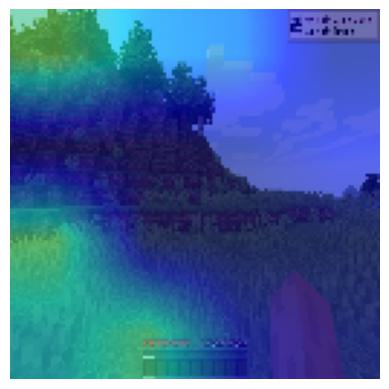}
            \subcaption{ResNet 128 +Aug}
        \end{minipage}
    \end{adjustbox}
    \begin{adjustbox}{valign=t}  
        \begin{minipage}{0.19\linewidth}
            \includegraphics[width=\textwidth,trim={0.5em 0.5em 0.5em 0.5em},clip]{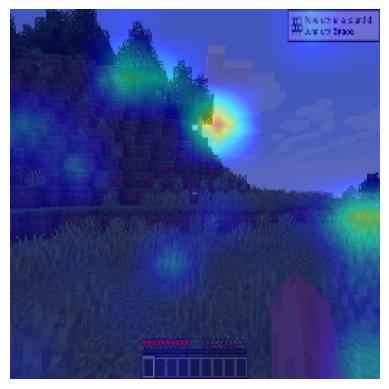}
            \subcaption{ResNet 256}
        \end{minipage}
    \end{adjustbox}
    \begin{adjustbox}{valign=t}  
        \begin{minipage}{0.19\linewidth}
            \includegraphics[width=\textwidth,trim={0.5em 0.5em 0.5em 0.5em},clip]{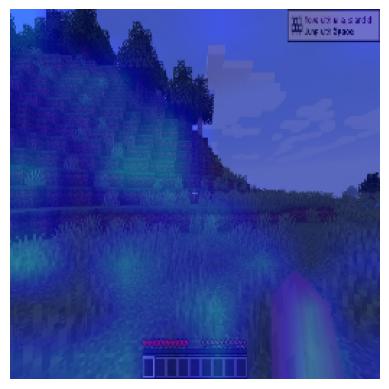}
            \subcaption{ResNet 256 +Aug}
        \end{minipage}
    \end{adjustbox}
    \begin{adjustbox}{valign=t}  
        \begin{minipage}{0.19\linewidth}
            \includegraphics[width=\textwidth,trim={0.5em 0.5em 0.5em 0.5em},clip]{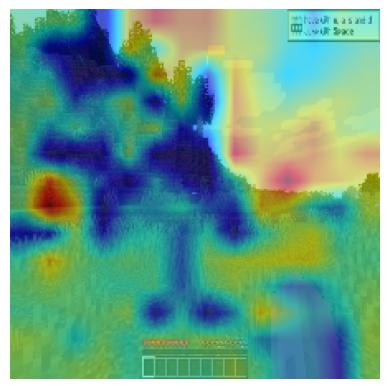}
            \subcaption{ViT Tiny}
        \end{minipage}
    \end{adjustbox}
    
    \begin{adjustbox}{valign=t}  
        \begin{minipage}{0.19\linewidth}
            \includegraphics[width=\textwidth,trim={0.5em 0.5em 0.5em 0.5em},clip]{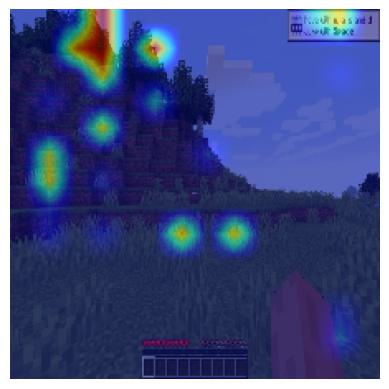}
            \subcaption{ViT Tiny +Aug}
        \end{minipage}
    \end{adjustbox}
    \begin{adjustbox}{valign=t}  
        \begin{minipage}{0.19\linewidth}
            \includegraphics[width=\textwidth,trim={0.5em 0.5em 0.5em 0.5em},clip]{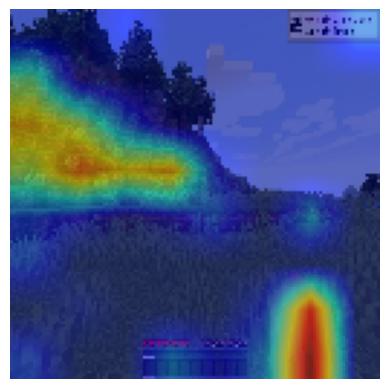}
            \subcaption{ViT 128}
        \end{minipage}
    \end{adjustbox}
    \begin{adjustbox}{valign=t}  
        \begin{minipage}{0.19\linewidth}
            \includegraphics[width=\textwidth,trim={0.5em 0.5em 0.5em 0.5em},clip]{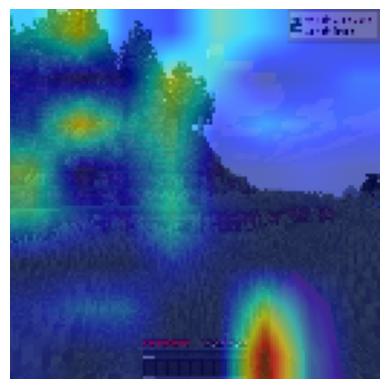}
            \subcaption{ViT 128 +Aug}
        \end{minipage}
    \end{adjustbox}
    \begin{adjustbox}{valign=t}  
        \begin{minipage}{0.19\linewidth}
            \includegraphics[width=\textwidth,trim={0.5em 0.5em 0.5em 0.5em},clip]{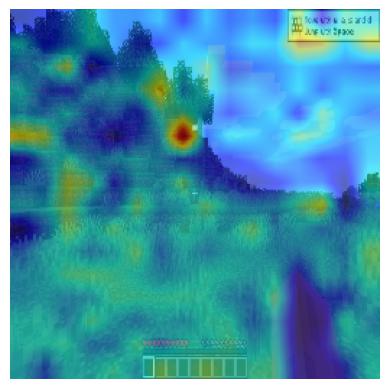}
            \subcaption{ViT 256}
        \end{minipage}
    \end{adjustbox}
    \begin{adjustbox}{valign=t}  
        \begin{minipage}{0.19\linewidth}
            \includegraphics[width=\textwidth,trim={0.5em 0.5em 0.5em 0.5em},clip]{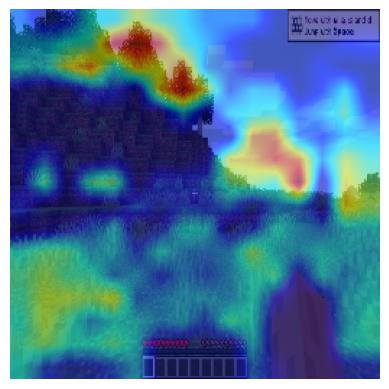}
            \subcaption{ViT 256 +Aug}
        \end{minipage}
    \end{adjustbox}

    \begin{adjustbox}{valign=t}  
        \begin{minipage}{0.19\linewidth}
            \includegraphics[width=\textwidth,trim={0.5em 0.5em 0.5em 0.5em},clip]{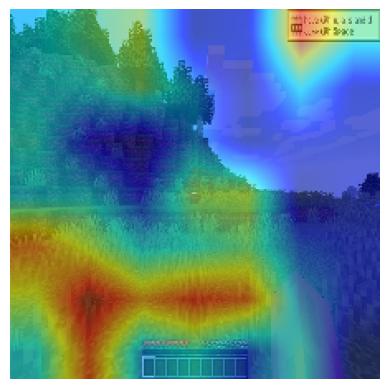}
            \subcaption{CLIP RN50}
        \end{minipage}
    \end{adjustbox}
    \begin{adjustbox}{valign=t}  
        \begin{minipage}{0.19\linewidth}
            \includegraphics[width=\textwidth,trim={0.5em 0.5em 0.5em 0.5em},clip]{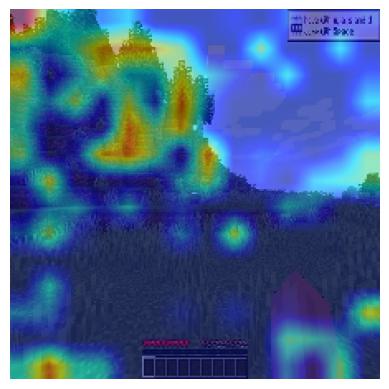}
            \subcaption{CLIP ViT-B/16}
        \end{minipage}
    \end{adjustbox}
    \begin{adjustbox}{valign=t}  
        \begin{minipage}{0.19\linewidth}
            \includegraphics[width=\textwidth,trim={0.5em 0.5em 0.5em 0.5em},clip]{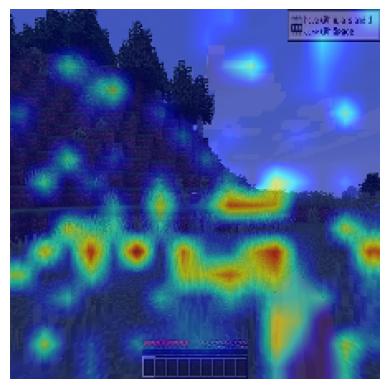}
            \subcaption{CLIP ViT-L/14}
        \end{minipage}
    \end{adjustbox}
    \begin{adjustbox}{valign=t}  
        \begin{minipage}{0.19\linewidth}
            \includegraphics[width=\textwidth,trim={0.5em 0.5em 0.5em 0.5em},clip]{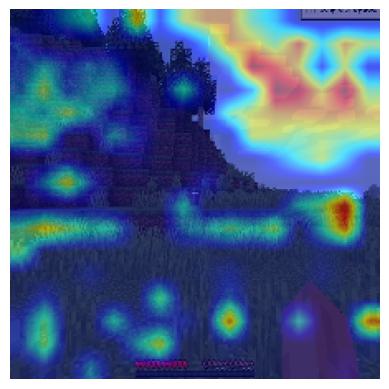}
            \subcaption{DINOv2 ViT-S/14}
        \end{minipage}
    \end{adjustbox}
    \begin{adjustbox}{valign=t}  
        \begin{minipage}{0.19\linewidth}
            \includegraphics[width=\textwidth,trim={0.5em 0.5em 0.5em 0.5em},clip]{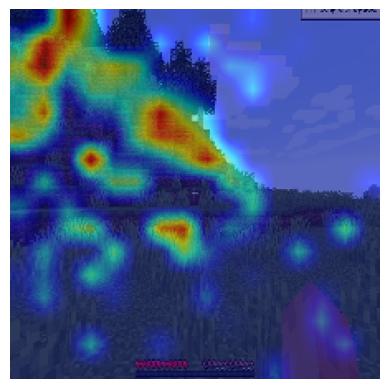}
            \subcaption{DINOv2 ViT-B/14}
        \end{minipage}
    \end{adjustbox}

    \begin{adjustbox}{valign=t}  
        \begin{minipage}{0.19\linewidth}
            \includegraphics[width=\textwidth,trim={0.5em 0.5em 0.5em 0.5em},clip]{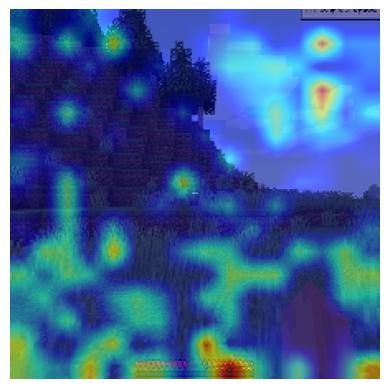}
            \subcaption{DINOv2 ViT-L/14}
        \end{minipage}
    \end{adjustbox}
    \begin{adjustbox}{valign=t}  
        \begin{minipage}{0.19\linewidth}
            \includegraphics[width=\textwidth,trim={0.5em 0.5em 0.5em 0.5em},clip]{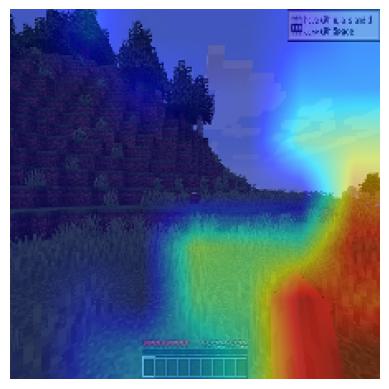}
            \subcaption{Focal Large}
        \end{minipage}
    \end{adjustbox}
    \begin{adjustbox}{valign=t}  
        \begin{minipage}{0.19\linewidth}
            \includegraphics[width=\textwidth,trim={0.5em 0.5em 0.5em 0.5em},clip]{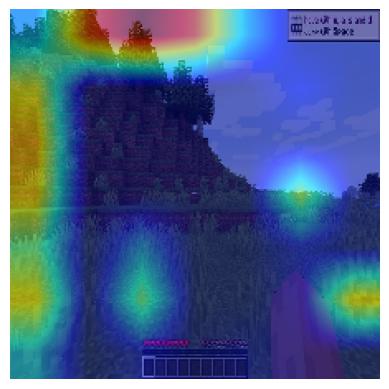}
            \subcaption{Focal XLarge}
        \end{minipage}
    \end{adjustbox}
    \begin{adjustbox}{valign=t}  
        \begin{minipage}{0.19\linewidth}
            \includegraphics[width=\textwidth,trim={0.5em 0.5em 0.5em 0.5em},clip]{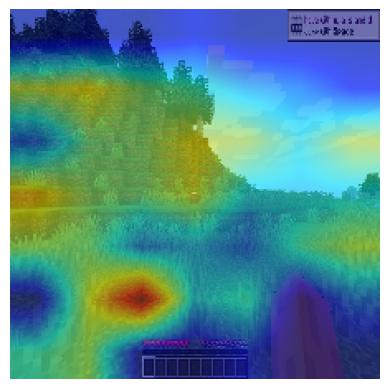}
            \subcaption{Focal Huge}
        \end{minipage}
    \end{adjustbox}
    \begin{adjustbox}{valign=t}  
        \begin{minipage}{0.19\linewidth}
            \includegraphics[width=\textwidth,trim={0.5em 0.5em 0.5em 0.5em},clip]{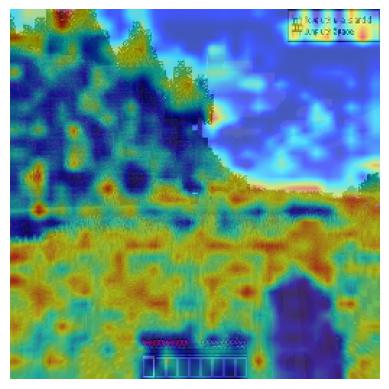}
            \subcaption{SD VAE}
        \end{minipage}
    \end{adjustbox}

    \caption{Grad-Cam visualisations for all encoders (seed 0) in Minecraft with policy action logits serving as the targets.}
    \label{fig:grad_cam_minerl_player123_frame500_actions}
\end{figure*}

\begin{figure*}[h]
    \centering
    \begin{adjustbox}{valign=t}  
        \begin{minipage}{0.335\linewidth} 
            \centering
            \includegraphics[width=\textwidth]{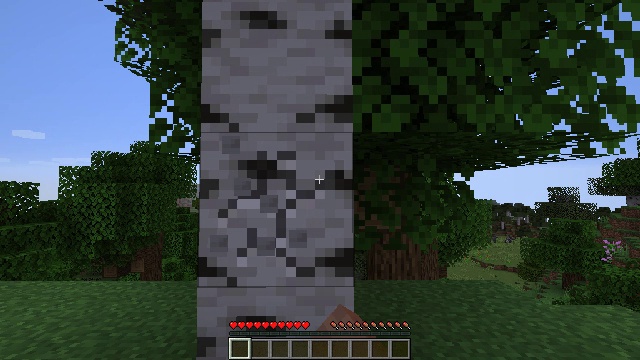}
            \subcaption{Original image}
        \end{minipage}
    \end{adjustbox}
    \hspace{3cm}
    \begin{adjustbox}{valign=t}  
        \begin{minipage}{0.2\linewidth}
            \includegraphics[width=\textwidth,trim={0.5em 0.5em 0.5em 0.5em},clip]{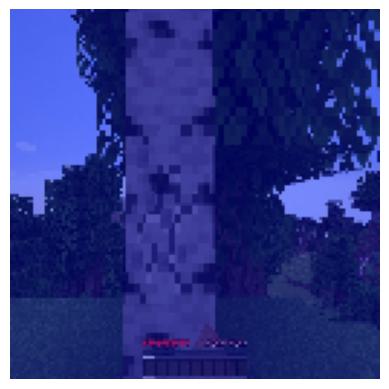}
            \subcaption{Impala ResNet}
        \end{minipage}
    \end{adjustbox}
    \begin{adjustbox}{valign=t}  
        \begin{minipage}{0.2\linewidth}
            \includegraphics[width=\textwidth,trim={0.5em 0.5em 0.5em 0.5em},clip]{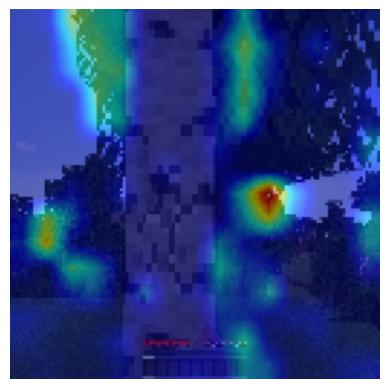}
            \subcaption{Impala ResNet +Aug}
        \end{minipage}
    \end{adjustbox}

    \begin{adjustbox}{valign=t}  
        \begin{minipage}{0.19\linewidth}
            \includegraphics[width=\textwidth,trim={0.5em 0.5em 0.5em 0.5em},clip]{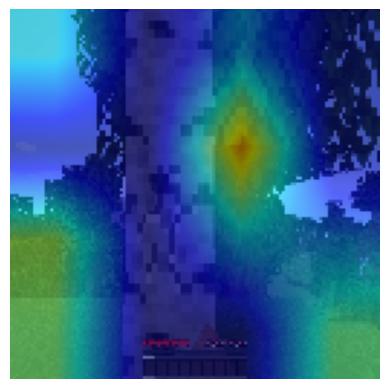}
            \subcaption{ResNet 128}
        \end{minipage}
    \end{adjustbox}
    \begin{adjustbox}{valign=t}  
        \begin{minipage}{0.19\linewidth}
            \includegraphics[width=\textwidth,trim={0.5em 0.5em 0.5em 0.5em},clip]{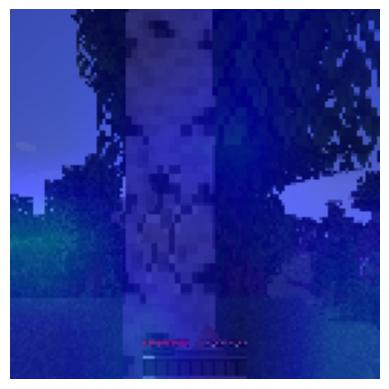}
            \subcaption{ResNet 128 +Aug}
        \end{minipage}
    \end{adjustbox}
    \begin{adjustbox}{valign=t}  
        \begin{minipage}{0.19\linewidth}
            \includegraphics[width=\textwidth,trim={0.5em 0.5em 0.5em 0.5em},clip]{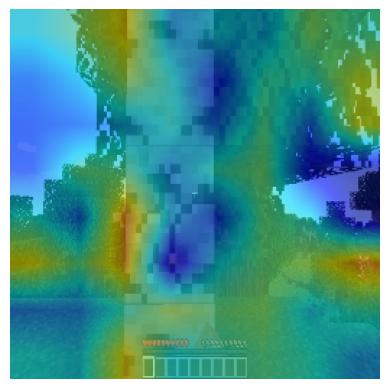}
            \subcaption{ResNet 256}
        \end{minipage}
    \end{adjustbox}
    \begin{adjustbox}{valign=t}  
        \begin{minipage}{0.19\linewidth}
            \includegraphics[width=\textwidth,trim={0.5em 0.5em 0.5em 0.5em},clip]{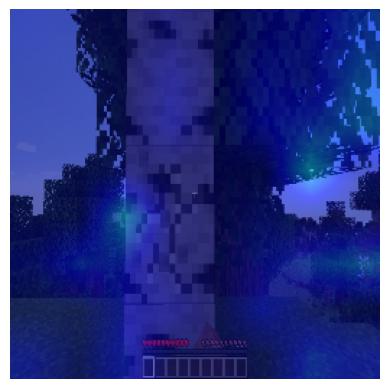}
            \subcaption{ResNet 256 +Aug}
        \end{minipage}
    \end{adjustbox}
    \begin{adjustbox}{valign=t}  
        \begin{minipage}{0.19\linewidth}
            \includegraphics[width=\textwidth,trim={0.5em 0.5em 0.5em 0.5em},clip]{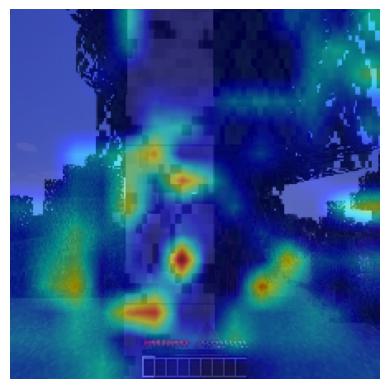}
            \subcaption{ViT Tiny}
        \end{minipage}
    \end{adjustbox}
    
    \begin{adjustbox}{valign=t}  
        \begin{minipage}{0.19\linewidth}
            \includegraphics[width=\textwidth,trim={0.5em 0.5em 0.5em 0.5em},clip]{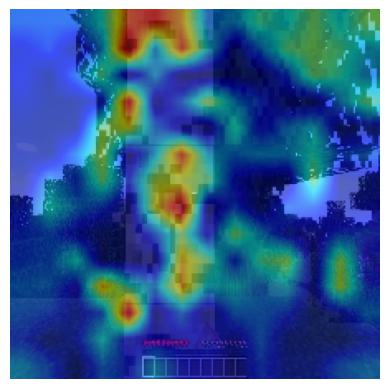}
            \subcaption{ViT Tiny +Aug}
        \end{minipage}
    \end{adjustbox}
    \begin{adjustbox}{valign=t}  
        \begin{minipage}{0.19\linewidth}
            \includegraphics[width=\textwidth,trim={0.5em 0.5em 0.5em 0.5em},clip]{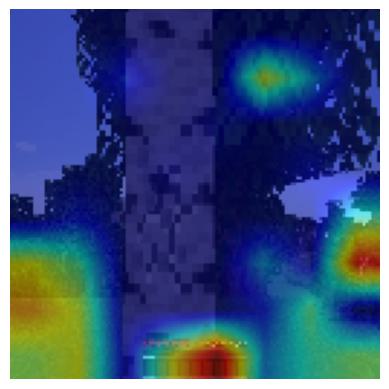}
            \subcaption{ViT 128}
        \end{minipage}
    \end{adjustbox}
    \begin{adjustbox}{valign=t}  
        \begin{minipage}{0.19\linewidth}
            \includegraphics[width=\textwidth,trim={0.5em 0.5em 0.5em 0.5em},clip]{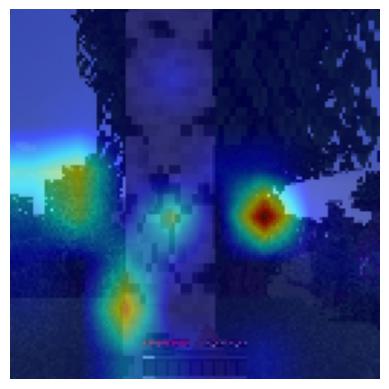}
            \subcaption{ViT 128 +Aug}
        \end{minipage}
    \end{adjustbox}
    \begin{adjustbox}{valign=t}  
        \begin{minipage}{0.19\linewidth}
            \includegraphics[width=\textwidth,trim={0.5em 0.5em 0.5em 0.5em},clip]{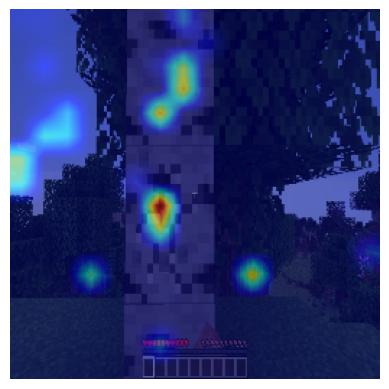}
            \subcaption{ViT 256}
        \end{minipage}
    \end{adjustbox}
    \begin{adjustbox}{valign=t}  
        \begin{minipage}{0.19\linewidth}
            \includegraphics[width=\textwidth,trim={0.5em 0.5em 0.5em 0.5em},clip]{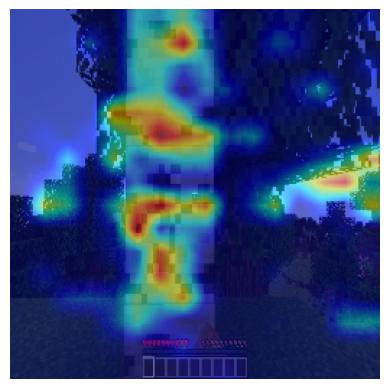}
            \subcaption{ViT 256 +Aug}
        \end{minipage}
    \end{adjustbox}

    \begin{adjustbox}{valign=t}  
        \begin{minipage}{0.19\linewidth}
            \includegraphics[width=\textwidth,trim={0.5em 0.5em 0.5em 0.5em},clip]{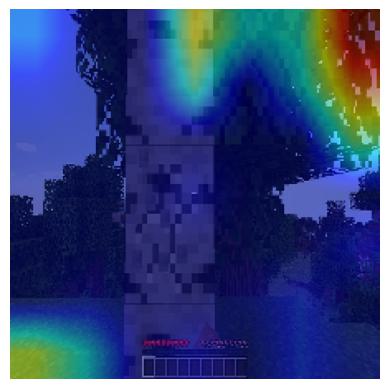}
            \subcaption{CLIP RN50}
        \end{minipage}
    \end{adjustbox}
    \begin{adjustbox}{valign=t}  
        \begin{minipage}{0.19\linewidth}
            \includegraphics[width=\textwidth,trim={0.5em 0.5em 0.5em 0.5em},clip]{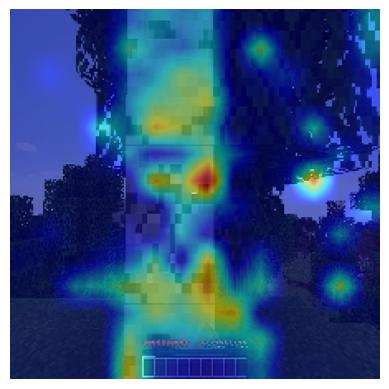}
            \subcaption{CLIP ViT-B/16}
        \end{minipage}
    \end{adjustbox}
    \begin{adjustbox}{valign=t}  
        \begin{minipage}{0.19\linewidth}
            \includegraphics[width=\textwidth,trim={0.5em 0.5em 0.5em 0.5em},clip]{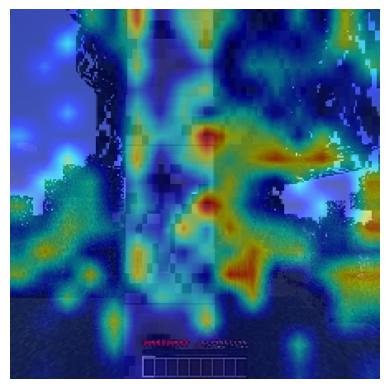}
            \subcaption{CLIP ViT-L/14}
        \end{minipage}
    \end{adjustbox}
    \begin{adjustbox}{valign=t}  
        \begin{minipage}{0.19\linewidth}
            \includegraphics[width=\textwidth,trim={0.5em 0.5em 0.5em 0.5em},clip]{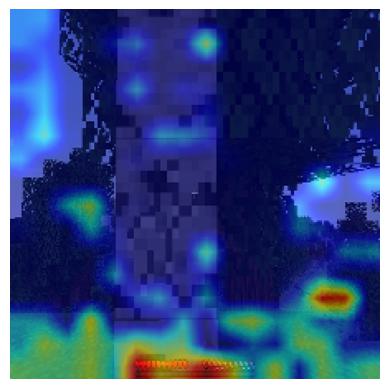}
            \subcaption{DINOv2 ViT-S/14}
        \end{minipage}
    \end{adjustbox}
    \begin{adjustbox}{valign=t}  
        \begin{minipage}{0.19\linewidth}
            \includegraphics[width=\textwidth,trim={0.5em 0.5em 0.5em 0.5em},clip]{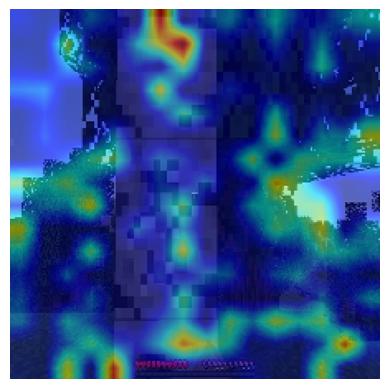}
            \subcaption{DINOv2 ViT-B/14}
        \end{minipage}
    \end{adjustbox}

    \begin{adjustbox}{valign=t}  
        \begin{minipage}{0.19\linewidth}
            \includegraphics[width=\textwidth,trim={0.5em 0.5em 0.5em 0.5em},clip]{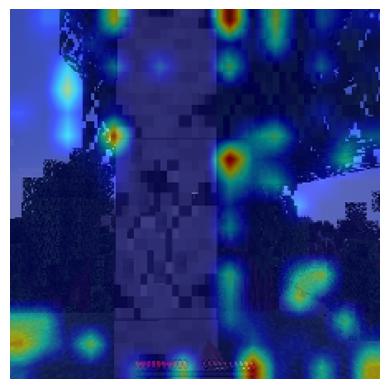}
            \subcaption{DINOv2 ViT-L/14}
        \end{minipage}
    \end{adjustbox}
    \begin{adjustbox}{valign=t}  
        \begin{minipage}{0.19\linewidth}
            \includegraphics[width=\textwidth,trim={0.5em 0.5em 0.5em 0.5em},clip]{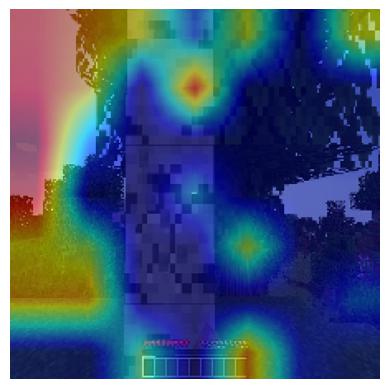}
            \subcaption{Focal Large}
        \end{minipage}
    \end{adjustbox}
    \begin{adjustbox}{valign=t}  
        \begin{minipage}{0.19\linewidth}
            \includegraphics[width=\textwidth,trim={0.5em 0.5em 0.5em 0.5em},clip]{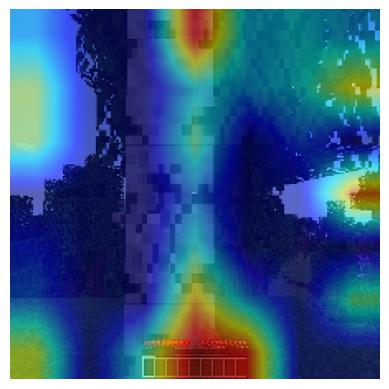}
            \subcaption{Focal XLarge}
        \end{minipage}
    \end{adjustbox}
    \begin{adjustbox}{valign=t}  
        \begin{minipage}{0.19\linewidth}
            \includegraphics[width=\textwidth,trim={0.5em 0.5em 0.5em 0.5em},clip]{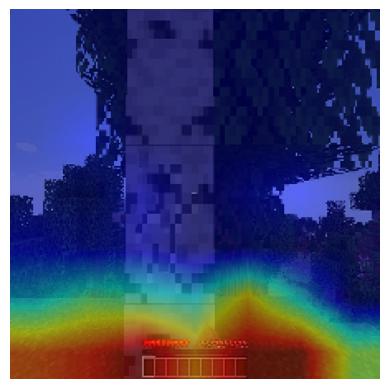}
            \subcaption{Focal Huge}
        \end{minipage}
    \end{adjustbox}
    \begin{adjustbox}{valign=t}  
        \begin{minipage}{0.19\linewidth}
            \includegraphics[width=\textwidth,trim={0.5em 0.5em 0.5em 0.5em},clip]{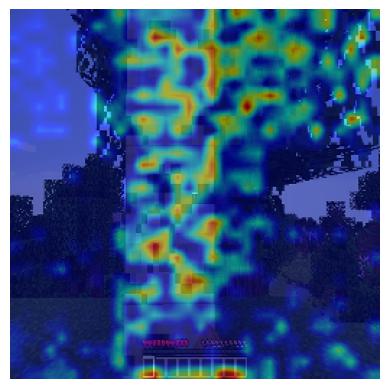}
            \subcaption{SD VAE}
        \end{minipage}
    \end{adjustbox}

    \caption{Grad-Cam visualisations for all encoders (seed 0) in Minecraft with policy action logits serving as the targets.}
    \label{fig:grad_cam_minerl_player112_frame_300_actions}
\end{figure*}

\begin{figure*}[h]
    \centering
    \begin{adjustbox}{valign=t}  
        \begin{minipage}{0.335\linewidth} 
            \centering
            \includegraphics[width=\textwidth]{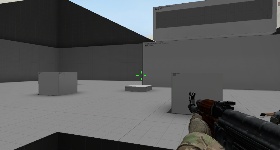}
            \subcaption{Original image}
        \end{minipage}
    \end{adjustbox}
    \hspace{3cm}
    \begin{adjustbox}{valign=t}  
        \begin{minipage}{0.2\linewidth}
            \includegraphics[width=\textwidth,trim={0.5em 0.5em 0.5em 0.5em},clip]{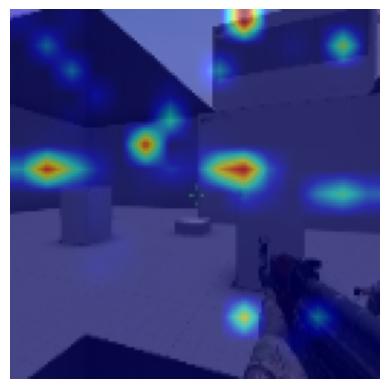}
            \subcaption{Impala ResNet}
        \end{minipage}
    \end{adjustbox}
    \begin{adjustbox}{valign=t}  
        \begin{minipage}{0.2\linewidth}
            \includegraphics[width=\textwidth,trim={0.5em 0.5em 0.5em 0.5em},clip]{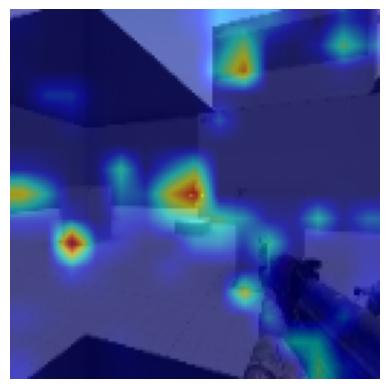}
            \subcaption{Impala ResNet +Aug}
        \end{minipage}
    \end{adjustbox}

    \begin{adjustbox}{valign=t}  
        \begin{minipage}{0.19\linewidth}
            \includegraphics[width=\textwidth,trim={0.5em 0.5em 0.5em 0.5em},clip]{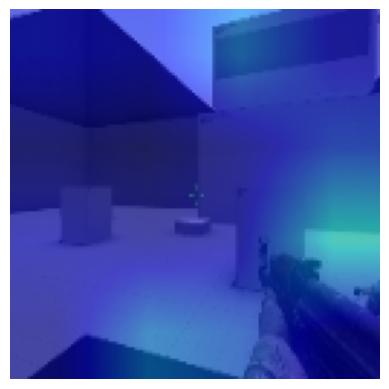}
            \subcaption{ResNet 128}
        \end{minipage}
    \end{adjustbox}
    \begin{adjustbox}{valign=t}  
        \begin{minipage}{0.19\linewidth}
            \includegraphics[width=\textwidth,trim={0.5em 0.5em 0.5em 0.5em},clip]{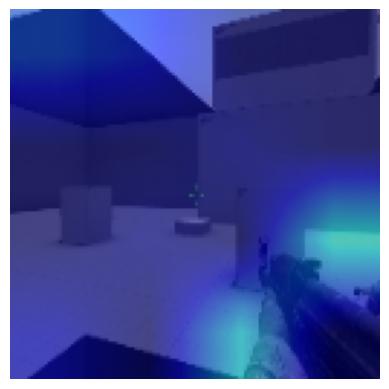}
            \subcaption{ResNet 128 +Aug}
        \end{minipage}
    \end{adjustbox}
    \begin{adjustbox}{valign=t}  
        \begin{minipage}{0.19\linewidth}
            \includegraphics[width=\textwidth,trim={0.5em 0.5em 0.5em 0.5em},clip]{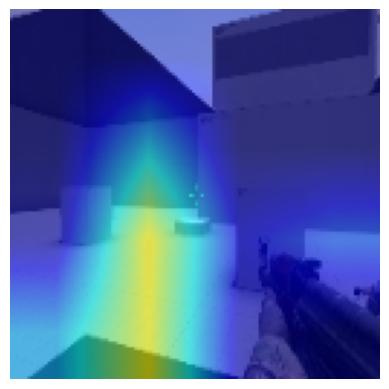}
            \subcaption{ResNet 256}
        \end{minipage}
    \end{adjustbox}
    \begin{adjustbox}{valign=t}  
        \begin{minipage}{0.19\linewidth}
            \includegraphics[width=\textwidth,trim={0.5em 0.5em 0.5em 0.5em},clip]{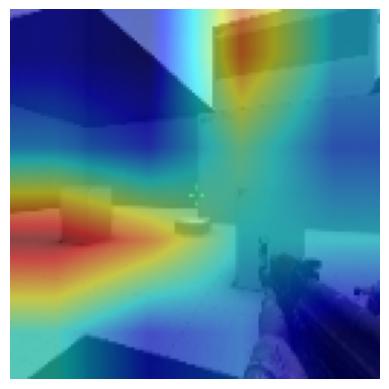}
            \subcaption{ResNet 256 +Aug}
        \end{minipage}
    \end{adjustbox}
    \begin{adjustbox}{valign=t}  
        \begin{minipage}{0.19\linewidth}
            \includegraphics[width=\textwidth,trim={0.5em 0.5em 0.5em 0.5em},clip]{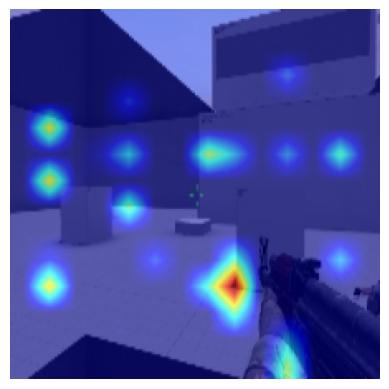}
            \subcaption{ViT Tiny}
        \end{minipage}
    \end{adjustbox}
    
    \begin{adjustbox}{valign=t}  
        \begin{minipage}{0.19\linewidth}
            \includegraphics[width=\textwidth,trim={0.5em 0.5em 0.5em 0.5em},clip]{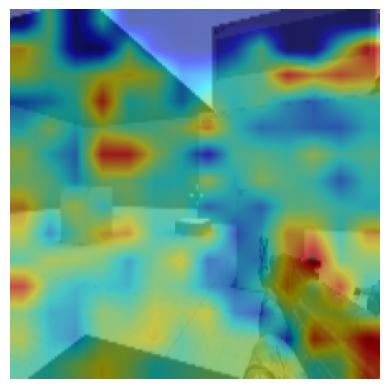}
            \subcaption{ViT Tiny +Aug}
        \end{minipage}
    \end{adjustbox}
    \begin{adjustbox}{valign=t}  
        \begin{minipage}{0.19\linewidth}
            \includegraphics[width=\textwidth,trim={0.5em 0.5em 0.5em 0.5em},clip]{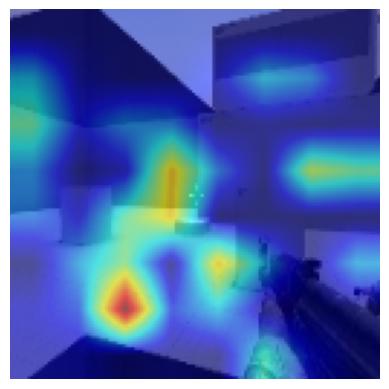}
            \subcaption{ViT 128}
        \end{minipage}
    \end{adjustbox}
    \begin{adjustbox}{valign=t}  
        \begin{minipage}{0.19\linewidth}
            \includegraphics[width=\textwidth,trim={0.5em 0.5em 0.5em 0.5em},clip]{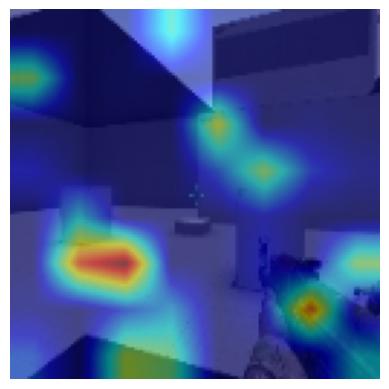}
            \subcaption{ViT 128 +Aug}
        \end{minipage}
    \end{adjustbox}
    \begin{adjustbox}{valign=t}  
        \begin{minipage}{0.19\linewidth}
            \includegraphics[width=\textwidth,trim={0.5em 0.5em 0.5em 0.5em},clip]{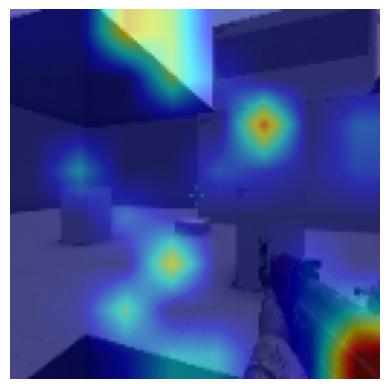}
            \subcaption{ViT 256}
        \end{minipage}
    \end{adjustbox}
    \begin{adjustbox}{valign=t}  
        \begin{minipage}{0.19\linewidth}
            \includegraphics[width=\textwidth,trim={0.5em 0.5em 0.5em 0.5em},clip]{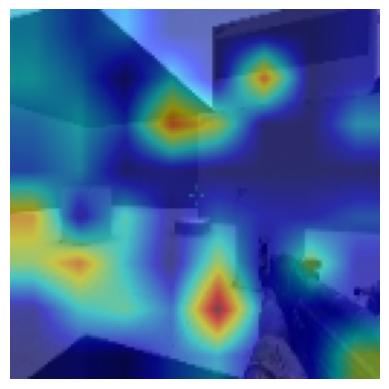}
            \subcaption{ViT 256 +Aug}
        \end{minipage}
    \end{adjustbox}

    \begin{adjustbox}{valign=t}  
        \begin{minipage}{0.19\linewidth}
            \includegraphics[width=\textwidth,trim={0.5em 0.5em 0.5em 0.5em},clip]{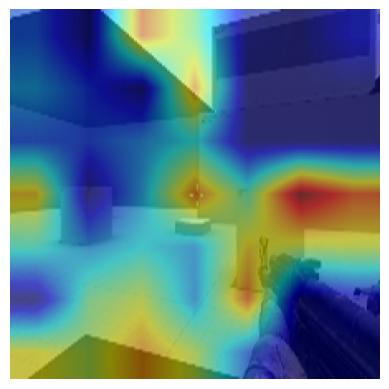}
            \subcaption{CLIP RN50}
        \end{minipage}
    \end{adjustbox}
    \begin{adjustbox}{valign=t}  
        \begin{minipage}{0.19\linewidth}
            \includegraphics[width=\textwidth,trim={0.5em 0.5em 0.5em 0.5em},clip]{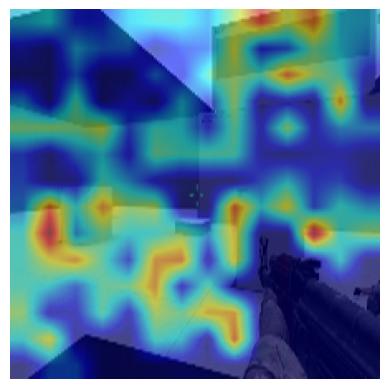}
            \subcaption{CLIP ViT-B/16}
        \end{minipage}
    \end{adjustbox}
    \begin{adjustbox}{valign=t}  
        \begin{minipage}{0.19\linewidth}
            \includegraphics[width=\textwidth,trim={0.5em 0.5em 0.5em 0.5em},clip]{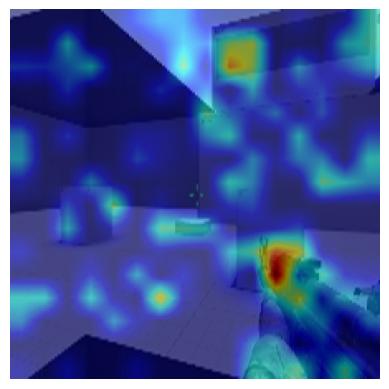}
            \subcaption{CLIP ViT-L/14}
        \end{minipage}
    \end{adjustbox}
    \begin{adjustbox}{valign=t}  
        \begin{minipage}{0.19\linewidth}
            \includegraphics[width=\textwidth,trim={0.5em 0.5em 0.5em 0.5em},clip]{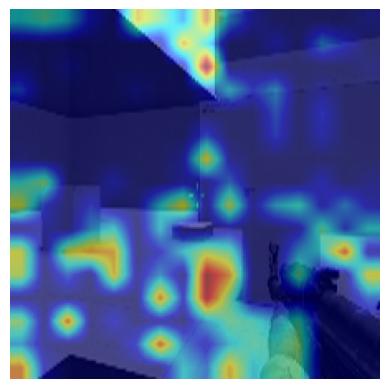}
            \subcaption{DINOv2 ViT-S/14}
        \end{minipage}
    \end{adjustbox}
    \begin{adjustbox}{valign=t}  
        \begin{minipage}{0.19\linewidth}
            \includegraphics[width=\textwidth,trim={0.5em 0.5em 0.5em 0.5em},clip]{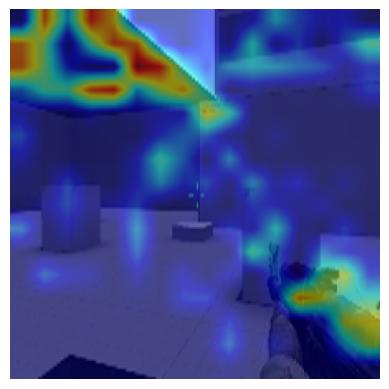}
            \subcaption{DINOv2 ViT-B/14}
        \end{minipage}
    \end{adjustbox}

    \begin{adjustbox}{valign=t}  
        \begin{minipage}{0.19\linewidth}
            \includegraphics[width=\textwidth,trim={0.5em 0.5em 0.5em 0.5em},clip]{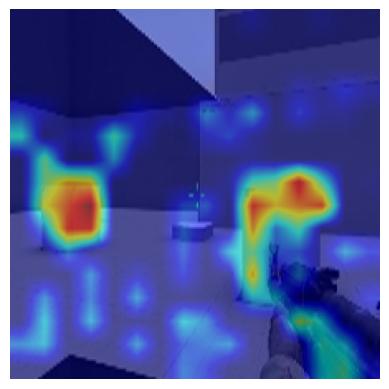}
            \subcaption{DINOv2 ViT-L/14}
        \end{minipage}
    \end{adjustbox}
    \begin{adjustbox}{valign=t}  
        \begin{minipage}{0.19\linewidth}
            \includegraphics[width=\textwidth,trim={0.5em 0.5em 0.5em 0.5em},clip]{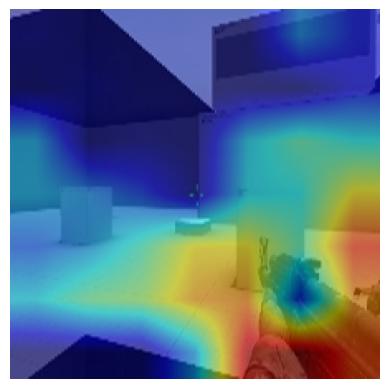}
            \subcaption{Focal Large}
        \end{minipage}
    \end{adjustbox}
    \begin{adjustbox}{valign=t}  
        \begin{minipage}{0.19\linewidth}
            \includegraphics[width=\textwidth,trim={0.5em 0.5em 0.5em 0.5em},clip]{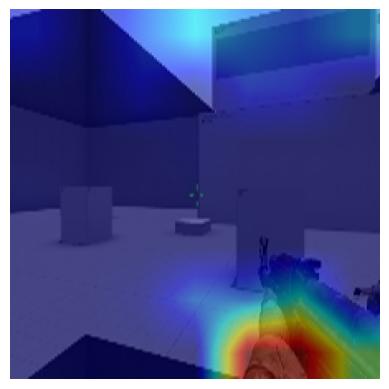}
            \subcaption{Focal XLarge}
        \end{minipage}
    \end{adjustbox}
    \begin{adjustbox}{valign=t}  
        \begin{minipage}{0.19\linewidth}
            \includegraphics[width=\textwidth,trim={0.5em 0.5em 0.5em 0.5em},clip]{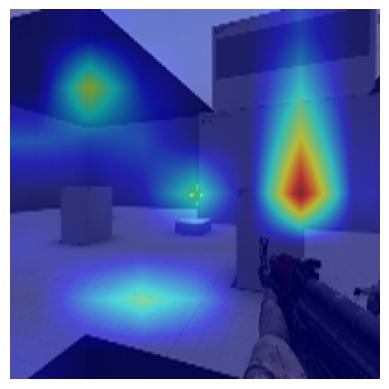}
            \subcaption{Focal Huge}
        \end{minipage}
    \end{adjustbox}
    \begin{adjustbox}{valign=t}  
        \begin{minipage}{0.19\linewidth}
            \includegraphics[width=\textwidth,trim={0.5em 0.5em 0.5em 0.5em},clip]{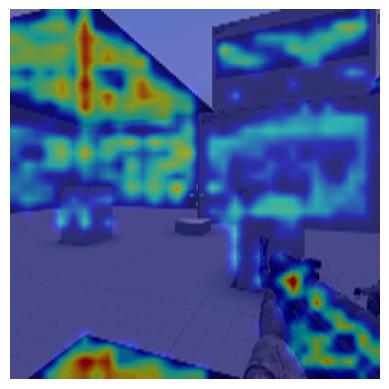}
            \subcaption{SD VAE}
        \end{minipage}
    \end{adjustbox}

    \caption{Grad-Cam visualisations for all encoders (seed 0) in Counter Strike with policy action logits serving as the targets.}
    \label{fig:grad_cam_csgo_hdf5_aim_july2021_expert_10_image_200}
\end{figure*}

\begin{figure*}[h]
    \centering
    \begin{adjustbox}{valign=t}  
        \begin{minipage}{0.335\linewidth} 
            \centering
            \includegraphics[width=\textwidth]{images/grad_cam/csgo/original_images/hdf5_aim_july2021_expert_10_image_800.jpg}
            \subcaption{Original image}
        \end{minipage}
    \end{adjustbox}
    \hspace{3cm}
    \begin{adjustbox}{valign=t}  
        \begin{minipage}{0.2\linewidth}
            \includegraphics[width=\textwidth,trim={0.5em 0.5em 0.5em 0.5em},clip]{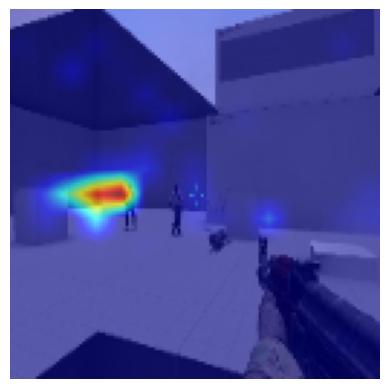}
            \subcaption{Impala ResNet}
        \end{minipage}
    \end{adjustbox}
    \begin{adjustbox}{valign=t}  
        \begin{minipage}{0.2\linewidth}
            \includegraphics[width=\textwidth,trim={0.5em 0.5em 0.5em 0.5em},clip]{images/grad_cam/csgo/end_to_end_encoders/csgo_e2e_impala_resnet_128_seed0_hdf5_aim_july2021_expert_10_image_800.jpg}
            \subcaption{Impala ResNet +Aug}
        \end{minipage}
    \end{adjustbox}

    \begin{adjustbox}{valign=t}  
        \begin{minipage}{0.19\linewidth}
            \includegraphics[width=\textwidth,trim={0.5em 0.5em 0.5em 0.5em},clip]{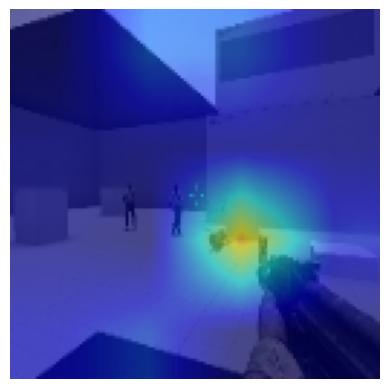}
            \subcaption{ResNet 128}
        \end{minipage}
    \end{adjustbox}
    \begin{adjustbox}{valign=t}  
        \begin{minipage}{0.19\linewidth}
            \includegraphics[width=\textwidth,trim={0.5em 0.5em 0.5em 0.5em},clip]{images/grad_cam/csgo/end_to_end_encoders/dungeons_csgo_own_resnet_128_seed_0_hdf5_aim_july2021_expert_10_image_800.jpg}
            \subcaption{ResNet 128 +Aug}
        \end{minipage}
    \end{adjustbox}
    \begin{adjustbox}{valign=t}  
        \begin{minipage}{0.19\linewidth}
            \includegraphics[width=\textwidth,trim={0.5em 0.5em 0.5em 0.5em},clip]{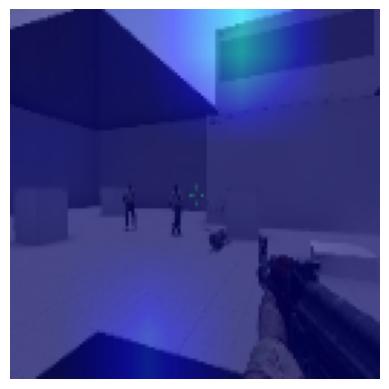}
            \subcaption{ResNet 256}
        \end{minipage}
    \end{adjustbox}
    \begin{adjustbox}{valign=t}  
        \begin{minipage}{0.19\linewidth}
            \includegraphics[width=\textwidth,trim={0.5em 0.5em 0.5em 0.5em},clip]{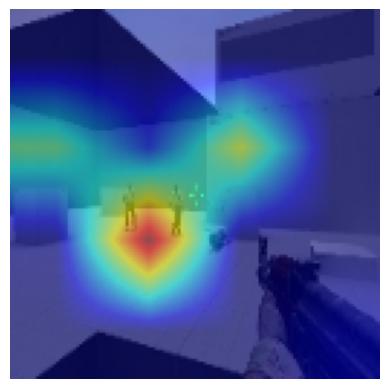}
            \subcaption{ResNet 256 +Aug}
        \end{minipage}
    \end{adjustbox}
    \begin{adjustbox}{valign=t}  
        \begin{minipage}{0.19\linewidth}
            \includegraphics[width=\textwidth,trim={0.5em 0.5em 0.5em 0.5em},clip]{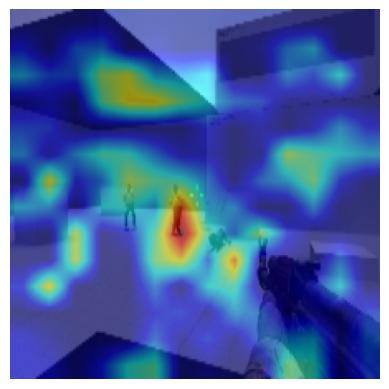}
            \subcaption{ViT Tiny}
        \end{minipage}
    \end{adjustbox}
    
    \begin{adjustbox}{valign=t}  
        \begin{minipage}{0.19\linewidth}
            \includegraphics[width=\textwidth,trim={0.5em 0.5em 0.5em 0.5em},clip]{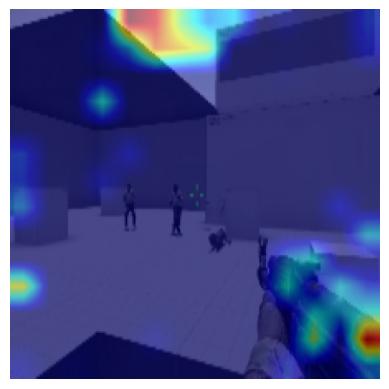}
            \subcaption{ViT Tiny +Aug}
        \end{minipage}
    \end{adjustbox}
    \begin{adjustbox}{valign=t}  
        \begin{minipage}{0.19\linewidth}
            \includegraphics[width=\textwidth,trim={0.5em 0.5em 0.5em 0.5em},clip]{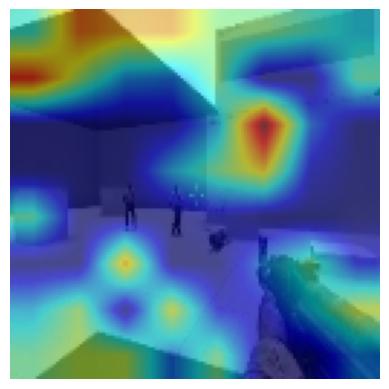}
            \subcaption{ViT 128}
        \end{minipage}
    \end{adjustbox}
    \begin{adjustbox}{valign=t}  
        \begin{minipage}{0.19\linewidth}
            \includegraphics[width=\textwidth,trim={0.5em 0.5em 0.5em 0.5em},clip]{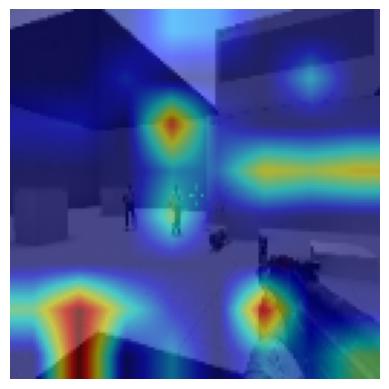}
            \subcaption{ViT 128 +Aug}
        \end{minipage}
    \end{adjustbox}
    \begin{adjustbox}{valign=t}  
        \begin{minipage}{0.19\linewidth}
            \includegraphics[width=\textwidth,trim={0.5em 0.5em 0.5em 0.5em},clip]{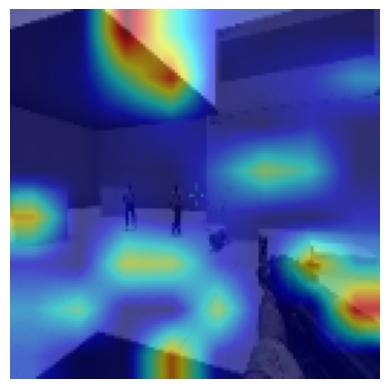}
            \subcaption{ViT 256}
        \end{minipage}
    \end{adjustbox}
    \begin{adjustbox}{valign=t}  
        \begin{minipage}{0.19\linewidth}
            \includegraphics[width=\textwidth,trim={0.5em 0.5em 0.5em 0.5em},clip]{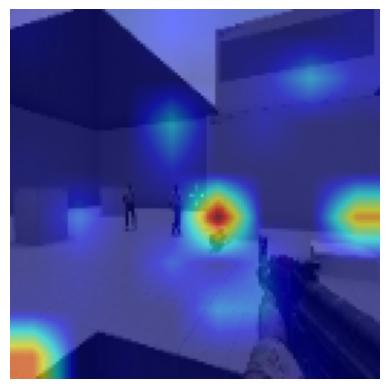}
            \subcaption{ViT 256 +Aug}
        \end{minipage}
    \end{adjustbox}

    \begin{adjustbox}{valign=t}  
        \begin{minipage}{0.19\linewidth}
            \includegraphics[width=\textwidth,trim={0.5em 0.5em 0.5em 0.5em},clip]{images/grad_cam/csgo/pretrained_encoders/csgo_pretrained_clip_rn50_seed0_hdf5_aim_july2021_expert_10_image_800.jpg}
            \subcaption{CLIP RN50}
        \end{minipage}
    \end{adjustbox}
    \begin{adjustbox}{valign=t}  
        \begin{minipage}{0.19\linewidth}
            \includegraphics[width=\textwidth,trim={0.5em 0.5em 0.5em 0.5em},clip]{images/grad_cam/csgo/pretrained_encoders/csgo_pretrained_clip_vitb_seed0_hdf5_aim_july2021_expert_10_image_800.jpg}
            \subcaption{CLIP ViT-B/16}
        \end{minipage}
    \end{adjustbox}
    \begin{adjustbox}{valign=t}  
        \begin{minipage}{0.19\linewidth}
            \includegraphics[width=\textwidth,trim={0.5em 0.5em 0.5em 0.5em},clip]{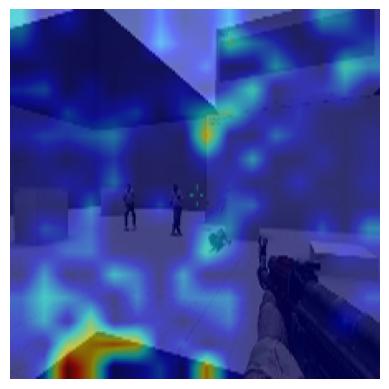}
            \subcaption{CLIP ViT-L/14}
        \end{minipage}
    \end{adjustbox}
    \begin{adjustbox}{valign=t}  
        \begin{minipage}{0.19\linewidth}
            \includegraphics[width=\textwidth,trim={0.5em 0.5em 0.5em 0.5em},clip]{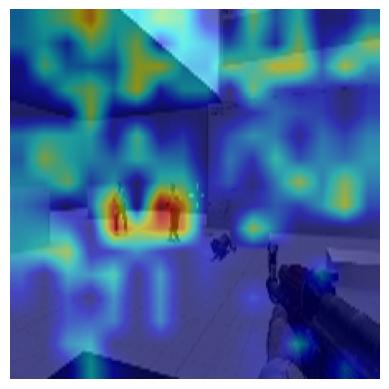}
            \subcaption{DINOv2 ViT-S/14}
        \end{minipage}
    \end{adjustbox}
    \begin{adjustbox}{valign=t}  
        \begin{minipage}{0.19\linewidth}
            \includegraphics[width=\textwidth,trim={0.5em 0.5em 0.5em 0.5em},clip]{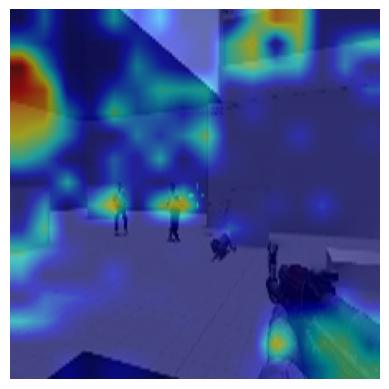}
            \subcaption{DINOv2 ViT-B/14}
        \end{minipage}
    \end{adjustbox}

    \begin{adjustbox}{valign=t}  
        \begin{minipage}{0.19\linewidth}
            \includegraphics[width=\textwidth,trim={0.5em 0.5em 0.5em 0.5em},clip]{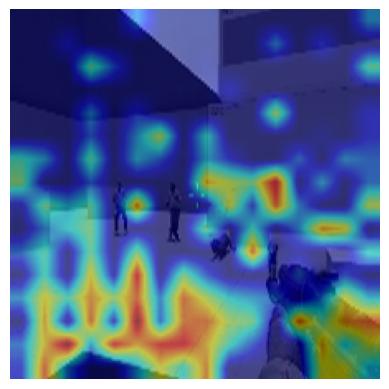}
            \subcaption{DINOv2 ViT-L/14}
        \end{minipage}
    \end{adjustbox}
    \begin{adjustbox}{valign=t}  
        \begin{minipage}{0.19\linewidth}
            \includegraphics[width=\textwidth,trim={0.5em 0.5em 0.5em 0.5em},clip]{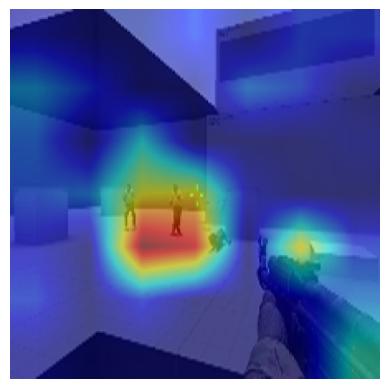}
            \subcaption{Focal Large}
        \end{minipage}
    \end{adjustbox}
    \begin{adjustbox}{valign=t}  
        \begin{minipage}{0.19\linewidth}
            \includegraphics[width=\textwidth,trim={0.5em 0.5em 0.5em 0.5em},clip]{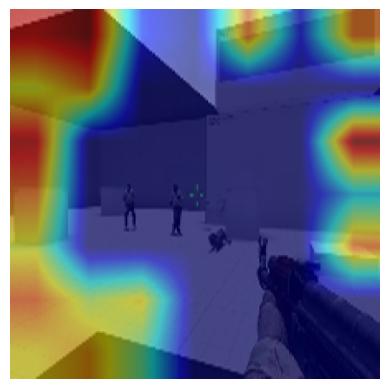}
            \subcaption{Focal XLarge}
        \end{minipage}
    \end{adjustbox}
    \begin{adjustbox}{valign=t}  
        \begin{minipage}{0.19\linewidth}
            \includegraphics[width=\textwidth,trim={0.5em 0.5em 0.5em 0.5em},clip]{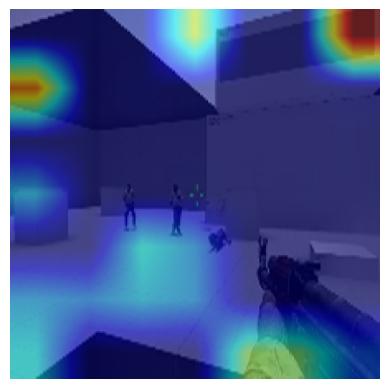}
            \subcaption{Focal Huge}
        \end{minipage}
    \end{adjustbox}
    \begin{adjustbox}{valign=t}  
        \begin{minipage}{0.19\linewidth}
            \includegraphics[width=\textwidth,trim={0.5em 0.5em 0.5em 0.5em},clip]{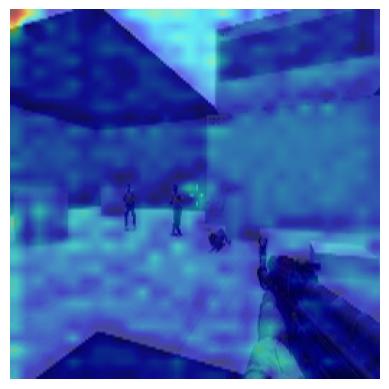}
            \subcaption{SD VAE}
        \end{minipage}
    \end{adjustbox}

    \caption{Grad-Cam visualisations for all encoders (seed 0) in Counter Strike with policy action logits serving as the targets.}
    \label{fig:grad_cam_csgo_hdf5_aim_july2021_expert_10_image_800}
\end{figure*}

\begin{figure*}[h]
    \centering
    \begin{adjustbox}{valign=t}  
        \begin{minipage}{0.335\linewidth} 
            \centering
            \includegraphics[width=\textwidth]{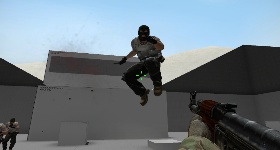}
            \subcaption{Original image}
        \end{minipage}
    \end{adjustbox}
    \hspace{3cm}
    \begin{adjustbox}{valign=t}  
        \begin{minipage}{0.2\linewidth}
            \includegraphics[width=\textwidth,trim={0.5em 0.5em 0.5em 0.5em},clip]{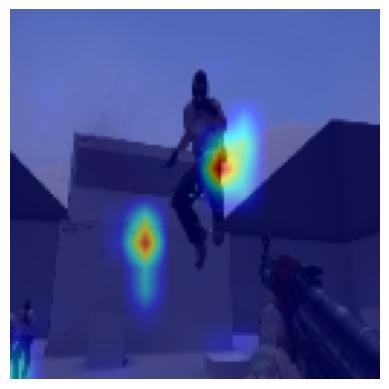}
            \subcaption{Impala ResNet}
        \end{minipage}
    \end{adjustbox}
    \begin{adjustbox}{valign=t}  
        \begin{minipage}{0.2\linewidth}
            \includegraphics[width=\textwidth,trim={0.5em 0.5em 0.5em 0.5em},clip]{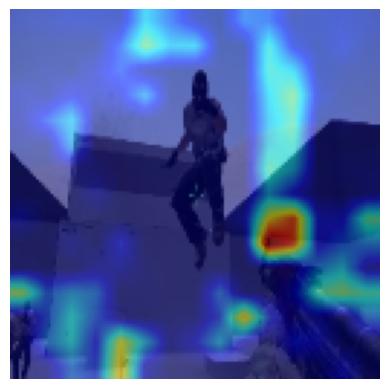}
            \subcaption{Impala ResNet +Aug}
        \end{minipage}
    \end{adjustbox}

    \begin{adjustbox}{valign=t}  
        \begin{minipage}{0.19\linewidth}
            \includegraphics[width=\textwidth,trim={0.5em 0.5em 0.5em 0.5em},clip]{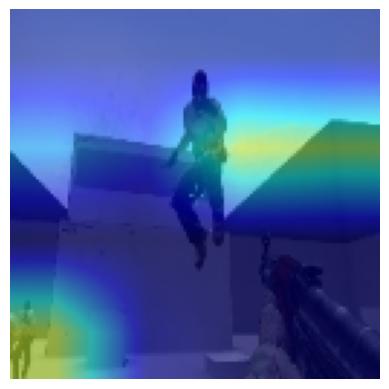}
            \subcaption{ResNet 128}
        \end{minipage}
    \end{adjustbox}
    \begin{adjustbox}{valign=t}  
        \begin{minipage}{0.19\linewidth}
            \includegraphics[width=\textwidth,trim={0.5em 0.5em 0.5em 0.5em},clip]{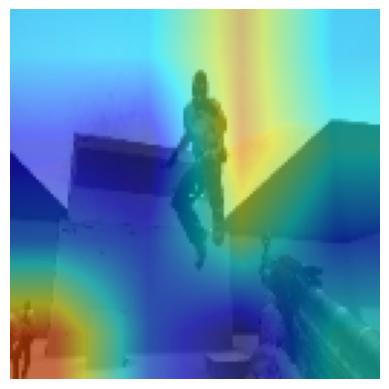}
            \subcaption{ResNet 128 +Aug}
        \end{minipage}
    \end{adjustbox}
    \begin{adjustbox}{valign=t}  
        \begin{minipage}{0.19\linewidth}
            \includegraphics[width=\textwidth,trim={0.5em 0.5em 0.5em 0.5em},clip]{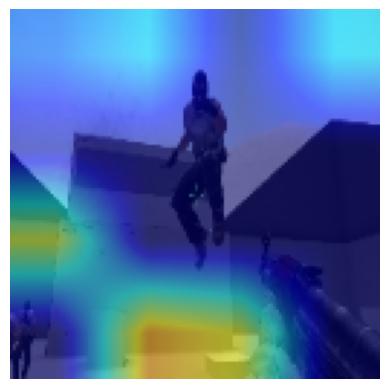}
            \subcaption{ResNet 256}
        \end{minipage}
    \end{adjustbox}
    \begin{adjustbox}{valign=t}  
        \begin{minipage}{0.19\linewidth}
            \includegraphics[width=\textwidth,trim={0.5em 0.5em 0.5em 0.5em},clip]{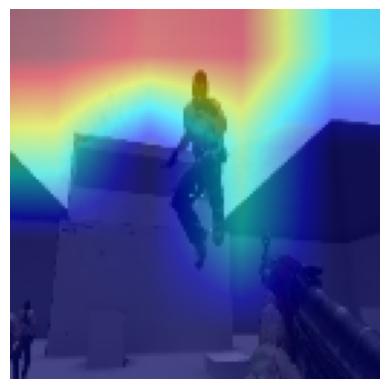}
            \subcaption{ResNet 256 +Aug}
        \end{minipage}
    \end{adjustbox}
    \begin{adjustbox}{valign=t}  
        \begin{minipage}{0.19\linewidth}
            \includegraphics[width=\textwidth,trim={0.5em 0.5em 0.5em 0.5em},clip]{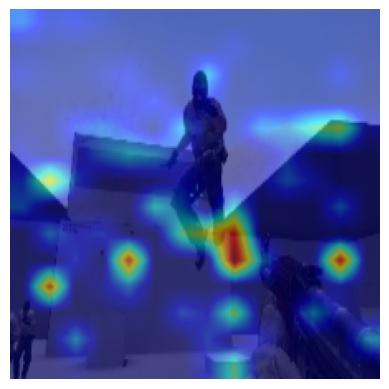}
            \subcaption{ViT Tiny}
        \end{minipage}
    \end{adjustbox}
    
    \begin{adjustbox}{valign=t}  
        \begin{minipage}{0.19\linewidth}
            \includegraphics[width=\textwidth,trim={0.5em 0.5em 0.5em 0.5em},clip]{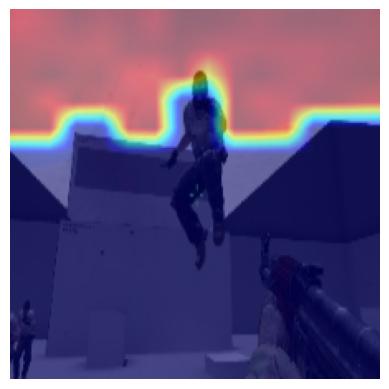}
            \subcaption{ViT Tiny +Aug}
        \end{minipage}
    \end{adjustbox}
    \begin{adjustbox}{valign=t}  
        \begin{minipage}{0.19\linewidth}
            \includegraphics[width=\textwidth,trim={0.5em 0.5em 0.5em 0.5em},clip]{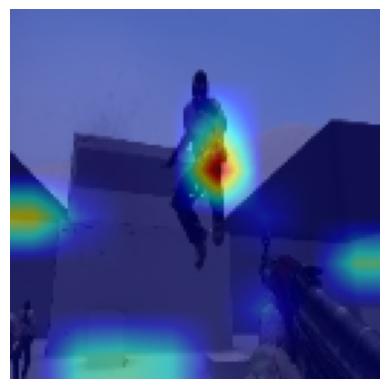}
            \subcaption{ViT 128}
        \end{minipage}
    \end{adjustbox}
    \begin{adjustbox}{valign=t}  
        \begin{minipage}{0.19\linewidth}
            \includegraphics[width=\textwidth,trim={0.5em 0.5em 0.5em 0.5em},clip]{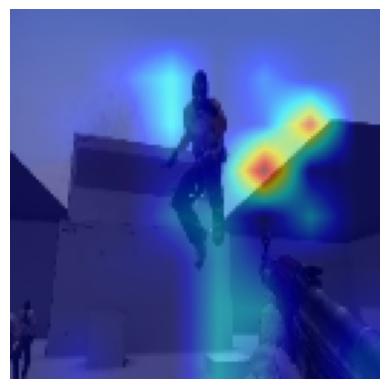}
            \subcaption{ViT 128 +Aug}
        \end{minipage}
    \end{adjustbox}
    \begin{adjustbox}{valign=t}  
        \begin{minipage}{0.19\linewidth}
            \includegraphics[width=\textwidth,trim={0.5em 0.5em 0.5em 0.5em},clip]{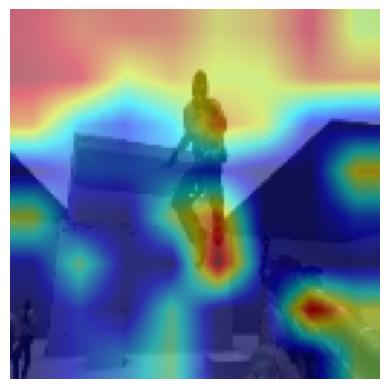}
            \subcaption{ViT 256}
        \end{minipage}
    \end{adjustbox}
    \begin{adjustbox}{valign=t}  
        \begin{minipage}{0.19\linewidth}
            \includegraphics[width=\textwidth,trim={0.5em 0.5em 0.5em 0.5em},clip]{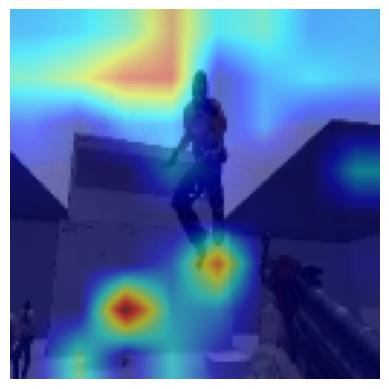}
            \subcaption{ViT 256 +Aug}
        \end{minipage}
    \end{adjustbox}

    \begin{adjustbox}{valign=t}  
        \begin{minipage}{0.19\linewidth}
            \includegraphics[width=\textwidth,trim={0.5em 0.5em 0.5em 0.5em},clip]{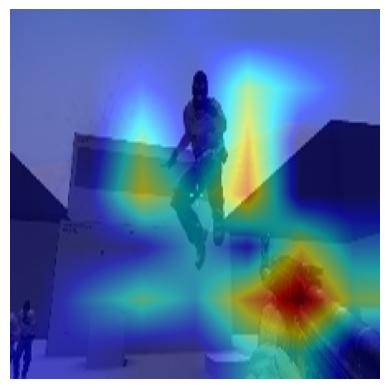}
            \subcaption{CLIP RN50}
        \end{minipage}
    \end{adjustbox}
    \begin{adjustbox}{valign=t}  
        \begin{minipage}{0.19\linewidth}
            \includegraphics[width=\textwidth,trim={0.5em 0.5em 0.5em 0.5em},clip]{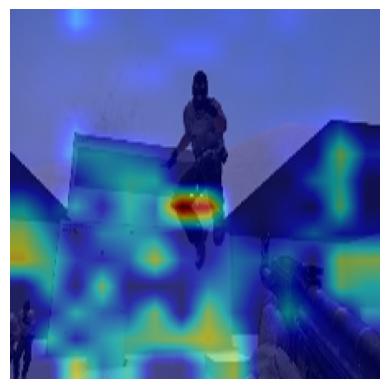}
            \subcaption{CLIP ViT-B/16}
        \end{minipage}
    \end{adjustbox}
    \begin{adjustbox}{valign=t}  
        \begin{minipage}{0.19\linewidth}
            \includegraphics[width=\textwidth,trim={0.5em 0.5em 0.5em 0.5em},clip]{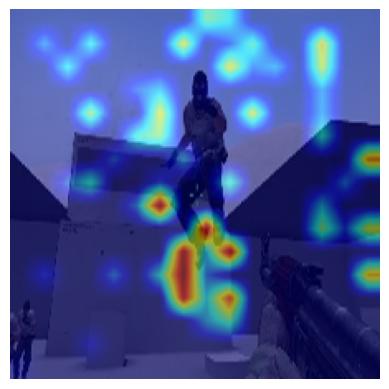}
            \subcaption{CLIP ViT-L/14}
        \end{minipage}
    \end{adjustbox}
    \begin{adjustbox}{valign=t}  
        \begin{minipage}{0.19\linewidth}
            \includegraphics[width=\textwidth,trim={0.5em 0.5em 0.5em 0.5em},clip]{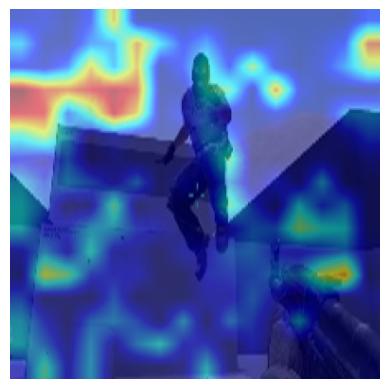}
            \subcaption{DINOv2 ViT-S/14}
        \end{minipage}
    \end{adjustbox}
    \begin{adjustbox}{valign=t}  
        \begin{minipage}{0.19\linewidth}
            \includegraphics[width=\textwidth,trim={0.5em 0.5em 0.5em 0.5em},clip]{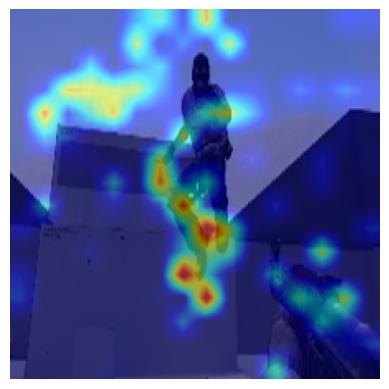}
            \subcaption{DINOv2 ViT-B/14}
        \end{minipage}
    \end{adjustbox}

    \begin{adjustbox}{valign=t}  
        \begin{minipage}{0.19\linewidth}
            \includegraphics[width=\textwidth,trim={0.5em 0.5em 0.5em 0.5em},clip]{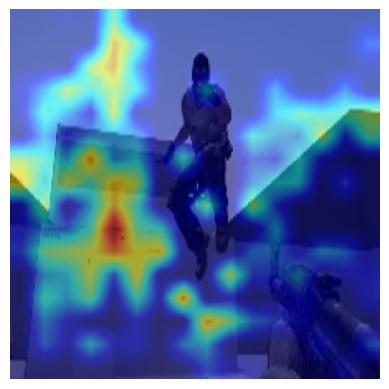}
            \subcaption{DINOv2 ViT-L/14}
        \end{minipage}
    \end{adjustbox}
    \begin{adjustbox}{valign=t}  
        \begin{minipage}{0.19\linewidth}
            \includegraphics[width=\textwidth,trim={0.5em 0.5em 0.5em 0.5em},clip]{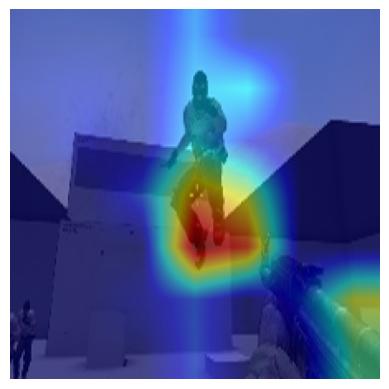}
            \subcaption{Focal Large}
        \end{minipage}
    \end{adjustbox}
    \begin{adjustbox}{valign=t}  
        \begin{minipage}{0.19\linewidth}
            \includegraphics[width=\textwidth,trim={0.5em 0.5em 0.5em 0.5em},clip]{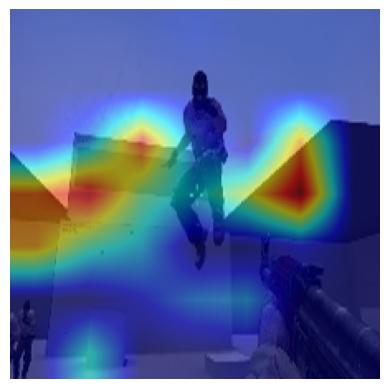}
            \subcaption{Focal XLarge}
        \end{minipage}
    \end{adjustbox}
    \begin{adjustbox}{valign=t}  
        \begin{minipage}{0.19\linewidth}
            \includegraphics[width=\textwidth,trim={0.5em 0.5em 0.5em 0.5em},clip]{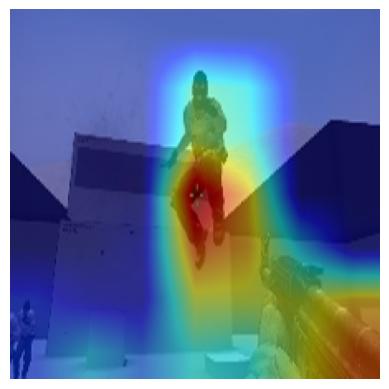}
            \subcaption{Focal Huge}
        \end{minipage}
    \end{adjustbox}
    \begin{adjustbox}{valign=t}  
        \begin{minipage}{0.19\linewidth}
            \includegraphics[width=\textwidth,trim={0.5em 0.5em 0.5em 0.5em},clip]{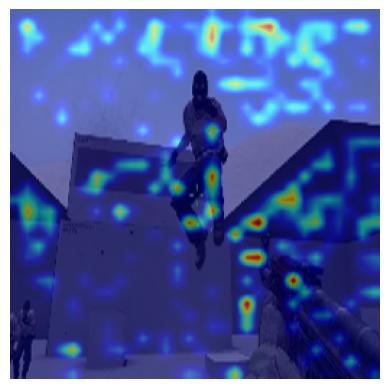}
            \subcaption{SD VAE}
        \end{minipage}
    \end{adjustbox}

    \caption{Grad-Cam visualisations for all encoders (seed 0) in Counter Strike with policy action logits serving as the targets.}
    \label{fig:grad_cam_csgo_hdf5_aim_july2021_expert_4_image_800}
\end{figure*}

\end{document}